\documentclass[runningheads]{llncs}
\usepackage{graphicx}

\usepackage{tikz}
\usepackage{comment}
\usepackage{amsmath,amssymb} 
\usepackage{color}
\usepackage{booktabs}
\usepackage{bm}
\usepackage{xcolor}

\usepackage[accsupp]{axessibility}  

\usepackage[pagebackref,breaklinks,colorlinks]{hyperref}

\definecolor{revcolor}{RGB}{255,50,0}
\definecolor{blue}{RGB}{0,0,255}

\def\boxit#1{%
  \smash{\fboxsep=0pt\llap{\rlap{\fbox{\strut\makebox[#1]{}}}~}}\ignorespaces
}

\begin{document}
\pagestyle{headings}
\mainmatter

\title{Neural Pixel Composition: \\ 3D-4D View Synthesis from Multi-Views}
\titlerunning{Neural Pixel Composition (NPC)}
\author{Aayush Bansal  \and Michael Zollhoefer}
\authorrunning{Bansal and Zollhoefer}
\institute{Reality Labs Research, Pittsburgh, USA \\
\email{\{aayushb4, zollhoefer\}@fb.com}\\
\url{http://www.aayushbansal.xyz/npc/}
}

\maketitle

\begin{abstract}

We present Neural Pixel Composition (NPC), a novel approach for continuous 3D-4D view synthesis given only a discrete set of multi-view observations as input.
Existing state-of-the-art approaches require dense multi-view supervision and an extensive computational budget.
The proposed formulation reliably operates on sparse and wide-baseline multi-view imagery and can be trained efficiently within a few seconds to $10$ minutes for hi-res (12MP) content, i.e., 200-400$\times$ faster convergence than existing methods. 
Crucial to our approach are two core novelties:
1) a representation of a pixel that contains color and depth information accumulated from multi-views for a particular location and time along a line of sight,
and 2) a multi-layer perceptron (MLP) that enables the composition of this rich information provided for a pixel location to obtain the final color output.
We experiment with a large variety of multi-view sequences, compare to existing approaches, and achieve better results in diverse and challenging settings.
Finally, our approach enables dense 3D reconstruction from sparse multi-views, where COLMAP, a state-of-the-art 3D reconstruction approach, struggles.


\end{abstract}

\section{Introduction}
\label{sec:intro}

\begin{figure*}[t]
\includegraphics[width=\linewidth]{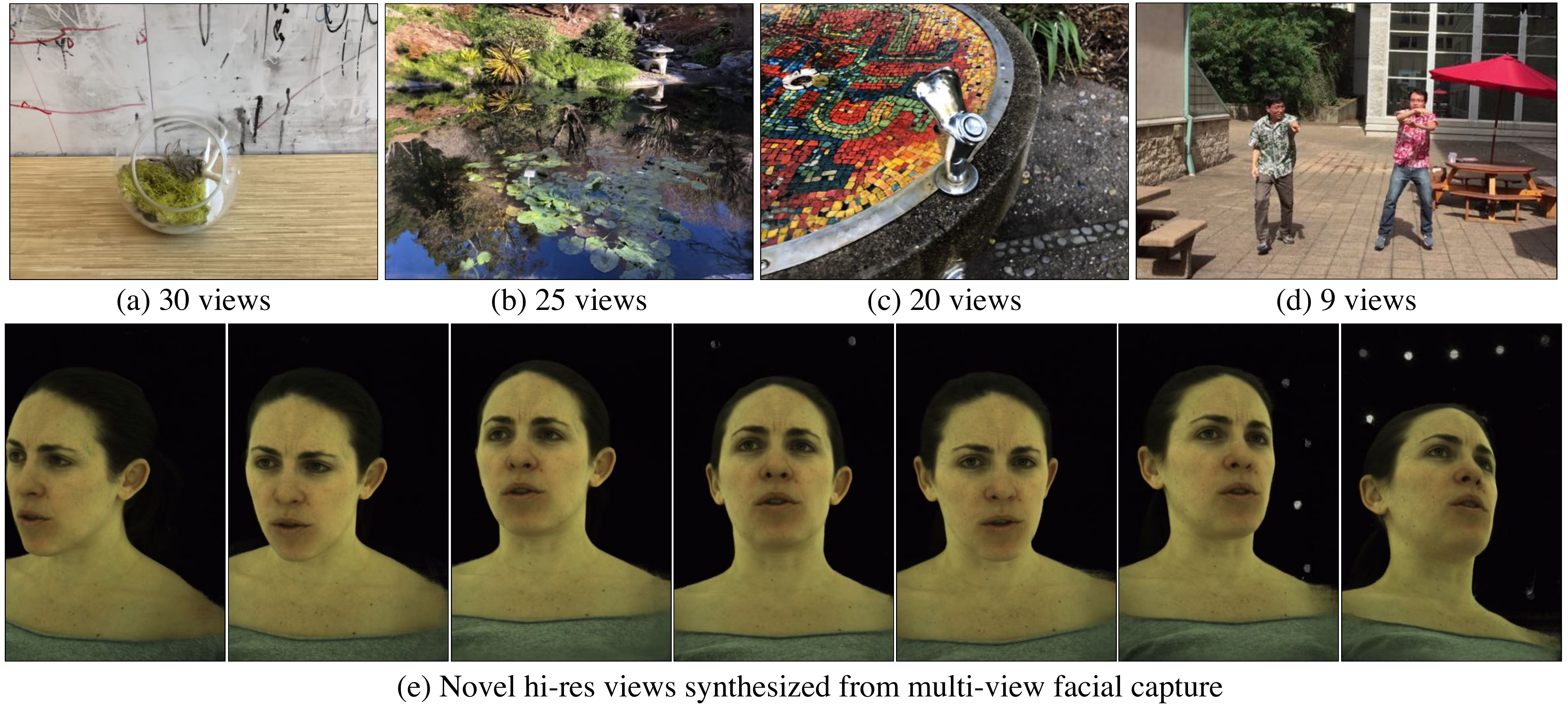}
\includegraphics[width=\linewidth]{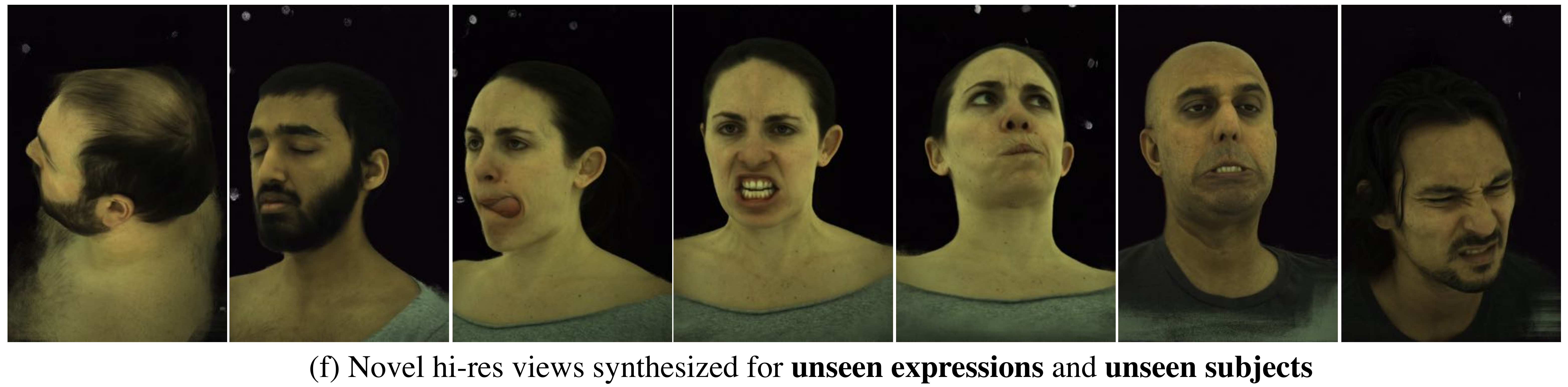}
\caption{We show \textbf{novel views} synthesized using our approach for a wide variety of multi-view sequences capturing static and dynamic environments with a varying number of views. Our approach nicely captures fine details, specular surfaces, as well as reflections as shown in {\bf (a)-(c)}. {\bf (d)} Our approach can operate on sparse and wide-baseline multi-view images with unbounded depth, and can be trained efficiently. {\bf (e)} We also show novel views capturing hi-res facial and hair details. The model is trained for a specific time instant for a given subject. This learned model {\bf (f)} generalizes to unseen expressions and unseen subjects. \textcolor{blue}{Best viewed in electronic format}.}
\label{fig:teaser_fig}
\vspace{-0.5cm}
\end{figure*}

Novel views can be readily generated if we have access to the underlying 6D plenoptic function $R(\bm{\theta}, \bm{d}, \tau)$ \cite{adelson1991plenoptic,mcmillan1995plenoptic} of the scene that models the radiance incident from direction $\bm{\theta} \in \mathbb{R}^2$ to a camera placed at position $\bm{d} \in \mathbb{R}^3$ at time $\tau$. 
Currently, no approach exists that can automatically reconstruct an efficient space- and-time representation of the plenoptic function given only a (potentially sparse) set of multi-view measurements of the scene as input. 
The core idea of image-based rendering \cite{mcmillan1999image,shade1998layered} is to generate novel views based on re-projected information from a set of calibrated source views.
This re-projection requires a high-quality estimate of the scene's geometry and is only correct for Lambertian materials, since the appearance of specular surfaces is highly view-dependent.
Building a dense 3D volume from multi-view inputs that provides correct 3D information for each pixel location is a non-trivial task.

Recent approaches such as Neural Radiance Fields (NeRF)~\cite{mildenhall2020nerf} and Neural Volumes~(NV)~\cite{Lombardi:2019} attempt to create rich 3D information along a ray of light by sampling 3D points at regular intervals given a min-max bound.
%
Radiance fields are highly flexible 3D scene representations that enables them to represent a large variety of scenes including semi-transparent objects.
The price to be paid for this flexibility is that current approaches are restricted to datasets that provide dense 3D observations~\cite{Lombardi:2019,mildenhall2020nerf,pumarola2020d,riegler2020free,Riegler_2021_CVPR,zhang2020nerf++}, can only model bounded scenes~\cite{SRF,Lombardi:2019,keypointnerf,mildenhall2020nerf,wang2021ibrnet,yu2020pixelnerf}, and require intensive computational resources~\cite{Lombardi:2019,mildenhall2020nerf,zhang2020nerf++}.
%
In contrast, we introduce a multi-view composition approach that combines the insights from image-based rendering~\cite{shum2000review} with the power of neural rendering~\cite{tewari2020NeuralRendering} by learning how to best aggregate information from different views given only imperfect depth estimates as input. Figure~\ref{fig:teaser_fig} shows novel views synthesized using our approach for different multi-view sequences.
%
Our approach can operate on sparse and wide-baseline multi-view imagery (assuming known camera parameters) and requires limited computational resources for operation. The model learned on a single time-instant for one subject (Fig~\ref{fig:teaser_fig}-(e)) generalizes to unseen time instances and unseen subjects without any fine-tuning (Fig~\ref{fig:teaser_fig}-(f)). 

\begin{figure}[t]
\includegraphics[width=\linewidth]{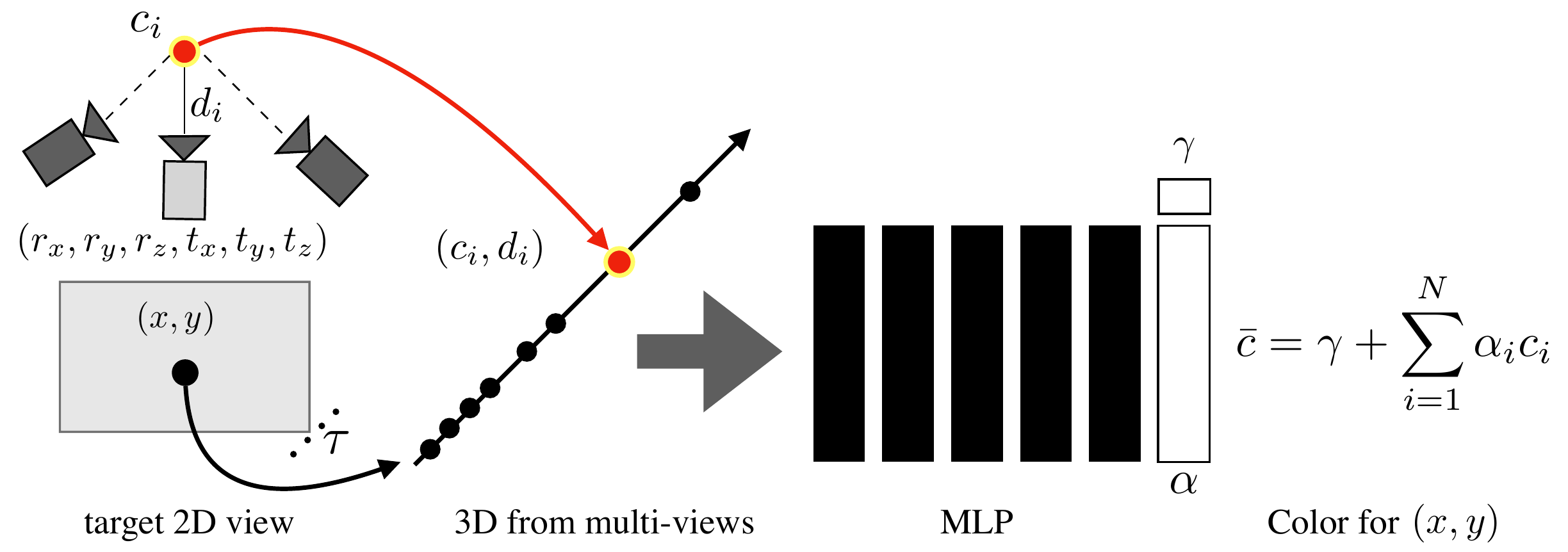}
\caption{\textbf{Color for a pixel location:}
Our goal is to estimate the color for every pixel location $(x,y)$ for a time $\tau$ given camera extrinsic parameters $(r_x, r_y, r_z, t_x, t_y, t_z)$.
We collect a rich 3D descriptor consisting of color ($\bm{c}$) and depth ($\bm{d}$) information from multiple stereo-pairs using an off-the-shelf disparity estimation module~\cite{Yang_2019_CVPR}.
We learn a multi-layer perceptron (MLP) to compose color and depth.
The final output color $\bar{c}$ is obtained by a simple dot-product of a blending weight $\bm{\alpha}$ (output of MLP) and the corresponding color samples.
$\gamma$ is a regressed color correction term per pixel.
}
\label{fig:intro}
\end{figure}

We accumulate rich 3D information (color and depth) for a pixel location using an off-the-shelf disparity estimation approach~\cite{Yang_2019_CVPR} given multiple stereo pairs as input.
We then learn a small multi-layer perceptron (MLP) for a given multi-view sequence that inputs the per-pixel information at a given camera position and outputs color at the location. Figure~\ref{fig:intro} illustrates the components of our approach.
We train an MLP for a sequence by sampling random pixels given multi-views.
In our experiments, we observe that a simple $5$-layer perceptron is sufficient to generate high-quality results.
Our model roughly requires $1$ GB of GPU memory and can be trained within a few seconds to $10$ minutes from scratch for a hi-res multi-view sequence.
The trained model allows us to perform a single forward-pass at test time for each pixel location in a target camera view. A single forward pass per pixel is more efficient than radiance field based approaches that require hundreds of samples along each ray. 
Finally, the alpha values ($\alpha_i$) allow us to perform dense 3D reconstruction of the scene by selecting appropriate depth values at a given location.

In summary, our \textbf{contributions} are:
\begin{itemize}
    \item A surprisingly simple, yet effective approach for view synthesis from calibrated multi-view images that works with limited computational resources on diverse multi-view sequences.
    \item Our approach offers a natural extension to the 4D view synthesis problem. Our approach can also generalize to unseen time instances.
    \item Our approach is able to obtain dense 3D reconstruction on challenging in-the-wild scenes.
\end{itemize}

\section{Related Work}
\label{sec:related}

Our novel view synthesis work is closely related to several research domains, such as classical 3D reconstruction and plenoptic modeling, as well as neural rendering for static and dynamic scenes.
In the following, we cover the most related approaches.
For a detailed discussion of neural rendering approaches, we refer to the surveys \cite{shum2000review,tewari2020NeuralRendering,tewari2021NeuralAdvances}.

\begin{figure*}[t]
\includegraphics[width=\linewidth]{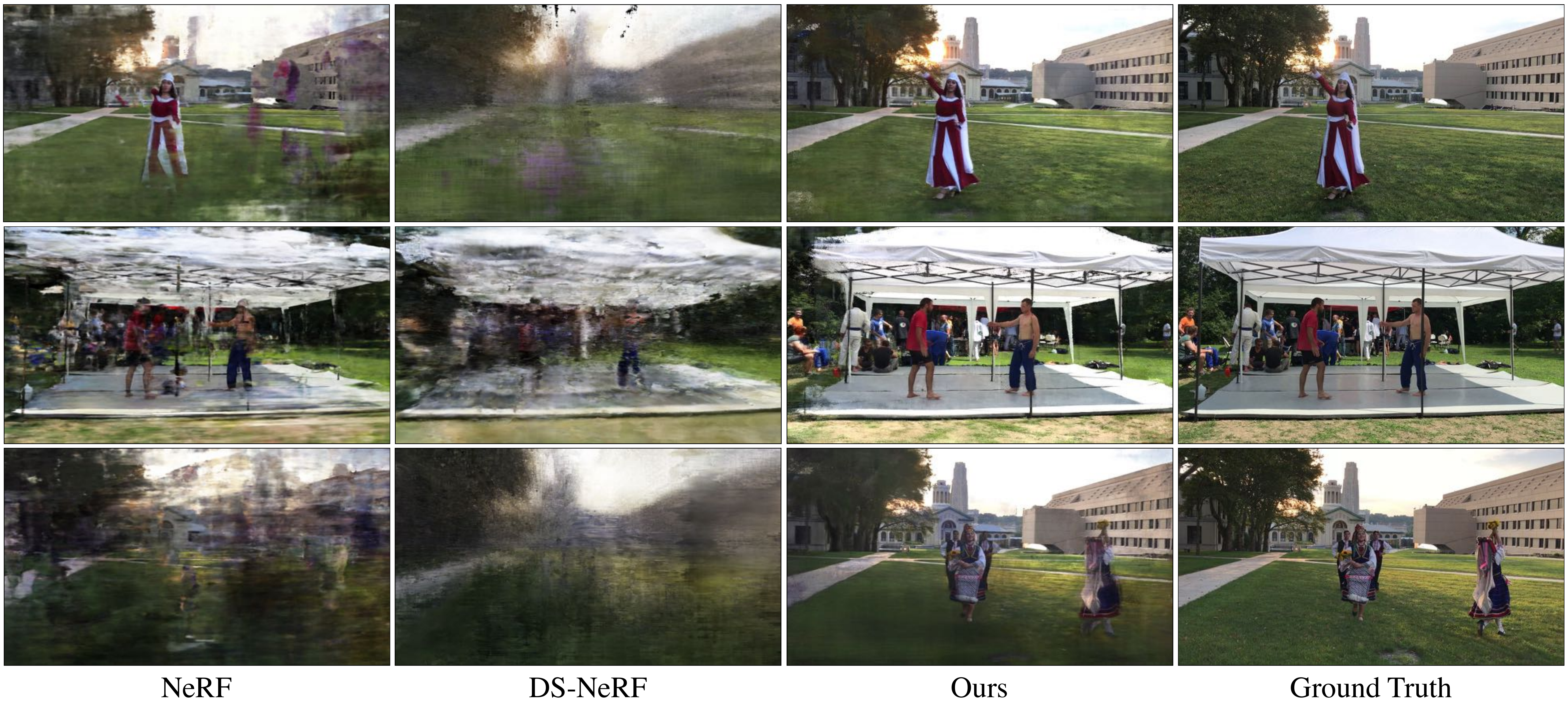}
\includegraphics[width=\linewidth]{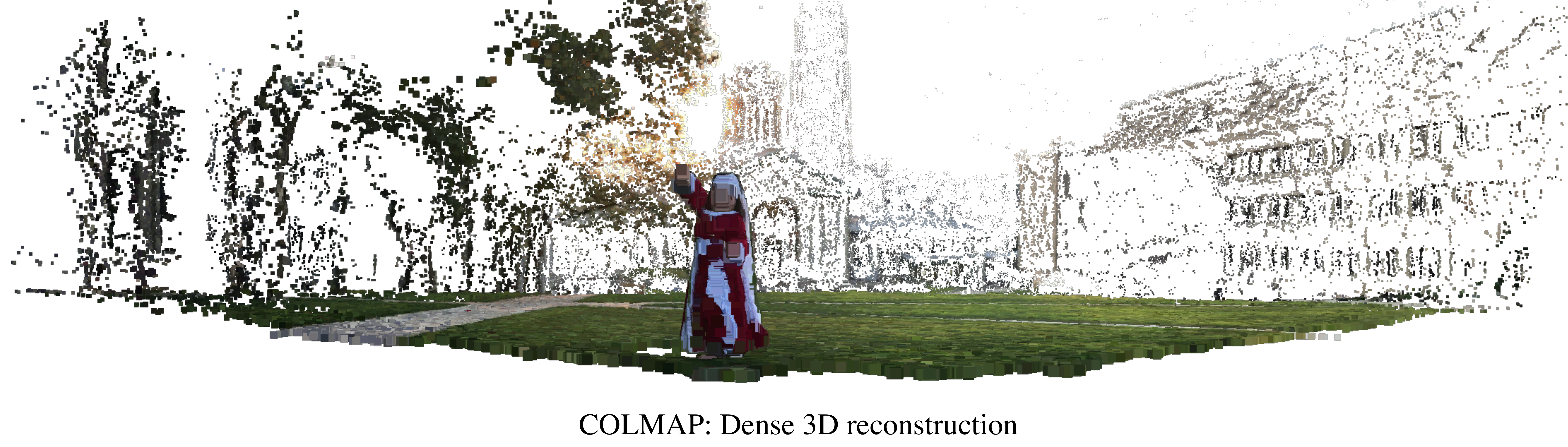}
\caption{\textbf{View synthesis given sparse and spread-out multi-views: } Our approach allows us to operate on sparse multi-views of unbounded scenes~\cite{Bansal_2020_CVPR}. We show novel view points for a fixed time instant for three unbounded scenes. Prior approaches such as NeRF~\cite{mildenhall2020nerf} and DS-NeRF~\cite{deng2021depth} lead to degenerate outputs on these sequences. We also show the 3D reconstruction using COLMAP~\cite{schoenberger2016mvs,schoenberger2016sfm} for the sequence in the top-row. We observe that dense 3D reconstruction from sparse views is non-trivial for COLMAP.} 
\vspace{-0.5cm}
\label{fig:teaser_fig_b}
\end{figure*}

\noindent\textbf{Plenoptic Modeling and NeRF:} Plenoptic function~\cite{adelson1991plenoptic,mcmillan1995plenoptic} does not require geometric modeling.
A plenoptic or a light-field camera~\cite{gortler1996lumigraph,levoy1996light,ng2005light} captures all possible rays of light (in a bounded scene), which in turns enables the synthesis of a new view via a per-ray look-up.
Recent approaches such as NeRF~\cite{mildenhall2020nerf} and follow-up work~\cite{MIPNerF,wu2021diver,zhang2020nerf++} employ a multi-layer perceptron (MLP) that infers color and opacity values at 3D locations along each camera ray.
These color and opacity values along the ray are then being integrated to obtain the final pixel color.
This requires:
1) dense multi-view inputs~\cite{SRF,yu2020pixelnerf};
2) perfect camera parameters~\cite{SCNeRF2021,lin2021barf};
and 3) a min-max bound to sample 3D points along a ray of light~\cite{riegler2020free,zhang2020nerf++}. We observe degenerate outputs if all three conditions are not met (as shown in Figure~\ref{fig:teaser_fig_b}).
Different approaches either use prior knowledge or a large number of multi-view sequences~\cite{SRF,wang2021ibrnet,yu2020pixelnerf}, additional geometric optimization~\cite{SCNeRF2021,lin2021barf}, or large capacity models to separately capture foreground and background~\cite{zhang2020nerf++}.
In this work, we use an off-the-shelf disparity estimation module~\cite{Yang_2019_CVPR} that allows us to accumulate 3D information for a given pixel location. 
A simple MLP provides us with blending parameters that enable the composition of color information.
This allows us to overcome the above-mentioned three challenges albeit using limited computational resources to train/test the model.

\noindent\textbf{3D Reconstruction and View Synthesis:} Another approach to solve the problem is to obtain dense 3D reconstruction from the input images~\cite{hartley2003multiple} and project 3D points to the target view.
There has been immense progress in  densely reconstructing the static world from multi-view imagery~\cite{furukawa2015multi,kanade2006historical}, internet scale photos~\cite{Agarwal2009,heinly_dissertation,schoenberger2016sfm,Snavely:2006}, and videos~\cite{schoenberger2016mvs}.
Synthesizing a novel view from accumulated 3D point clouds may not be consistent due to varying illumination, specular material, and different cameras used for the capture of the various viewpoints.
Riegler et al.~\cite{riegler2020free,Riegler_2021_CVPR} use a neural network to obtain consistent visuals given a dense 3D reconstruction.
This works well for dense multi-view observations~\cite{Knapitsch2017}.
However, 3D reconstruction is sparse given wide-baseline views or scenes with specular surfaces.
This is highlighted in Figure~\ref{fig:teaser_fig_b}, which shows 3D reconstruction results of COLMAP~\cite{schoenberger2016sfm,schoenberger2016mvs} using one of the sequences. Recently, DS-NeRF~\cite{deng2021depth} use sparse 3D points from COLMAP along with NeRF to learn better and faster view synthesis. As shown in Figure~\ref{fig:teaser_fig_b}, adding explicit depth information enables DS-NeRF to capture scene structure but still struggles with details.

\noindent\textbf{Layered Depth and Multi-Plane Images:} Closely related to our work are layered depth images~\cite{meyer1998interactive,mildenhall2019local,oliveira1999image,Soft3DReconstruction,shade1998layered,zhou2018stereo} that learn an alpha composition for multi-plane images at discrete depth positions.
In this work, we did not restrict our approach to 2D planes or specific depth locations.
Instead, we learn a representation for a pixel at arbitrary depth locations.
A pixel-wise representation not only allows us to interpolate, but also to extrapolate, and obtain dense 3D reconstruction results.
Since we have a pixel-wise representation, we are able to generate $12$MP resolution images without any modifications of our approach. Prior work has demonstrated results on a maximum of $2$MP resolution content.

\noindent\textbf{4D View Synthesis:} Most approaches are restricted to 3D view synthesis~\cite{mildenhall2019local,mildenhall2020nerf} and would require drastic modifications~\cite{du2021nerflow,pumarola2020d} to be applied to the 4D view-synthesis problem.
Lombardi et al.~\cite{lombardi2021mixture} employ a mixture of animated volumetric primitives  to model the dynamic appearance of human heads from dense multi-view observations.
Open4D~\cite{Bansal_2020_CVPR} requires foreground and background modeling for 4D visualization.
Our work does not require major modifications to extend to 4D view-synthesis.
In addition, we do not require explicit foreground-background modeling for 4D view synthesis.
We demonstrate our approach on the challenging Open4D dataset~\cite{Bansal_2020_CVPR} where the minimum distance between two cameras is $50$cm. 
Our composition model trained on a single time instant also enables us to do 4D visualization for unseen time instances. Finally, the model learned for view synthesis enable dense 3D reconstruction on multi-view content. To our knowledge, no prior work has demonstrated these results for 3D-4D multi-view view synthesis. 

\section{Method}
\label{sec:method}

\begin{figure*}[t]
\includegraphics[width=\linewidth]{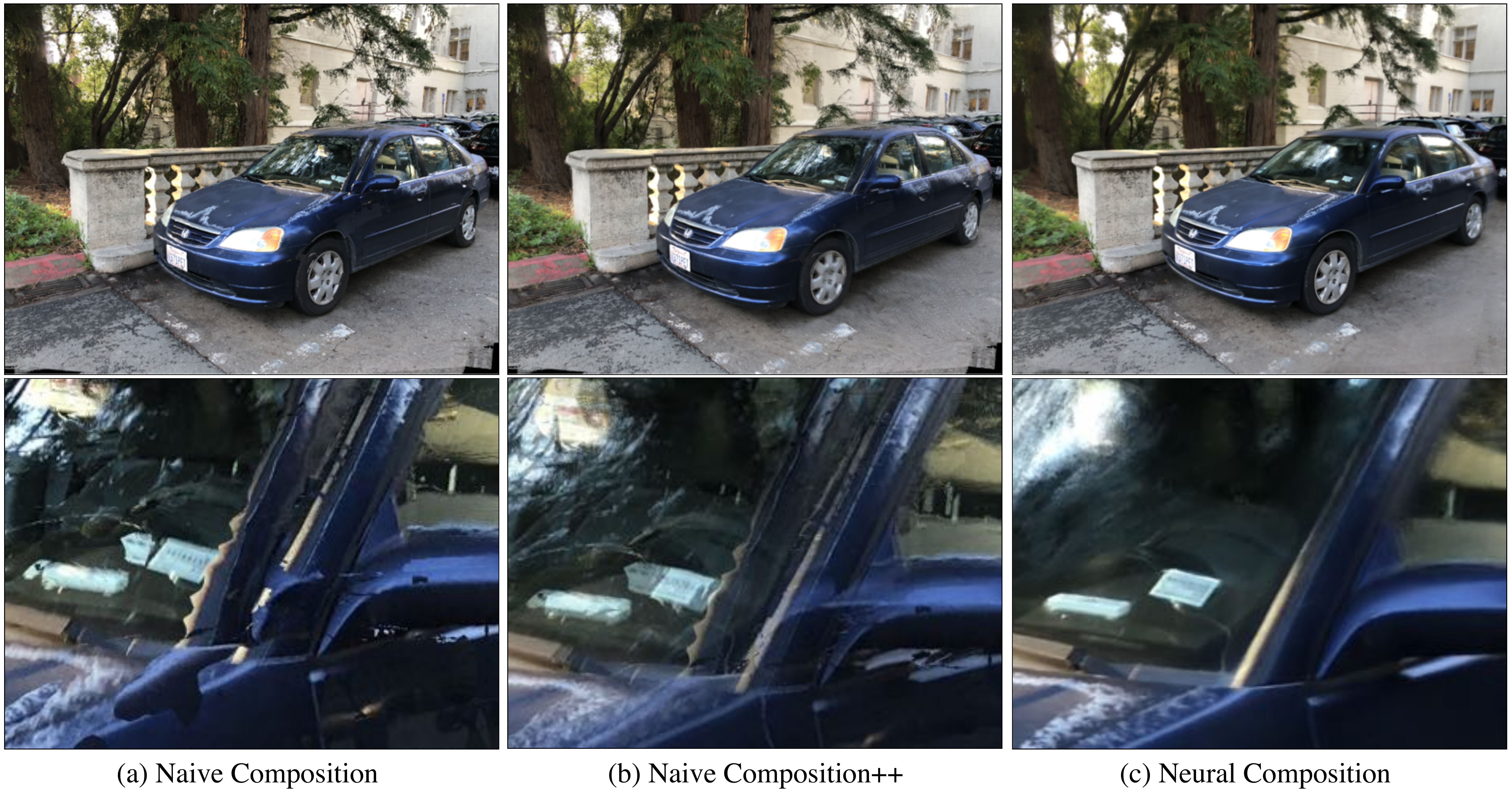}
\caption{\textbf{Naive Composition vs. Neural Composition: } The baseline naively uses multiple stereo-pairs to generate the final output ({\bf Naive Composition}). {\bf (a)} For each pixel location, we select the color value for the closest depth location. {\bf (b)} We also take the average of color values for the three closest depth locations ({\bf Naive Composition++}). {\bf (c)} We contrast these results with {\bf Neural Composition} which uses an MLP to compose the color values. We observe that the MLP nicely composes the color values despite noisy depth estimates and fills the missing regions. }
\vspace{-0.4cm}
\label{fig:composition}
\end{figure*}

We are given $M$ multi-view images with camera parameters (intrinsics and extrinsics) as input.
Our goal is to learn a function, $f$, that inputs pixel information $(\textbf{p})$, $\textbf{p} \in \mathbb{R}^{N_p}$, and outputs color $(\bm{\bar{c}} \in \mathbb{R}^{3})$ 
at that location, i.e.,  $f:\textbf{p}\rightarrow{\bm{\bar{c}}}$.
Learning such a function is challenging since we live in a 3D-4D world and images provide only 2D measurements.
We present two crucial components:
1) a representation of a \textbf{pixel} that contains relevant multi-view information for high-fidelity view synthesis;
and 2) a multi-layer perceptron (\textbf{MLP}) that inputs the pixel information and outputs the color.

\noindent\textbf{Overview: } We input a pixel location $(x,y)$ given corresponding camera parameters $(r_x, r_y, r_z, t_x, t_y, t_z)$ at time, $\tau$, along with an array of possible 3D points along the line of sight.
The $i^{th}$ location of this array contains depth ($d_i$) and color ($c_i$).
The MLP outputs alpha ($\alpha_i$) values for the $i^{th}$ location that allow us to obtain the final color at $(x,y)$. The MLP also outputs gamma, $\bm{\gamma} \in \mathbb{R}^{3}$, which is a correction term learned by the model. We get the final color at pixel location $(x,y)$ as: $\bar{c} = \gamma + \sum_{i=1}^{N} {\alpha_i}{c_i}$, where $N$ is the number of points in the array.  

We describe our representation of a pixel in Sec.~\ref{ssec:pixel} and the MLP in Sec.~\ref{ssec:mlp}.

\begin{figure*}[t]
\includegraphics[width=\linewidth]{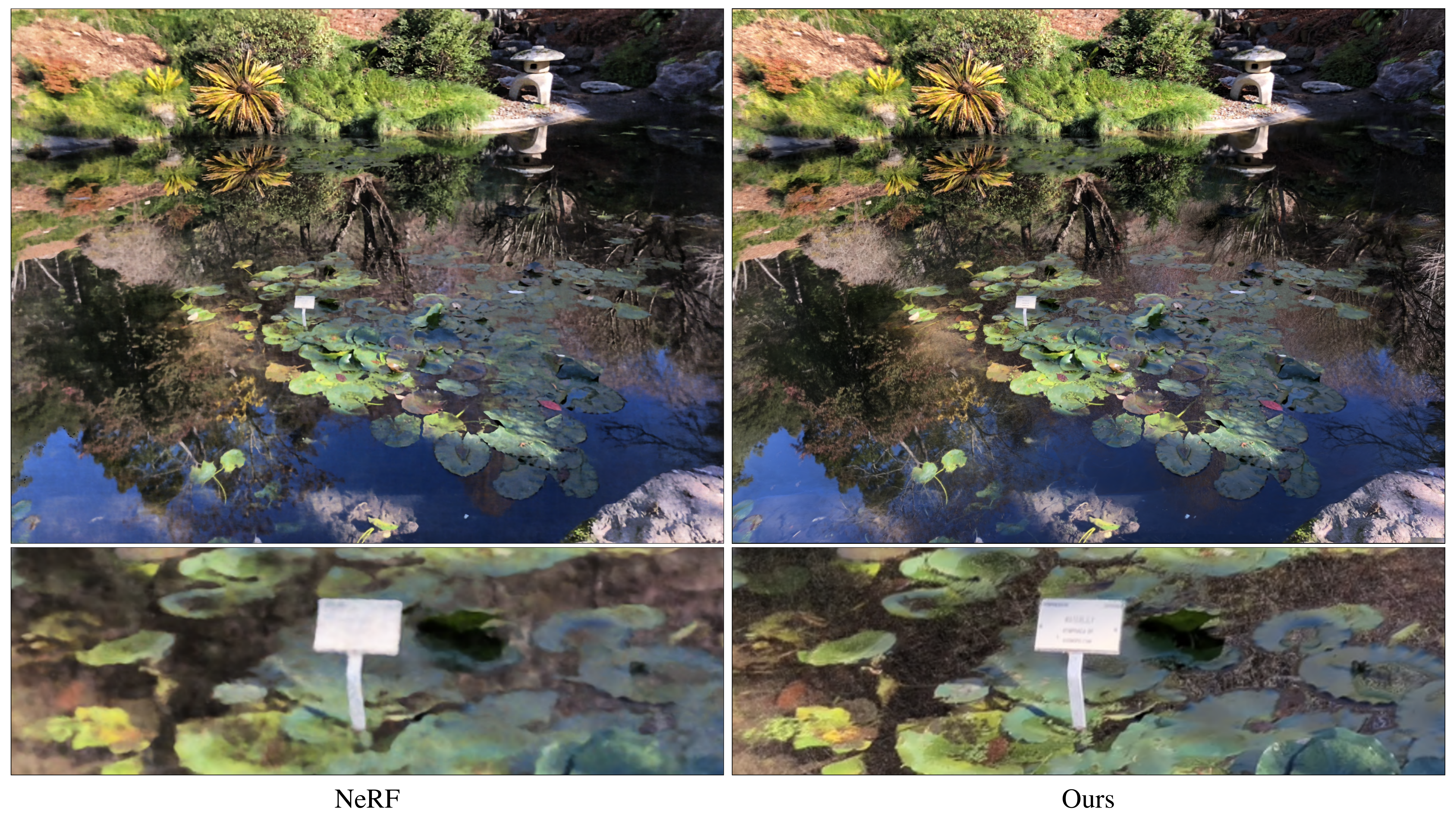}
\caption{\textbf{High-Resolution 12 MP View Synthesis:} Our approach allows us to capture small details better than \textbf{NeRF} on a novel held-out scene. We train the NeRF model for $2M$ iterations that take $64$ hours of training on a single NVIDIA V100 GPU. Ours is trained for $10$ minutes. \textcolor{blue}{Best viewed in electronic format}.}
\label{fig:scalable12MP}
\end{figure*}

\subsection{Representation of a Pixel}
\label{ssec:pixel}
Given a pixel location $(x,y)$ for a camera position $(r_x, r_y, r_z, t_x, t_y, t_z)$, our goal is to collect dense 3D information that contains depth and color information at all possible 3D points along a line of sight.
We obtain 3D points via two-view geometry~\cite{hartley2003multiple} by forming ${M \choose 2}$ stereo-pairs.
The estimated disparity between a stereo pair provides the depth for the 3D point locations.
Multiple stereo pairs allow us to densely populate 3D points along the rays.

\paragraph{Color and Depth Array:}
We use multiple stereo pairs to build an array of depth ($\textbf{d}$) and color ($\textbf{c}$) for a pixel.
We store the values in order of increasing depth, i.e., $d_{i+1} \geq d_{i}$.
The array is similar to a ray of light that travels in a particular direction connecting the 3D points.
We limit the number of 3D points to be $N$.
If there are less than $N$ depth observations, we set $d_i = 0$ and $c_i = (0,0,0)$.
If there are more than $N$ observation, we clip to the closest $N$ 3D points.

\paragraph{Uncertainty Array:} 
In this work, we use an off-the-shelf disparity estimation module from Yang et al.~\cite{Yang_2019_CVPR}.
This approach provides an estimate of uncertainty (entropy) for each prediction.
We also keep an array of uncertainty values ({$\bm{\mathfrak{H}}$}) of equal size as the depth array (obtained from disparity and camera parameters), s.t., $\mathfrak{H}_i\in[0,1]$, where a higher value represents higher uncertainty.
The uncertainty allow us to suppress noise or uncertain 3D points. 

\paragraph{Encoding Spatial Information:}
For each pixel, we concatenate its spatial location, i.e., $(x,y)$ location and camera position $(r_x, r_y, r_z, t_x, t_y, t_z)$.
We employ high-frequency positional encoding~\cite{mildenhall2019local} to represent spatial information of a pixel for a given camera position.
We normalize the pixel coordinates, s.t., $x\in[-1,1]$ and $y\in[-1,1]$.

\paragraph{Incorporating Temporal Information:}
Our approach enables a natural extension to incorporate temporal information.
Given a temporal sequence with $T$ frames, we represent each time instant as a Gaussian distribution with peak at the frame $\tau$.
We concatenate the color, depth, and uncertainty array alongside the spatial and temporal information in a single $N_p$-dimensional array, where $N_p$ is sum of the dimensions of each term.  
We input this array to the MLP to compute the color output at the pixel location.

\subsection{Neural Composition via Multi-Layer Perceptron (MLP)}
\label{ssec:mlp}
Our goal is to output blending values $\alpha$ that enable us to take the appropriate linear combination of color values in the color-array.
A naive way is to directly use the output of the last layer of the MLP as an alpha array and compute a dot product with the color-array:
\begin{align}
f(x, y, \tau, r_x, r_y, r_z, t_x, t_y, t_z, \bm{c}, \bm{d}, \bm{\mathfrak{H}}) = \bf{w}.
\end{align}
While this is reasonable, it assumes that the MLP will implicitly understand the relationship between color ($\bm{c}$), depth ($\bm{d}$), and uncertainty ($\bm{\mathfrak{H}}$).
This is challenging to learn. 
In this work, we observe that explicitly using the depth and uncertainty with the output of the MLP ($\bf{w}$) enables better view synthesis.
We, therefore, define:
\begin{align}
\alpha_i = \frac{(1-\mathfrak{H}_{i})e^{-(w_{i}d_{i} - \mu)^2}}{\sum_{j=1}^{N} (1-\mathfrak{H}_{j})e^{-(w_{j}d_{j} - \mu)^2}}, 
\end{align} where $\mu = \frac{1}{N}\sum_{j=1}^{N}{w_j}{d_j}$.
The Gaussian distribution forces the model to select color values belonging to depth location that are: 
1)  closest to the average depth value;
and 2) are confident and less noisy.
We employ these alpha values together with the original color array to predict the final values ($\bm{\bar{c}}$):
\begin{align}
    \bar{c} = \sum_{i=1}^{N} {\alpha_i}{c_i} + \gamma,
\end{align}
where $\gamma$ is an additional correction term that helps us to obtain sharp outputs. Note that the fifth layer of the MLP outputs $\alpha$ and $\gamma$ values.

\paragraph{Multi-Layer Perceptron:}
We employ a $5$-layer perceptron.
Each linear function has $256$ activations followed by a non-linear ReLU activation function.
We train the MLP in a self-supervised manner using a photometric $\ell_1$-loss:
\begin{align}
    \min_{f} \mathcal{L} =  \sum_{k=1}^{m} \big|\big| c_k - \bar{c}_k \big|\big|_{1},
\end{align}
where $c_k$ and $\bar{c}_k$ are the ground truth color and predicted color respectively for the $k^{th}$ pixel, and $m$ is the number of randomly sampled pixels from the $M$ images.
We train the MLP from scratch using the Adam optimizer~\cite{kingma2014adam}.
We randomly sample $4$ images, and sample $256$ pixels from each image for every forward/backward pass.
The learning rate is kept constant at $0.0002$ for the first $5$ epochs and is then linearly decayed to zero over next $5$ epochs. We observe that composition model converges around in a few seconds of training on a single GPU with $1$ GB GPU memory. Figure~\ref{fig:scalable12MP} contrasts our results with NeRF on one such dense multi-view sequence.

\noindent\textbf{Naive Composition}: One can also naively use the pixel representation to generate the final output by selecting the color value for the closest depth location. A slightly nuanced version is to take average of color values for three closest depth location ({\bf Naive Composition++}). We use this naive composition for comparisons in our work. Figure~\ref{fig:composition} shows the importance of using neural composition via MLP over naive composition. We believe it is an importance baseline for view synthesis as this simple nearest-neighbor based method generates results without any training.

\section{3D Multi-View View Synthesis}
\label{sec:exp}

We study various aspects of 3D view synthesis using our approach: {\bf (1)} synthesizing novel views given sparse and unconstrained multi-views (Sec.~\ref{ssec:sparse3d}); {\bf (2)} synthesizing hi-res $12$MP content (Sec.~\ref{ssec:hi-res}); and {\bf (3)} scenes with unbounded depth and influence of the number of views (Sec.~\ref{ssec:number}). We then demonstrate our approach on hi-resolution studio capture in Sec.~\ref{ssec:studio} and show that our approach can generalize to unseen subjects and unknown time instances from a single time-instant. We study convergence in Sec.~\ref{ssec:conv} where we observe that our model gets close to convergence within a few seconds of learning. Finally, there are more analysis in Appendix~\ref{ss:more}.

\subsection{Sparse and Unconstrained Multi-Views} 
\label{ssec:sparse3d}

We use $24$ sequences of sparse and unconstrained real-world samples from the Open4D dataset~\cite{Bansal_2020_CVPR}. Open4D consists of temporal sequences. We use certain time instants for 3D view synthesis. The minimum distance between two adjacent cameras is $50$cm in these sequences. We contrast our approach with NeRF~\cite{mildenhall2020nerf}. We also study DS-NeRF~\cite{deng2021depth}, which additionally employs sparse 3D point clouds from COLMAP for training NeRF. DS-NeRF has shown promising results given the sparse views from LLFF dataset~\cite{mildenhall2019local}. Both approaches are trained for $200,000$ iterations (roughly $420$ minutes) per sequence on a NVIDIA V100 GPU.  Table~\ref{tab:3d-sparse} compares the performance of different methods on held-out views from these sequences using PSNR, SSIM, and LPIPS (AlexNet)~\cite{zhang2018perceptual}. In this work, we observe that these three evaluation criteria are not self-sufficient in determining the relative ranking of different methods. While PSNR and SSIM may favor smooth or blurry results~\cite{zhang2018perceptual}, LPIPS may ignore the structural consistency in images. We posit that it is important to look at all three criteria and not one.  
Figure~\ref{fig:teaser_fig_b} shows the qualitative performance of our approach on these challenging sequences. We observe degenerate results using NeRF on these sequences. DS-NeRF also results in degenerate outputs most of the times except for the scenes with bounded depth. Our approach is able to generate high-quality results (with details such as faces, hair, dress, etc.) in this setting both qualitatively and quantitatively.  Total time taken to process (pre-processing multi-view content and training the composition model) a sequence is less than $10$ minutes. We also observe that a naive pixel composition can also yield meaningful results better than prior work. However, we obtain better pixel composition using MLPs. The details of these sequences are available in Appendix~\ref{app:sparse3d}. 

\begin{table}
\scriptsize{
\setlength{\tabcolsep}{6pt}
\def\arraystretch{1.2}
\center
\begin{tabular}{@{}l c c  c c c}
\toprule
\textbf{24 sequences} &  & \textbf{PSNR}$\uparrow$ & \textbf{SSIM}$\uparrow$   & \textbf{LPIPS}~\cite{zhang2018perceptual} $\downarrow$ \\ 
\midrule
LLFF~\cite{mildenhall2019local} &   &  15.187 $\pm$ 2.166 & 0.384 $\pm$ 0.082  & 0.602 $\pm$ 0.090   \\
NeRF~\cite{mildenhall2020nerf} & & 13.693 $\pm$ 2.050 &  0.317 $\pm$ 0.094  & 0.713 $\pm$ 0.089 \\ 
DS-NeRF~\cite{deng2021depth}   &. &  14.531 $\pm$ 2.603  & 0.316 $\pm$ 0.099  & 0.757 $\pm$ 0.040 \\ 
DS-NeRF{\bf **}~\cite{deng2021depth}   &. &  15.346 $\pm$ 2.276  & 0.389 $\pm$ 0.076  & 0.716 $\pm$ 0.048 \\ 
 \midrule
Naive Composition & & 15.480 $\pm$ 1.928  & 0.372 $\pm$ 0.061  & 0.665 $\pm$ 0.065 \\ 
Naive Composition++ &   & 16.244 $\pm$ 2.186  & 0.442 $\pm$ 0.074  & 0.616 $\pm$ 0.063 \\ 
 \midrule
Ours                          &   &  {\bf 17.946 $\pm$ 1.471}  & {\bf 0.562 $\pm$ 0.077} & {\bf 0.534 $\pm$ 0.061} \\
\bottomrule
\end{tabular}
\caption{\textbf{Sparse and Unconstrained Multi-Views:} We evaluate on the $24$ sparse and unconstrained multi-view sequences of the Open4D dataset~\cite{Bansal_2020_CVPR}. We train NeRF~\cite{mildenhall2020nerf} and DS-NeRF~\cite{deng2021depth} models for each sequence. DS-NeRF~\cite{deng2021depth} employs additional depth along with NeRF. We trained two versions of DS-NeRF. One where we use the same model as NeRF with additional depth supervision. The second version is DS-NeRF{\bf **} with tuned hyperparameters. We also use the off-the-shelf LLFF model. We observe degenerate outputs using LLFF, NeRF and DS-NeRF, especially for unbounded scenes. However, our approach is able to reliably generate novel views in twenty times less time. Training a NeRF/DS-NeRF model takes roughly $420$ minutes per sequence whereas our approach take 10 minutes (including pre-processing multi-view content). We also generate results using naive composition and obtain better results than prior work. We observe that the MLP allows us to do better composition than naive composition.}
\label{tab:3d-sparse}
}
\vspace{-1.cm}
\end{table}

\subsection{High-Resolution (12MP) View Synthesis} 
\label{ssec:hi-res}

\begin{figure}
\includegraphics[width=\linewidth]{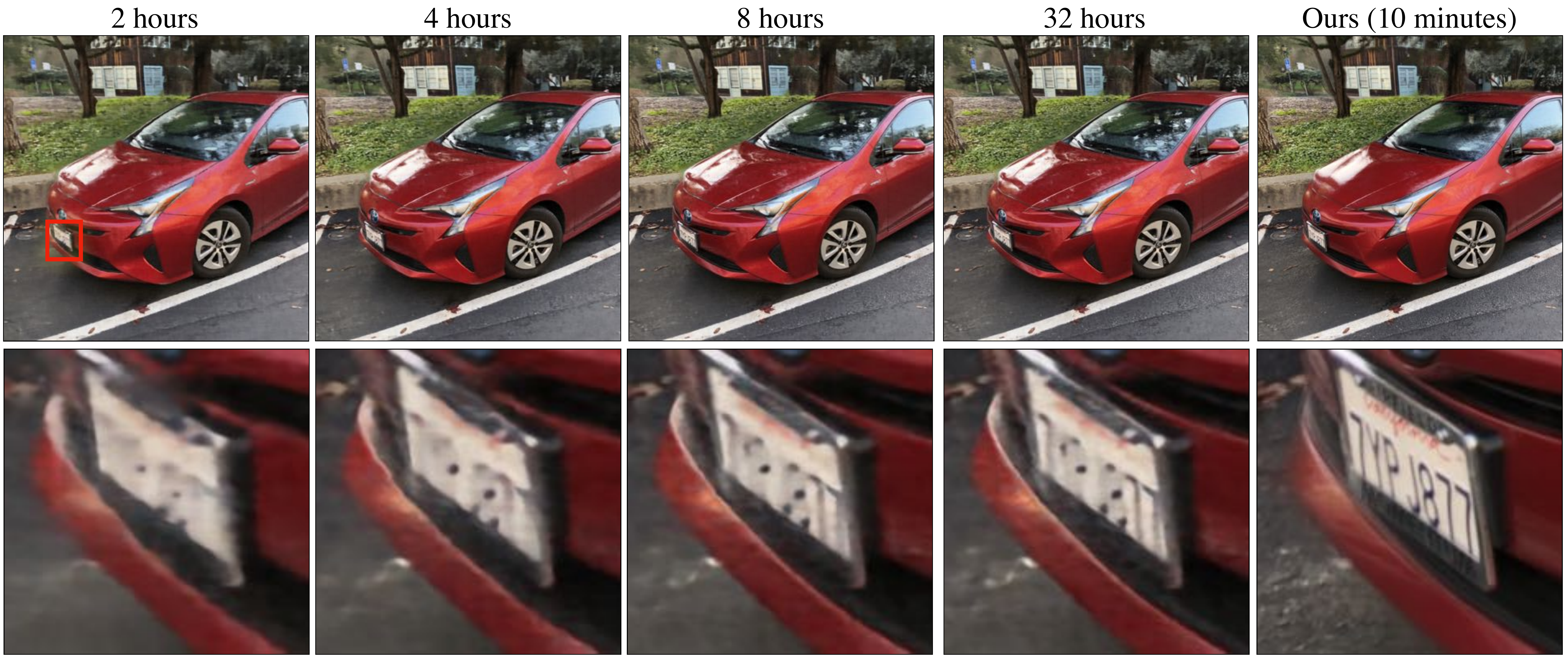}
\includegraphics[width=\linewidth]{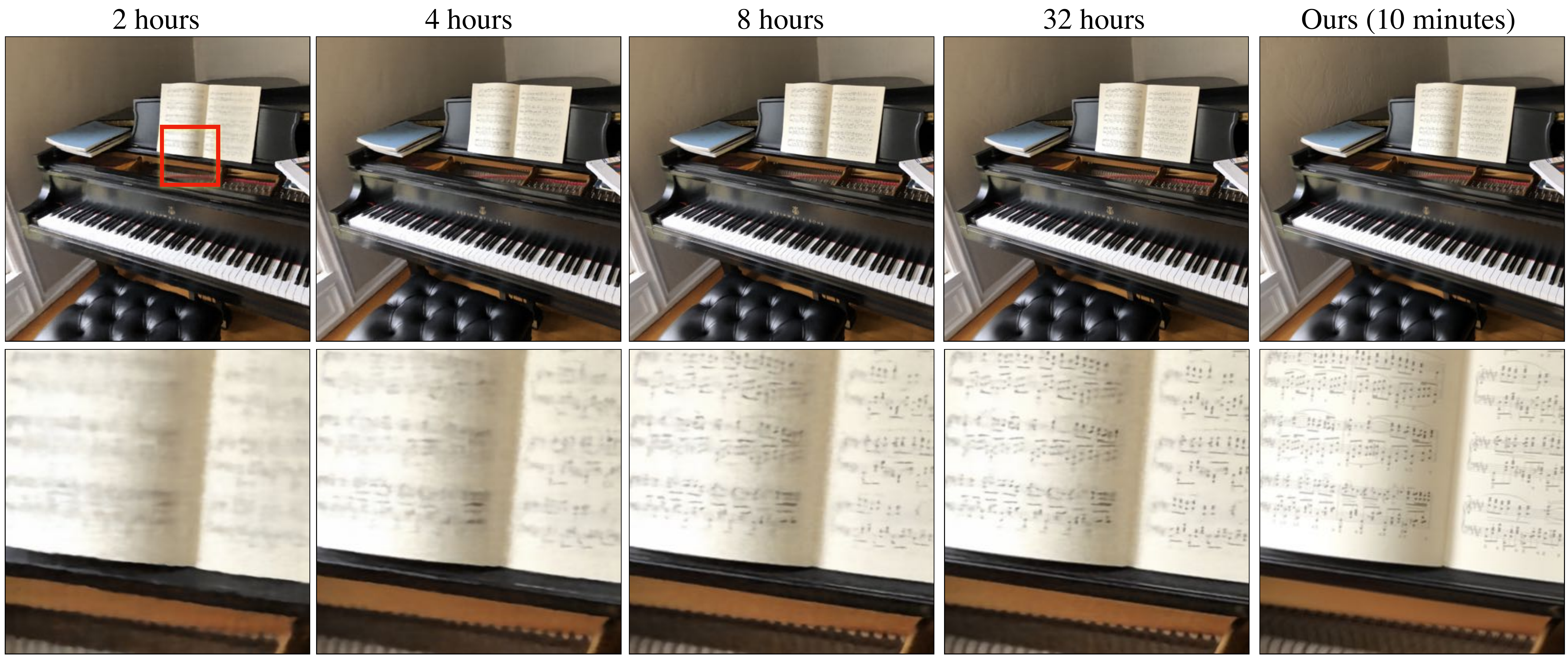}
\caption{\textbf{Improvement in NeRF over time: } We show the progression (first $32$ hours) of improvement for the NeRF model. We observe that results improve over time as details become clearer over time. We contrast this with our approach that can generate sharp results in only $10$ minutes. \textcolor{blue}{Best viewed in electronic format}. }
\label{fig:time}
\end{figure}

We use twelve high-resolution ($4032\times3024$) multi-view sequences from the LLFF dataset~\cite{mildenhall2019local} that contain challenging specular surfaces. In this setting, we train NeRF~\cite{mildenhall2020nerf} on these sequences for $2, 000,000$ iterations which take approximately $64$ hours on a single NVIDIA V100 GPU ($10,000$ iterations take $20$ minutes). Performance saturates at $1M$ iterations after 32 hours of training. We also show the performance for {\tt vanilla} NeRF that is trained for $200,000$ iterations and takes $400-420$ minutes to train. We train our model for $10$ epochs, which takes around $10$ minutes on a single GPU and only $1$GB GPU of memory. We estimate disparity~\cite{Yang_2019_CVPR} for multiple stereo pairs at one-fourth resolution for these sequences. Disparity estimation using the off-the-shelf model takes less than $5$ minutes per sequence on a single GPU. Table~\ref{tab:3d-scalable} contrasts the performance of NeRF models at different intervals of training using PSNR, SSIM, and LPIPS (AlexNet). We compute the average of per-frame statistics as the number of samples in the test set for these $12$ sequences are roughly the same. We once again observe that it is crucial to include all three evaluation criteria. Figure~\ref{fig:time} shows the results of NeRF at different intervals of time. We observe that the NeRF model improves over time and captures sharp results as suggested by LPIPS. Our method enables sharper outputs as compared to NeRF. Interestingly, NeRF does not capture details even for training samples when trained sufficiently long (64 hours) which suggests that it is non-trivial to capture details using NeRF on held-out samples. The qualitative and quantitative analysis suggest that we can efficiently generate results on 12MP images without drastically increasing the computational resources.  We also show the performance of naive composition to generate the final outputs. We observe that MLPs allow us to obtain better results. The details of these sequences are available in Appendix~\ref{app:hi-res}.

\begin{table}
\scriptsize{
\setlength{\tabcolsep}{6pt}
\def\arraystretch{1.2}
\center
\begin{tabular}{@{}l c c  c c c}
\toprule
\textbf{12 sequences} & & \textbf{PSNR}$\uparrow$ & \textbf{SSIM}$\uparrow$   & \textbf{LPIPS} $\downarrow$ \\ 
\midrule
\textbf{NeRF}~\cite{mildenhall2020nerf} &   &  &  &   \\
2 hours &   & 21.151 $\pm$ 2.783 & 0.577 $\pm$ 0.157   & 0.662 $\pm$ 0.099 \\ 
4 hours &   & 21.469 $\pm$ 2.881 & 0.588 $\pm$ 0.153  & 0.628 $\pm$ 0.096  \\ 
{\tt vanilla} &   &  21.625 $\pm$ 2.933 &  0.596 $\pm$ 0.150  &  0.605 $\pm$ 0.092 \\
8 hours &   & 21.674 $\pm$  2.958 & 0.598 $\pm$ 0.149   & 0.599 $\pm$ 0.091 \\ 
16 hours &   & 21.734 $\pm$ 2.981 & 0.602 $\pm$ 0.148   & 0.586 $\pm$ 0.088 \\ 
32 hours &   & 21.741 $\pm$ 2.985 & 0.602  $\pm$ 0.147   & 0.584 $\pm$ 0.087\\ 
64 hours   &.  & {\bf 21.741  $\pm$ 2.985}  & {\bf 0.602  $\pm$ 0.147}  & 0.584 $\pm$ 0.087 \\ 
\midrule
Naive Composition                        &    &  16.008 $\pm$ 2.315  & 0.415  $\pm$ 0.142  &  0.427 $\pm$ 0.068 \\ 
Naive Composition++                          &    &   17.022 $\pm$  2.483 &  0.460 $\pm$  0.144 & {\bf 0.406 $\pm$ 0.066}\\
\midrule
\textbf{Ours} (10 minutes) &   &  &    & \\
$K=50, N=50$                     &   &  20.834 $\pm$ 2.784  & 0.594 $\pm$  0.136    & 0.426 $\pm$ 0.075 \\
\boxit{4.15in} $K=100, N=100$                    &   & 20.953 $\pm$ 2.805 & 0.598 $\pm$ 0.136    &  0.460 $\pm$ 0.078\\
$K=200, N=200$                   &   & 20.783 $\pm$ 2.749  &  0.593 $\pm$ 0.135  & 0.494 $\pm$ 0.081  \\
$K=ALL, N=200${\bf *}                  &   & 20.712 $\pm$ 2.656  & 0.591 $\pm$ 0.134    &  0.497 $\pm$ 0.077 \\
\midrule
\textbf{Ours} (50 minutes) &   &  &    & \\
$K=50, N=50$                     &   & 20.777 $\pm$ 2.809  & 0.591 $\pm$ 0.137  & 0.416 $\pm$ 0.075  \\
$K=100, N=100$                      &    & 21.006 $\pm$ 2.869  & 0.597  $\pm$ 0.136 &  0.448 $\pm$ 0.080 \\ 
$K=200, N=200$                     &   & 20.924 $\pm$ 2.847  & 0.592 $\pm$ 0.134   & 0.477 $\pm$ 0.082  \\
$K=ALL, N=200${\bf *}                   &   & 20.825 $\pm$ 2.708  & 0.589 $\pm$ 0.132    & 0.480 $\pm$ 0.079 \\
\midrule
\textbf{Ours} (250 minutes) &   &  &     \\
$K=50, N=50$                      &    & 20.582  $\pm$ 2.751  & 0.585 $\pm$ 0.135 &  0.409 $\pm$ 0.076 \\ 
$K=100, N=100$                      &   & 20.916 $\pm$ 2.874   & 0.593 $\pm$ 0.135  &  0.433 $\pm$ 0.078  \\
$K=200, N=200$                     &   & 20.640 $\pm$ 3.234  & 0.582 $\pm$ 0.139    &  0.474 $\pm$ 0.095\\
$K=ALL, N=200${\bf *}                   &   & 20.548 $\pm$ 3.104 & 0.580 $\pm$ 0.137     & 0.478 $\pm$ 0.092\\
\bottomrule
\end{tabular}
\caption{\textbf{Hi-Res (12MP) View Synthesis:} We evaluate on $12$ sequences from LLFF containing specular surfaces on original $4032\times3024$ resolution. The details of these sequences are available in Appendix~\ref{app:hi-res}. We contrast the performance of our approach with different intervals of training a NeRF model. Performance saturates at $1M$ iterations after 32 hours of training. Our composition model converges quickly in a few minutes. Here, we show the results of our composition model trained for $10$ epochs that takes around $10$ minutes, $50$ epochs that takes less than $1$ hour. Training our model require 1 GB of GPU memory for training. We also show the results when the model is trained for $250$ epochs. For each setting, we vary the number of stereo pairs ($K$) and number of 3D points ($N$). We observe that using a few stereo-pairs gives competitive and better results than using all the pairs. We posit that noise introduced by using more stereo pairs might be responsible for the lower performance. Finally, we study the benefit of using an MLP for composing per-pixel color and depth information. The MLP allows us to obtain better results than a naive composition (Fig.~\ref{fig:composition}). We refer the reader to Figure~\ref{fig:time} for visual comparisons. We observe that details become better for NeRF when trained for long. However, our approach captures more details in a few minutes as compared to $32$ hours of training of a NeRF model. Consistent with the observation of Zhang et al.~\cite{zhang2018perceptual}, PSNR may favor averaged/blurry results while LPIPS favors sharp results. }
\label{tab:3d-scalable}
}
\vspace{-1.cm}
\end{table}

We also vary the number of stereo pairs ($K$) to synthesize the target view. We observe that we can get better results with a few stereo pairs than using all pairs. Synthesizing a new view for a dense multi-view sequence can be achieved by looking at the local neighborhood of the target location instead of using all the views. Local neighborhood is determined based on position in world space, i.e., we use stereo-pairs corresponding to the closest camera and then next and so on, unless we have $K$ samples. This allows us to speed-up training and testing.

\begin{figure}
\includegraphics[width=\linewidth]{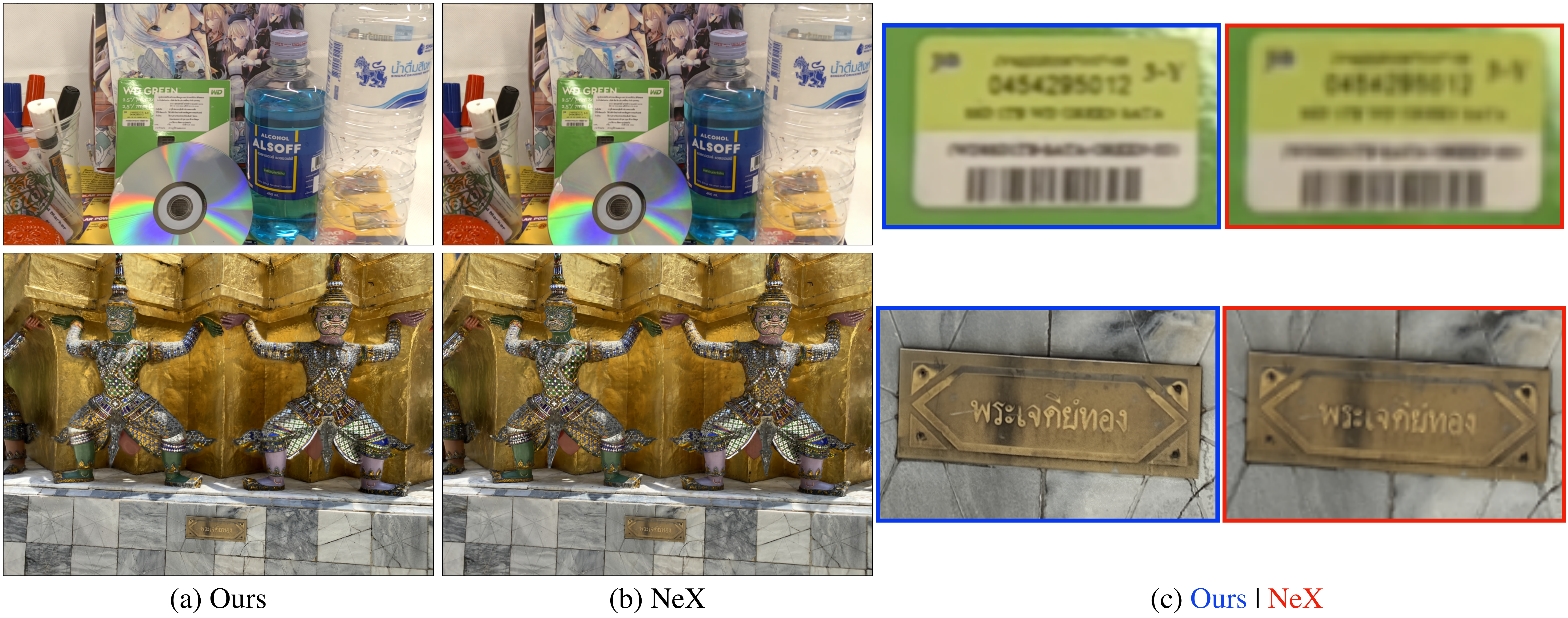}
\caption{\textbf{Shiny Dataset: } {\bf (a)}  We contrast the results of our approach with {\bf (b)} NeX~\cite{Wizadwongsa2021NeX}  on held-out views. {\bf (c)} Our approach is able to capture the details better than NeX such as the text ({\tt 0454295012 3-Y}) in the top-row and the details on the plate and stone in the bottom row. \textcolor{blue}{Best viewed in electronic format}.}
\label{fig:shiny}
\end{figure}

\noindent\textbf{Shiny Dataset: } We use $8$ multi-view sequences from the Shiny Dataset~\cite{Wizadwongsa2021NeX} that consists of multi-views captured for specular surfaces. The resolution of $6$ sequences (less than $60$ samples in each) in this dataset is $4032\times3024$, and the remaining two ({\tt cd} and {\tt labs} have more than $300$ samples) have resolution $1920\times1080$. We train NeRF on the original resolution of these sequences for $2M$ iterations (64 hours per GPU). We contrast the performance with our approach that is trained for 10 epochs and 50 epochs. Table~\ref{tab:shiny} shows the performance of different methods. We follow the evaluation criteria (average of per-sequence PSNR, multi-channel SSIM, LPIPS\footnote{We, however, use LPIPS via AlexNet (alex) instead of VGG-Net (vgg) to fit 12MP images on a single GPU.}) from NeX~\cite{Wizadwongsa2021NeX}. We also add the results generated by NeX~\cite{Wizadwongsa2021NeX} that synthesizes on one-fourth resolution for these sequences. We do a simple $4\times$-upsampling of their results to target resolution for an apples-to-apples comparison. Our model trained for 10 minutes achieves results close to the best performance. Figure~\ref{fig:shiny} contrasts our method with NeX. We observe small details are better captured by our method.

\begin{table}
\scriptsize{
\setlength{\tabcolsep}{6pt}
\def\arraystretch{1.2}
\center
\begin{tabular}{@{}l c c  c c}
\toprule
\textbf{8 sequences} & & \textbf{PSNR}$\uparrow$ & MC\textbf{SSIM}$\uparrow$   & \textbf{LPIPS} $\downarrow$\\
\midrule
{\tt 4032$\times$3024} &    &  &  &   \\
\textbf{NeRF}          &   &   &  &     \\
{\tt vanilla}         &   & 21.141 $\pm$ 3.528  & 0.735 $\pm$ 0.155  & 0.528 $\pm$ 0.157     \\
2M iterations         &   & 21.457 $\pm$ 3.657  & 0.751 $\pm$ 0.155  &   0.498 $\pm$ 0.153  \\
\midrule
Naive Composition &   & 16.624 $\pm$ 2.906 & 0.648 $\pm$ 0.197 &   0.342 $\pm$ 0.096 \\
Naive Composition++ &   & 17.535 $\pm$ 2.698  & 0.688 $\pm$ 0.184  &  0.317 $\pm$ 0.107 \\
 \midrule
\textbf{Ours} (10 minutes) &   &  &    & \\
$K=50, N=50$                     &   & 22.430 $\pm$ 4.748  & 0.795 $\pm$ 0.142   & 0.256 $\pm$ 0.108 \\
\boxit{4.55in}$K=100, N=100$                    &   & 22.868 $\pm$ 4.588  & 0.802 $\pm$ 0.140    & 0.269 $\pm$ 0.120\\
$K=200, N=200$                   &   &  23.016 $\pm$ 4.698 & 0.803 $\pm$ 0.144    & 0.285 $\pm$ 0.132\\
$K=ALL, N=200${\bf *}                   &   & 22.090 $\pm$ 4.263 & 0.786 $\pm$ 0.154   &  0.332 $\pm$ 0.145 \\
\midrule
\textbf{Ours} (50 minutes) &   &  &  &   \\
$K=50, N=50$                     &  & 22.261 $\pm$ 4.812 & 0.791 $\pm$ 0.144  &  {\bf 0.252  $\pm$  0.105} \\
$K=100, N=100$                     &   & 22.739 $\pm$ 4.637 & 0.801 $\pm$ 0.142 &  0.258 $\pm$ 0.113   \\
$K=200, N=200$                     &  & {\bf 23.020 $\pm$ 4.690} & {\bf 0.805 $\pm$ 0.143} &   0.271 $\pm$ 0.123  \\
$K=ALL, N=200${\bf *}                   &   & 21.788 $\pm$ 4.243  & 0.780 $\pm$ 0.154   &  0.317 $\pm$ 0.137 \\
\midrule
{\tt resized to original resolution} &    &  &  &   \\
NeRF~\cite{mildenhall2020nerf} &    & 22.009 $\pm$ 3.148  & 0.757 $\pm$ 0.156  & 0.487 $\pm$ 0.180 \\
NeX~\cite{Wizadwongsa2021NeX} &    & 22.292 $\pm$ 3.137  & 0.774 $\pm$ 0.152  & 0.423  $\pm$ 0.156\\
\bottomrule
\end{tabular}
\caption{\textbf{Shiny dataset:} We study our approach on the $8$ real sequences from the Shiny dataset~\cite{Wizadwongsa2021NeX}. NeRF is trained for $2M$ iterations taking approx $64$ hours. We also add the results of $4\times$ bi-linearly upsampled results from  NeX~\cite{Wizadwongsa2021NeX} on these sequences. Our approach gets competitive performance in only a few minutes. }
\label{tab:shiny}
}
\vspace{-0.8cm}
\end{table}

\noindent\textbf{Standard LLFF Sequences: } We quantitatively evaluate our approach on 8 forward-facing real-world multi-view sequences~\cite{mildenhall2020nerf} in Table~\ref{tab:llff}. We use the original hi-res ($4032\times3024$) undistorted images provided by Wizadwongsa et al.~\cite{Wizadwongsa2021NeX}. We once again train NeRF models for these hi-res sequences for $2M$ iterations (64 hours per GPU). Training the model for long allows us to get better performing NeRF models for these sequences. We follow the evaluation criteria (average of per-sequence PSNR, multi-channel SSIM, LPIPS) from NeX~\cite{Wizadwongsa2021NeX}. We also add the results reported by NeX~\cite{Wizadwongsa2021NeX}. These results were generated on one-fourth resolution. We upsample them to the desired resolution. We report the performance of our approach (without any modification for these sequences) trained for 10 and 50 epochs. Our approach underperform both PSNR and SSIM but achieves a competitive LPIPS score. However, we can generate novel hi-res views ($12$MP) in a few minutes with limited computational resources. 

\begin{table}
\scriptsize{
\setlength{\tabcolsep}{6pt}
\def\arraystretch{1.2}
\center
\begin{tabular}{@{}l c c  c c}
\toprule
\textbf{8 sequences} & & \textbf{PSNR}$\uparrow$ & MC\textbf{SSIM}$\uparrow$   & \textbf{LPIPS} $\downarrow$\\
\midrule
{\tt 4032$\times$3024} &    &  &  &   \\
\textbf{NeRF}        &   &   &  &     \\
{\tt vanilla}         &   & 25.192 $\pm$ 3.681  & 0.881 $\pm$ 0.063  &  0.396 $\pm$ 0.084   \\
2M iterations         &   & {\bf 25.666 $\pm$ 3.833}  & {\bf 0.887 $\pm$ 0.062} &   0.372 $\pm$ 0.080  \\
\midrule
Naive Composition &   & 17.147 $\pm$ 2.878  & 0.687 $\pm$ 0.134  &  0.528 $\pm$ 0.116 \\
Naive Composition++ &   & 18.280 $\pm$ 2.852  & 0.732 $\pm$ 0.124  & 0.475 $\pm$ 0.118   \\
 \midrule
\textbf{Ours} (10 minutes) &   &  &    & \\
$K=50, N=50$                     &   & 22.561 $\pm$  3.361 &  0.848 $\pm$ 0.078   &  {\bf 0.347 $\pm$ 0.085} \\
\boxit{4.55in}$K=100, N=100$                    &   & 22.951 $\pm$ 3.564 & 0.854 $\pm$ 0.077   & 0.361 $\pm$ 0.087 \\
$K=200, N=200$                   &   & 22.930 $\pm$  3.612  & 0.854 $\pm$ 0.078    & 0.380 $\pm$  0.096\\
$K=ALL, N=200${\bf *}                   &   & 21.650 $\pm$ 2.605  &  0.841 $\pm$ 0.072   & 0.416 $\pm$ 0.079  \\
\midrule
\textbf{Ours} (50 minutes) &   &  &  &   \\
$K=50, N=50$                    & & 22.335 $\pm$ 3.316  &  0.839 $\pm$ 0.086 &    0.355 $\pm$ 0.099 \\
$K=100, N=100$                   &  & 23.020 $\pm$ 3.500   & 0.851 $\pm$ 0.079  & 0.356 $\pm$ 0.093    \\
$K=200, N=200$                   &  &  23.237 $\pm$ 3.673  & 0.853 $\pm$ 0.082  &  0.369 $\pm$ 0.105   \\
$K=ALL, N=200${\bf *}                   &   & 21.650 $\pm$ 2.605 & 0.841 $\pm$ 0.072    & 0.400 $\pm$ 0.090 \\
\midrule
{\tt resized to original resolution} &    &  &  &   \\
SRN~\cite{sitzmann2019scene} &  & 21.147 $\pm$ 3.140   & 0.821 $\pm$ 0.078 & 0.594 $\pm$ 0.113    \\
LLFF~\cite{mildenhall2019local} &  &  23.334 $\pm$ 3.315 & 0.863 $\pm$ 0.064 &  0.431 $\pm$ 0.091   \\
NeRF &  & 25.076 $\pm$ 3.432   & 0.871 $\pm$ 0.062  &   0.439 $\pm$ 0.103  \\
NeX &  & 25.430 $\pm$ 3.503  & 0.881 $\pm$ 0.058 &  0.387 $\pm$ 0.077   \\
\bottomrule
\end{tabular}
\caption{\textbf{Real forward-facing dataset:} We study our approach on the original resolution of the $8$ real sequences from Mildenhall et al.~\cite{mildenhall2019local}. We also add the results of $4\times$ bi-linearly upsampled results from  NeX~\cite{Wizadwongsa2021NeX} on these sequences. Our approach underperform PSNR and SSIM but competitive LPIPS score. }
\label{tab:llff}
}
\vspace{-.75cm}
\end{table}

\begin{figure}
\includegraphics[width=\linewidth]{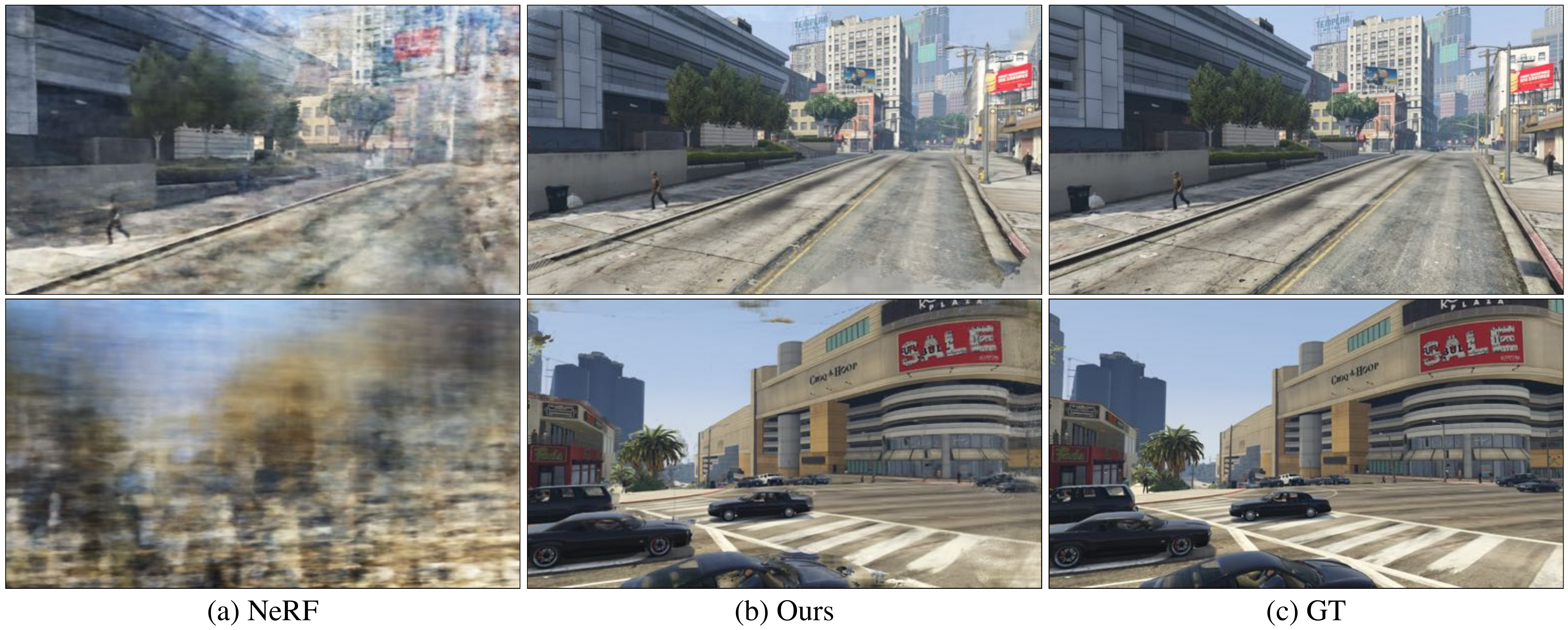}
\caption{\textbf{Synthetic scenes with unbounded depth, Num Views = $10$ :} {\bf (a)}  We show held-out views synthesized using NeRF models trained for synthetic scenes with unbounded depth. In the this experiment, the number of views at training time is $10$. {\bf (b)} We then show results using our approach. Our approach synthesizes detailed novel views despite lower number of views available for training. {\bf (c)} The ground truth is shown for reference.}
\label{fig:num-views-10}
\end{figure}

\subsection{Unbounded Scenes and Varying Number of Views} 
\label{ssec:number}

\begin{figure}
\includegraphics[width=\linewidth]{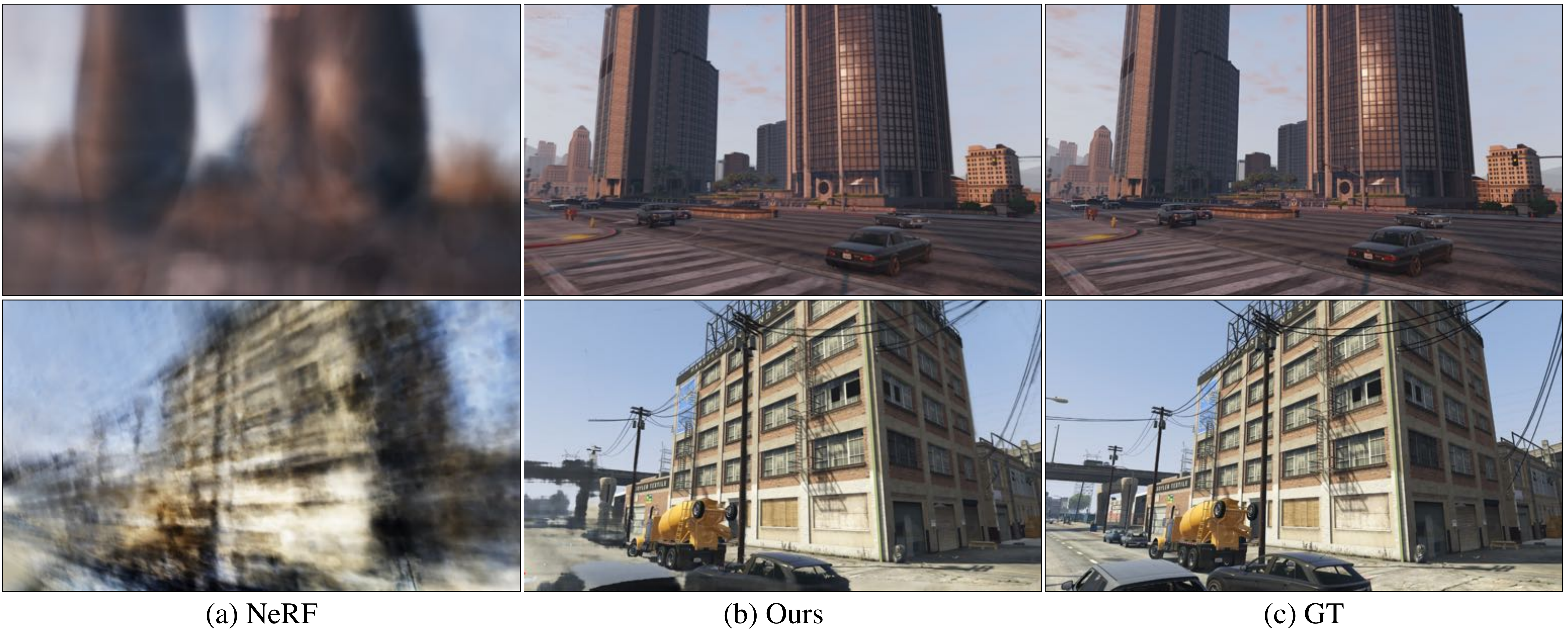}
\caption{\textbf{Synthetic scenes with unbounded depth, Num Views = $50$:} {\bf (a)}  We show held-out views synthesized using NeRF models trained for synthetic scenes with unbounded depth. In this experiment, the number of views at training time is $50$. {\bf (b)} We then show results using our approach. While NeRF struggles for scenes with unbounded depth, our approach is able to synthesize high-quality and detailed novel views. {\bf (c)} The ground truth is shown for reference. }
\label{fig:num-views-50}
\vspace{-0.7cm}
\end{figure}

\begin{figure*}[t]
\centering
\includegraphics[width=0.95\linewidth]{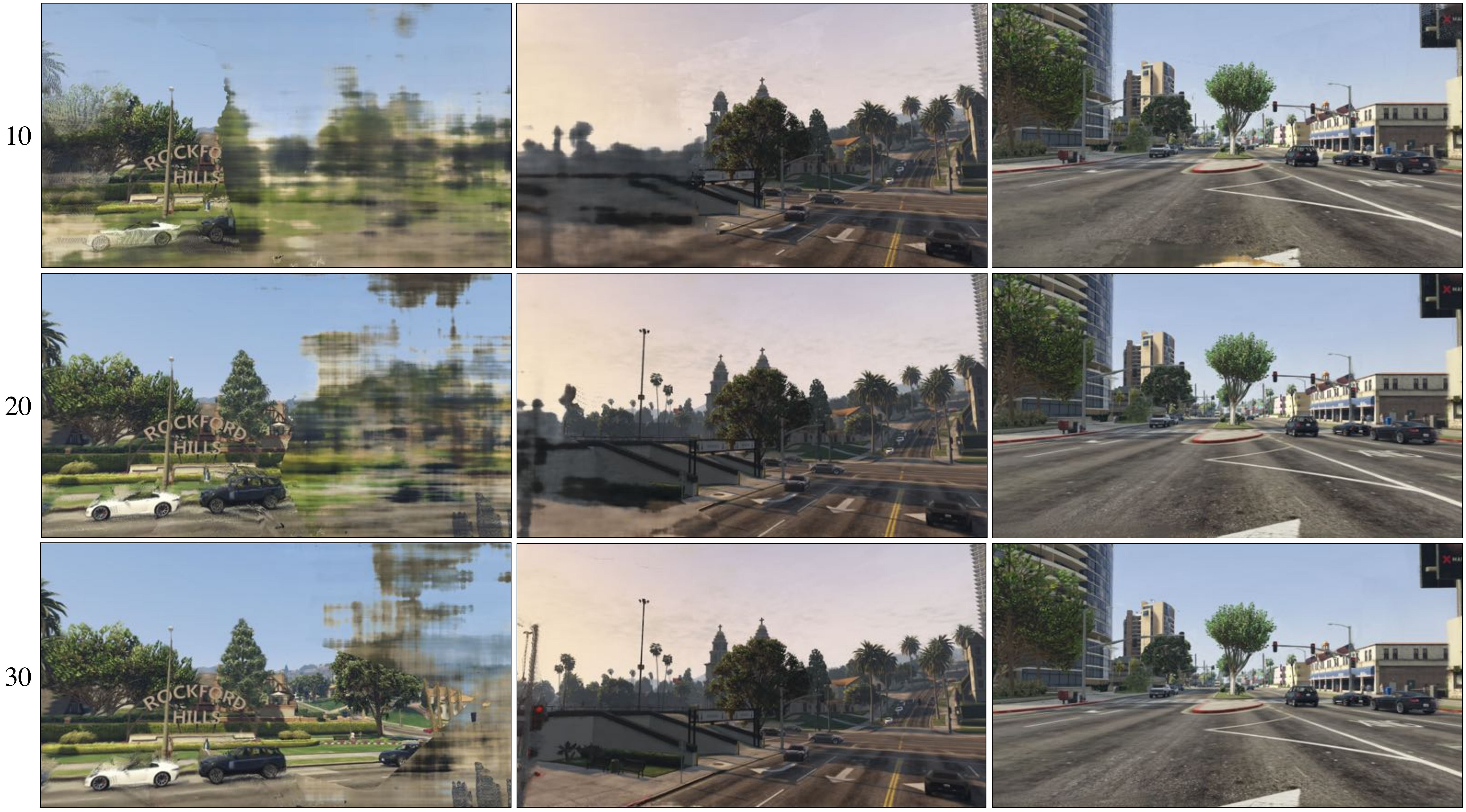}
\includegraphics[width=0.95\linewidth]{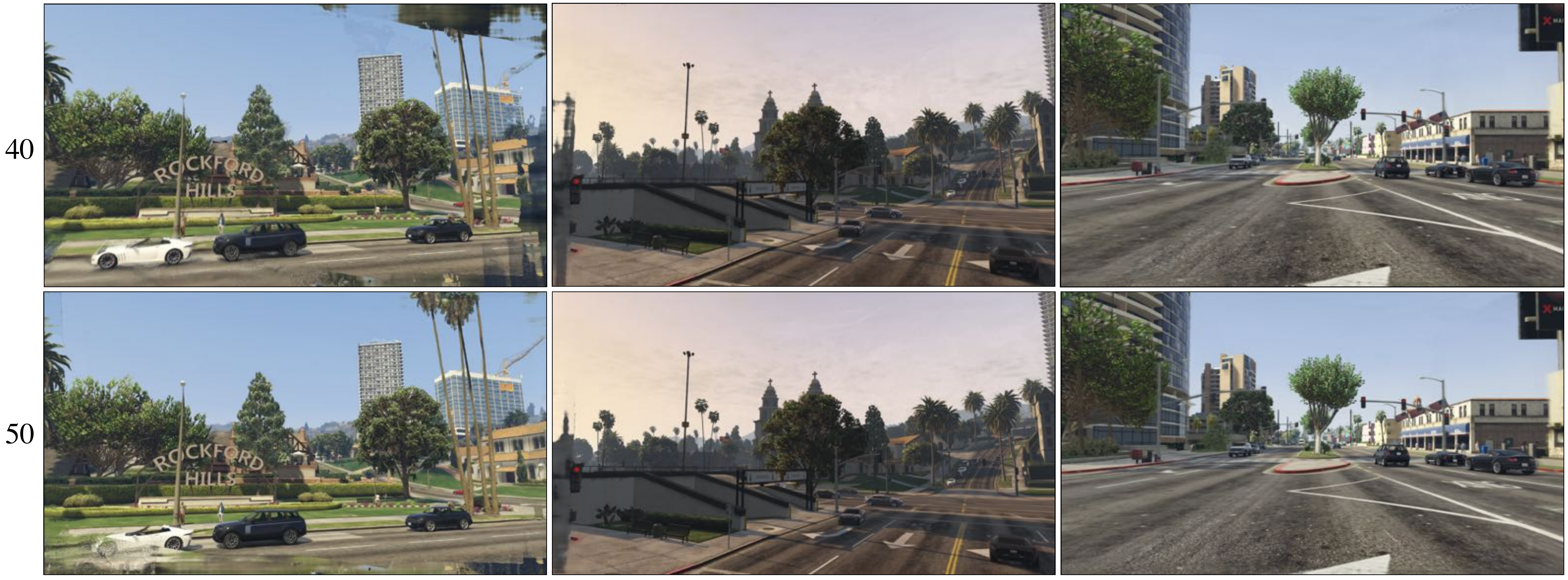}
\caption{\textbf{Varying number of views: } We vary the number of views for training our model. We show three examples here: (1) {\bf left} column shows drastic performance improvement as we increase the number of views; (2) {\bf middle} column shows improvement when increasing from $10$ to $30$ and then saturating; and (3) {\bf right} column where the improvement is little as we increase the number of views. In general, we observe that performance improves as we increase the number of views.  }
\label{fig:num-views}
\end{figure*}

We study the influence of the number of views on the quality of synthesized views. We use challenging synthetic multi-view sequences from MVS-Synth dataset~\cite{DeepMVS} that consist of different unbounded scenes. We use the first $13$ sequences with unbounded depth from this dataset for our analysis. Each sequence consists of $100$ frames. We use $50$ frames ($1920\times1080$ resolution) for evaluation, and train models by varying the number of views between $\{10,20,30,40,50\}$. The details about train-test splits are available in Appendix~\ref{app:synth}. The ground truth camera parameters are provided for these sequences. For this analysis, we train $65$ NeRF~\cite{mildenhall2020nerf} models (each for $200,000$ iterations taking roughly $420$ minutes per model) and $65$ models for our approach. Our approach takes $10-20$ minutes per sequence depending on the number of views. We contrast the performance of two methods in Table~\ref{tab:3d-synth-views}. Without any modification, our approach can generate better results. We can also generate better results with fewer views. For e.g., our method can get better results with $10$ views than NeRF with $50$ views on these sequences. Figure~\ref{fig:num-views-10} and Figure~\ref{fig:num-views-50} shows the comparison of our approach with NeRF when using $10$ and $50$ views respectively. We show the improvement in performance when using more views in Figure~\ref{fig:num-views}. Consistent with the quantitative analysis (Table~\ref{tab:3d-synth-views}), we see better results visually when increasing the number of views.

\begin{figure}
\includegraphics[width=\linewidth]{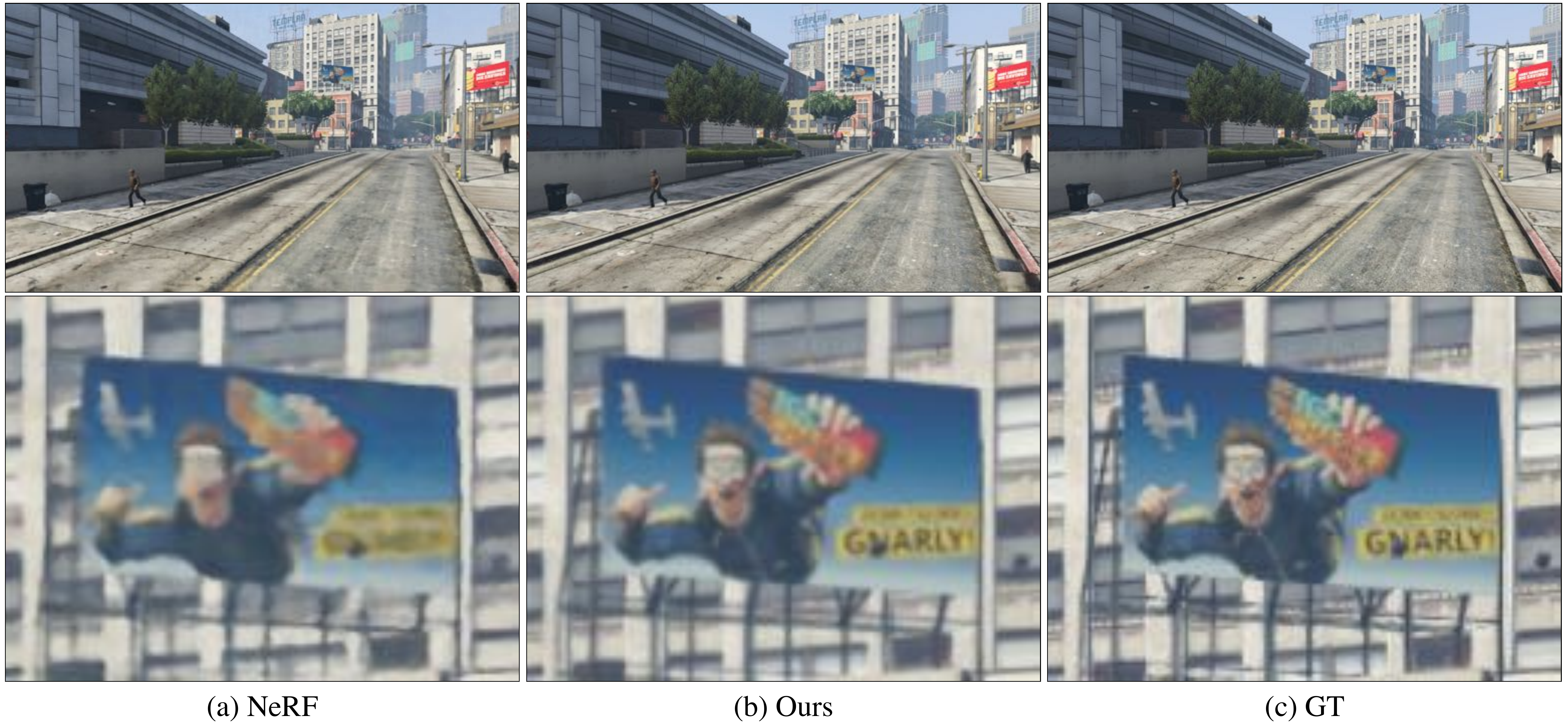}
\includegraphics[width=\linewidth]{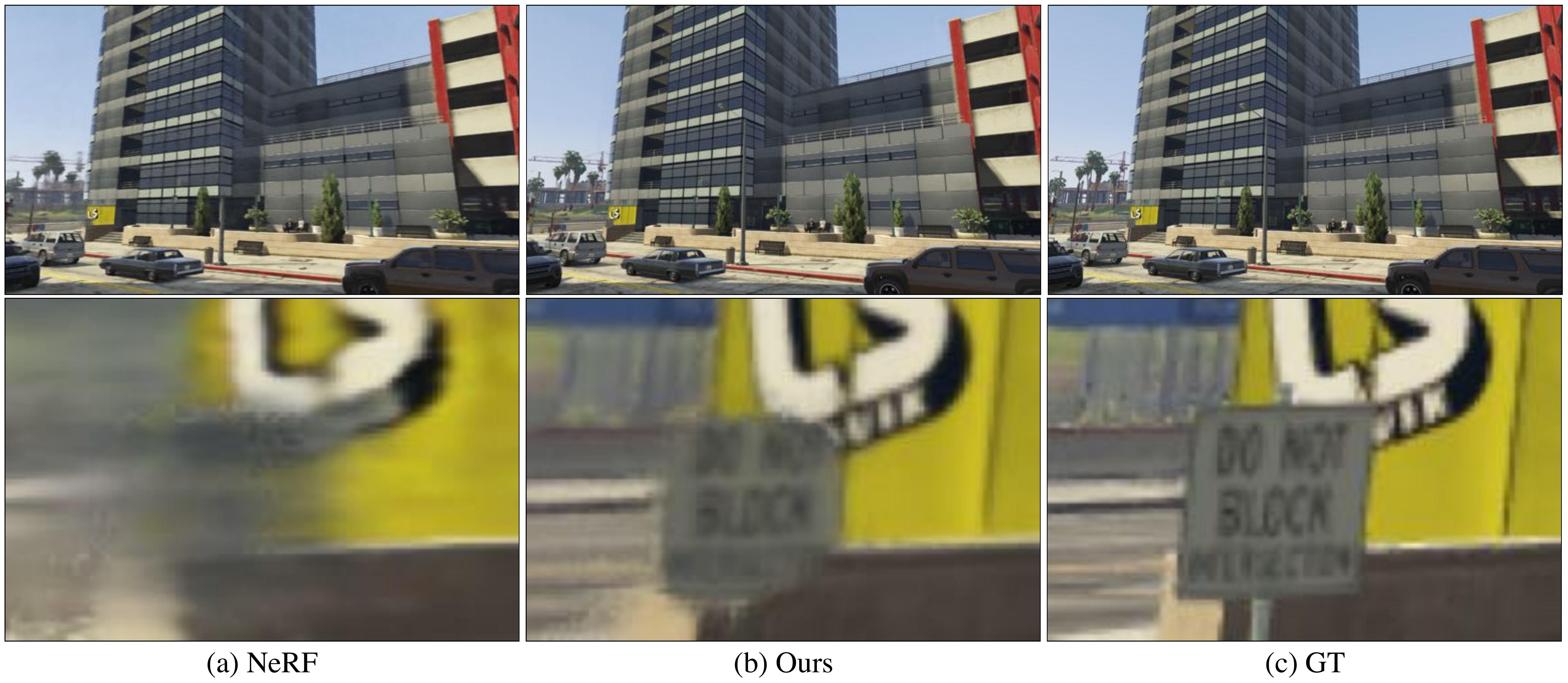}
\includegraphics[width=\linewidth]{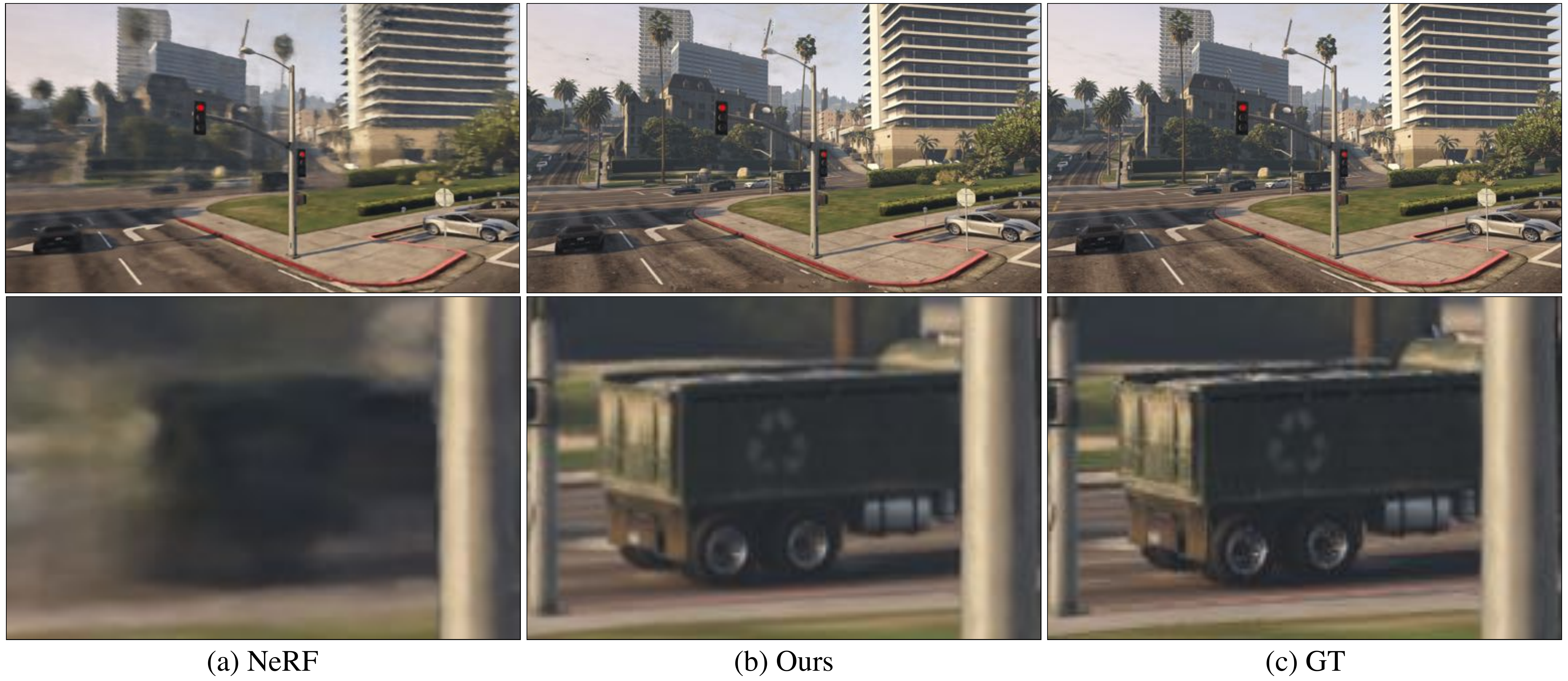}
\caption{\textbf{Best performing NeRF output on a synthetic sequence, Num Views = $50$:} {\bf (a)}  We cherry-pick the best performing synthesized result on a held-out view synthesized using NeRF trained on a sequence with unbounded depth. {\bf (b)} We then show results using our approach. We zoom in to the billboard in the center of image (top-example), towards the bottom-left in second example, and on the truck in the middle in bottom-example. Our approach captures details better than NeRF. {\bf (c)} The ground truth is shown for reference. }
\label{fig:best-nerf}
\vspace{-1.0cm}
\end{figure}

\noindent\textbf{Estimating camera parameters via SfM: } We repeat the above experiment with different camera parameters. We estimate camera parameters (intrinsics and extrinsics) from multi-views using Agisoft Metashape (a professional software used widely in industry). Table~\ref{tab:3d-synth-views-sfm} shows the performance of two approaches. We observe similar trends. The performance of NeRF improved drastically with camera parameters estimated using SfM. However, it still underperforms in comparison to our method by a large margin. We show the best performing result of NeRF on a held-out view from one of these sequences in Figure~\ref{fig:best-nerf}. We observe that our approach captures details better than NeRF. Note, we also tried COLMAP to obtain camera poses and point clouds. However, COLMAP struggles on some of these sequences.

\begin{table}
\scriptsize{
\setlength{\tabcolsep}{6pt}
\def\arraystretch{1.2}
\center
\begin{tabular}{@{}l c c  c c c}
\toprule
\textbf{13 sequences} &  & \textbf{PSNR}$\uparrow$ & \textbf{SSIM}$\uparrow$   & \textbf{LPIPS} $\downarrow$ \\ 
\midrule
{\tt num-views=$10$} &    &  &  &   \\
NeRF  & & 14.534 $\pm$ 2.001  & 0.499 $\pm$ 0.102  & 0.712 $\pm$ 0.096\\ 
Ours &   & {\bf 19.439 $\pm$ 4.375}   & {\bf 0.697 $\pm$ 0.128}  & {\bf 0.410 $\pm$ 0.177} \\
\midrule
{\tt num-views=$20$} &    &  &  &   \\
NeRF  & & 15.585 $\pm$ 2.086  & 0.524 $\pm$ 0.099   & 0.707 $\pm$ 0.096 \\ 
Ours. &   &  {\bf 23.651 $\pm$ 4.045}  & {\bf 0.813 $\pm$ 0.096}  &  {\bf 0.241 $\pm$ 0.120} \\
\midrule
{\tt num-views=$30$} &    &  &  &   \\
NeRF  & & 16.113 $\pm$ 2.091  & 0.536 $\pm$ 0.096   & 0.710 $\pm$ 0.094 \\ 
Ours &  & {\bf 25.357 $\pm$ 3.709}   & {\bf 0.846 $\pm$ 0.081}  & {\bf 0.201 $\pm$ 0.094} \\
\midrule
{\tt num-views=$40$} &    &  &  &   \\
NeRF  & & 16.561 $\pm$ 2.039 & 0.548 $\pm$ 0.094   & 0.708 $\pm$ 0.093 \\ 
Ours. &   & {\bf 26.083 $\pm$ 3.691}   & {\bf 0.865  $\pm$ 0.072}  & {\bf 0.178 $\pm$ 0.077}  \\
\midrule
{\tt num-views=$50$} &    &  &  &   \\
NeRF  & & 16.771 $\pm$ 1.955  & 0.553 $\pm$ 0.093   & 0.712 $\pm$ 0.090 \\ 
Ours &   & {\bf 26.829 $\pm$ 3.621}  & {\bf 0.878 $\pm$ 0.064}  & {\bf 0.161 $\pm$ 0.070} \\
\bottomrule
\end{tabular}
\caption{\textbf{Synthetic Multi-View Sequences of Unbounded Scenes:} We vary the number of views to synthesize target views using synthetic multi-view data. The held-out sequences are fixed in these analysis. We observe that our approach is able to generate better results with fewer views. The performance for both approaches improves as we increase the number of views. However, our method gets a substantial boost in performance as we increase the number of views.}
\label{tab:3d-synth-views}
}
\vspace{-0.9cm}
\end{table}

\begin{table}
\scriptsize{
\setlength{\tabcolsep}{6pt}
\def\arraystretch{1.2}
\center
\begin{tabular}{@{}l c c  c c c}
\toprule
\textbf{13 sequences} &  & \textbf{PSNR}$\uparrow$ & \textbf{SSIM}$\uparrow$   & \textbf{LPIPS} $\downarrow$ \\ 
\midrule
{\tt num-views=$10$} &    &  &  &   \\
NeRF  & & 16.150 $\pm$ 4.195 & 0.541 $\pm$ 0.139 & 0.619 $\pm$ 0.158 \\
Ours &   & {\bf 18.460 $\pm$ 4.099} & {\bf 0.656 $\pm$ 0.129} &  {\bf 0.451 $\pm$ 0.167} \\
\midrule
{\tt num-views=$20$} &    &  &  &    \\
NeRF  & & 18.171 $\pm$ 4.543 & 0.582 $\pm$ 0.135  &  0.594 $\pm$ 0.171 \\
Ours. &   & {\bf 22.414 $\pm$ 4.197} & {\bf 0.766 $\pm$ 0.126}  & {\bf 0.289 $\pm$ 0.147}  \\
\midrule
{\tt num-views=$30$} &    &  &  &   \\
NeRF  & & 19.725 $\pm$ 4.759  & 0.619 $\pm$ 0.135  & 0.557 $\pm$ 0.179 \\ 
Ours & & {\bf 24.191 $\pm$ 4.219}  & {\bf 0.803 $\pm$ 0.122}  & {\bf 0.243 $\pm$ 0.137}  \\
\midrule
{\tt num-views=$40$} &    &  &  &   \\
NeRF  & & 20.074 $\pm$ 4.673  &  0.627 $\pm$ 0.132   &  0.556 $\pm$ 0.178 \\
Ours. &   & {\bf 24.832 $\pm$ 4.110}  & {\bf 0.822 $\pm$ 0.117}  & {\bf 0.218 $\pm$ 0.125}   \\
\midrule
{\tt num-views=$50$} &    &  &  &   \\
NeRF  & & 20.244 $\pm$ 4.611  & 0.631 $\pm$ 0.129  &  0.556 $\pm$ 0.178\\
Ours &  & {\bf 25.529 $\pm$ 4.212}  & {\bf 0.836 $\pm$ 0.112}  & {\bf 0.198 $\pm$ 0.116}  \\
\bottomrule
\end{tabular}
\caption{\textbf{Camera Parameters Estimated using SfM for Synthetic Multi-View Sequences:} We repeat the experiment in Table~\ref{tab:3d-synth-views} with camera parameters estimated using Agisoft Metashape given multi-views from a sequence. Once again, we observe similar trends. Interestingly, NeRF improves upon previous results. However, it still underperforms as compared to our method by a large margin.}
\label{tab:3d-synth-views-sfm}
}
\vspace{-0.9cm}
\end{table}

\subsection{Hi-Res Studio Capture}
\label{ssec:studio}

\begin{figure}
\includegraphics[width=\linewidth]{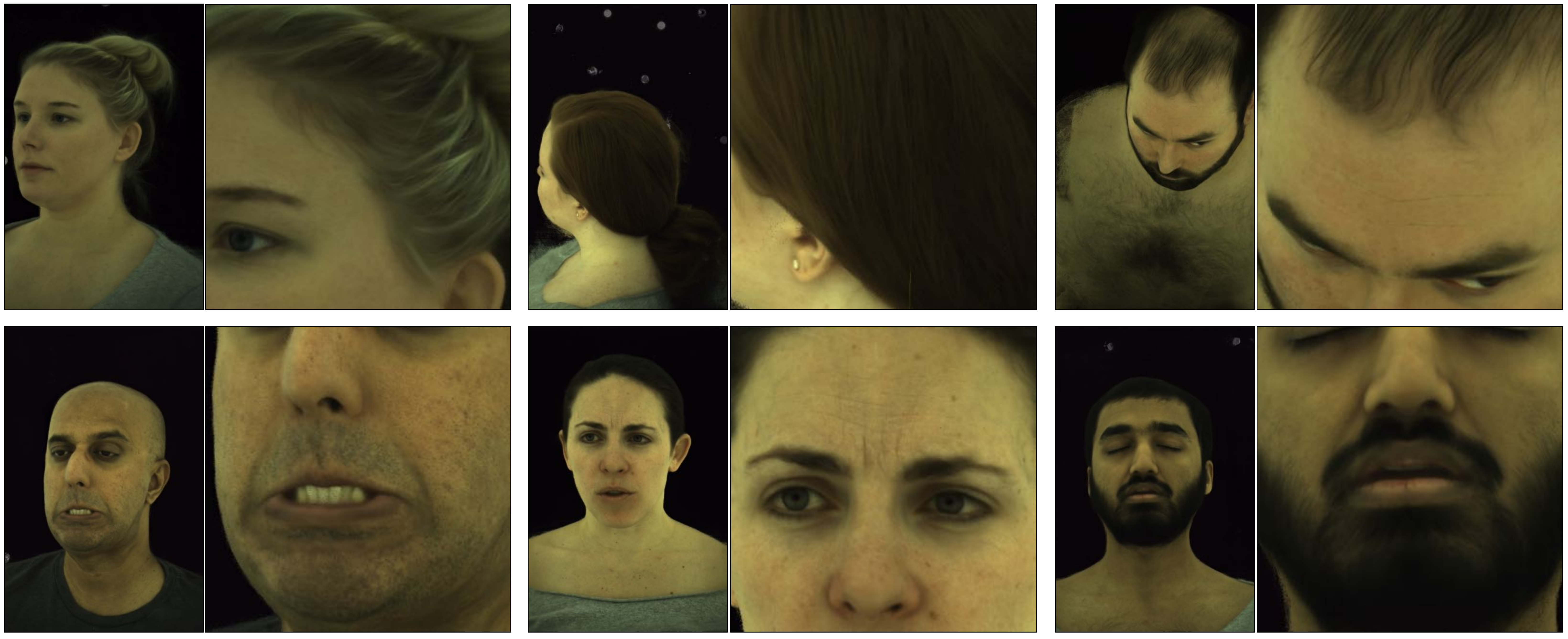}
\caption{\textbf{Hi-Res Facial Details: } Our approach allows us to capture hi-res facial details. We show novel views synthesized for various subjects and emphasize different regions on the face to show details such as hair, eyes, teeth, and skin details.}
\label{fig:faces}
\end{figure}

\noindent\textbf{Multi-View Facial Capture: }  We employ multi-view hi-res facial captures. We can synthesize hi-resolution novel views with a few minutes of training without any modification and without using any expert knowledge such as facial details, foreground-background etc. Figure~\ref{fig:faces} shows novel views synthesized and facial details (such as hair, eyes, wrinkles, teeth, etc.) captured using a model trained for a specific subject.

\begin{figure}
\centering
\includegraphics[width=0.95\linewidth]{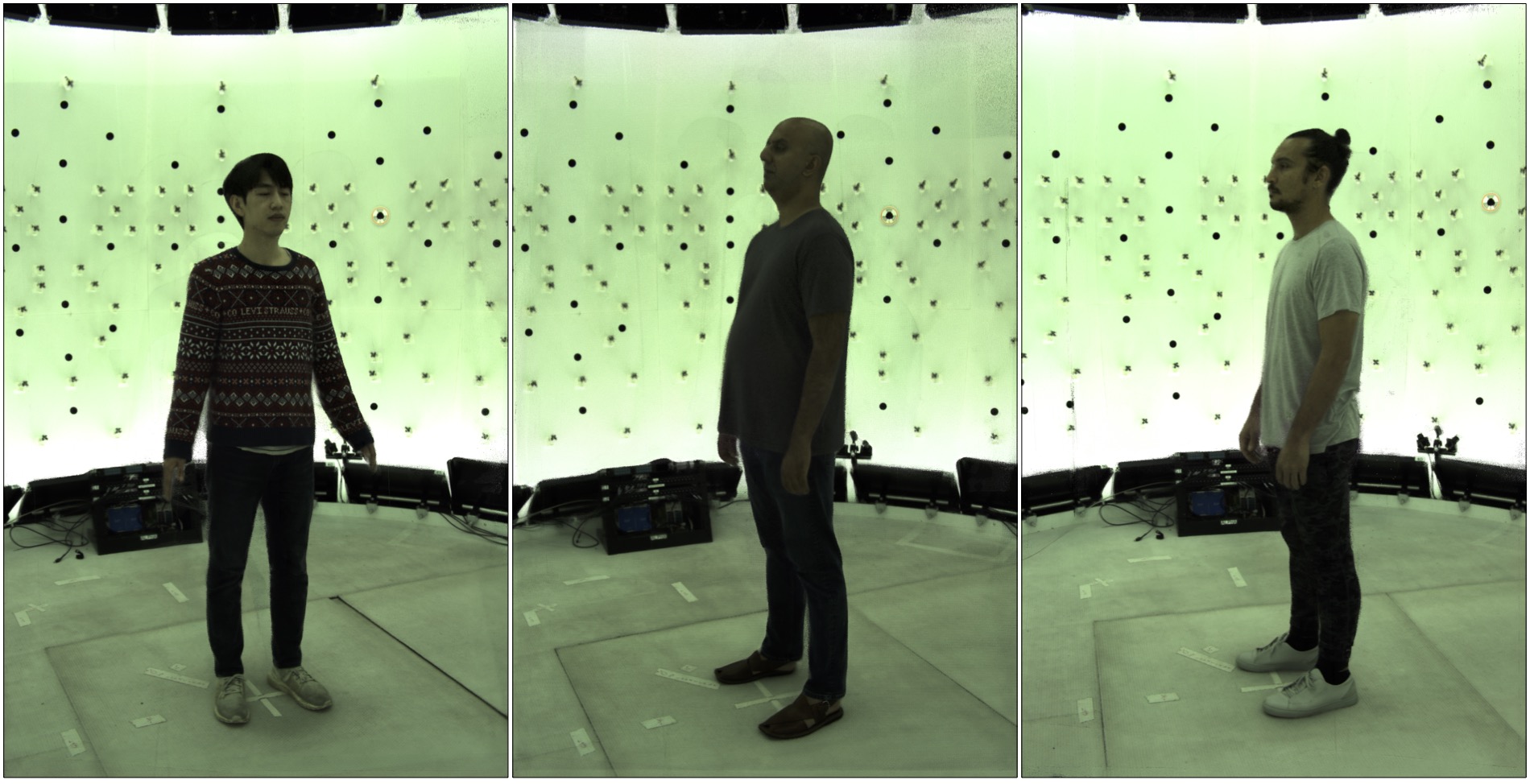}
\includegraphics[width=0.95\linewidth]{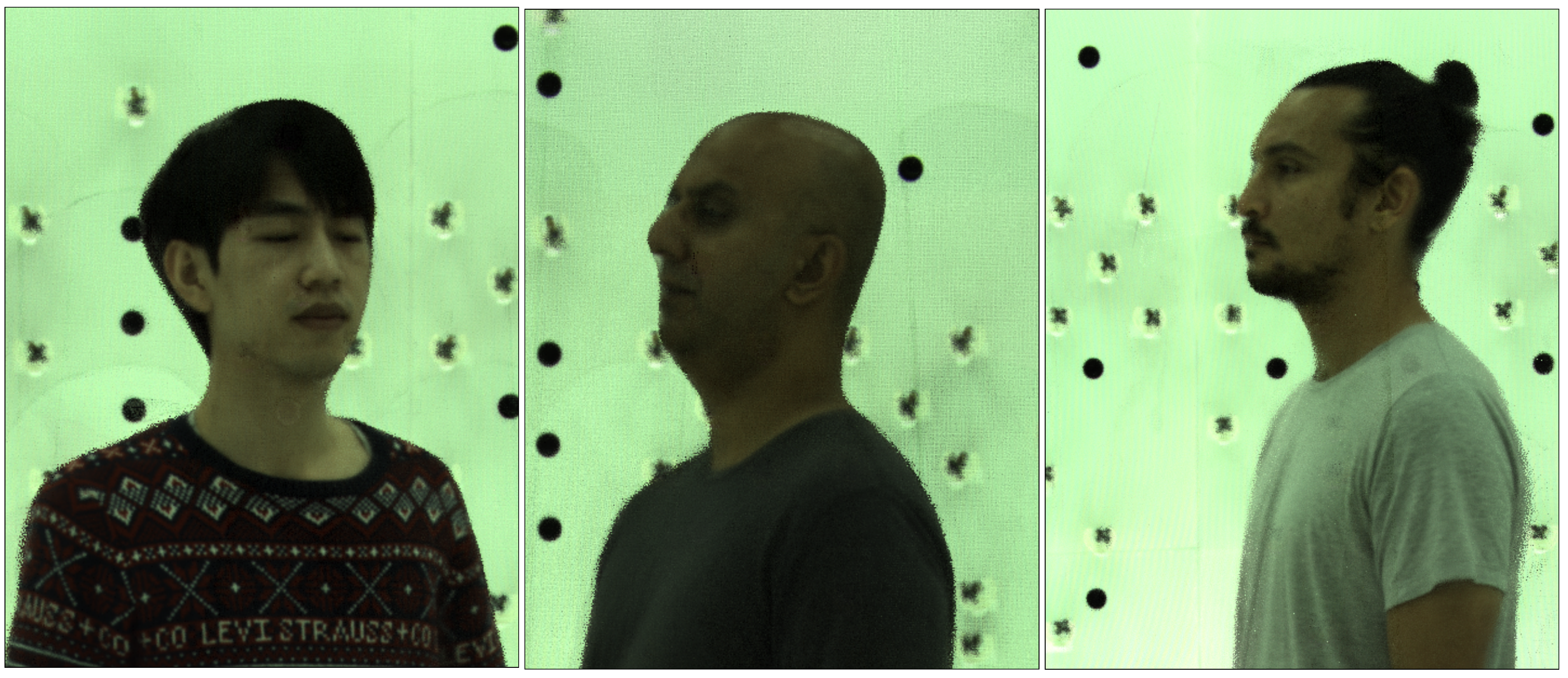}
\caption{\textbf{Hi-Res Body Synthesis: } Our approach allows us to synthesize high quality  novel views of human bodies. In the bottom-row, we zoom to see the details captured on the face for each of three subjects. \textcolor{blue}{Best viewed in electronic format}.  }
\label{fig:body}
\end{figure}

\noindent\textbf{Multi-View Full Body Capture: } Our approach also enables us to synthesize full-bodies from hi-res multi-view captures. Once again, we did not use any human-body specific information. Figure~\ref{fig:body} shows novel views synthesized and body details captured using a model trained for a specific subject.

\begin{figure}
\includegraphics[width=\linewidth]{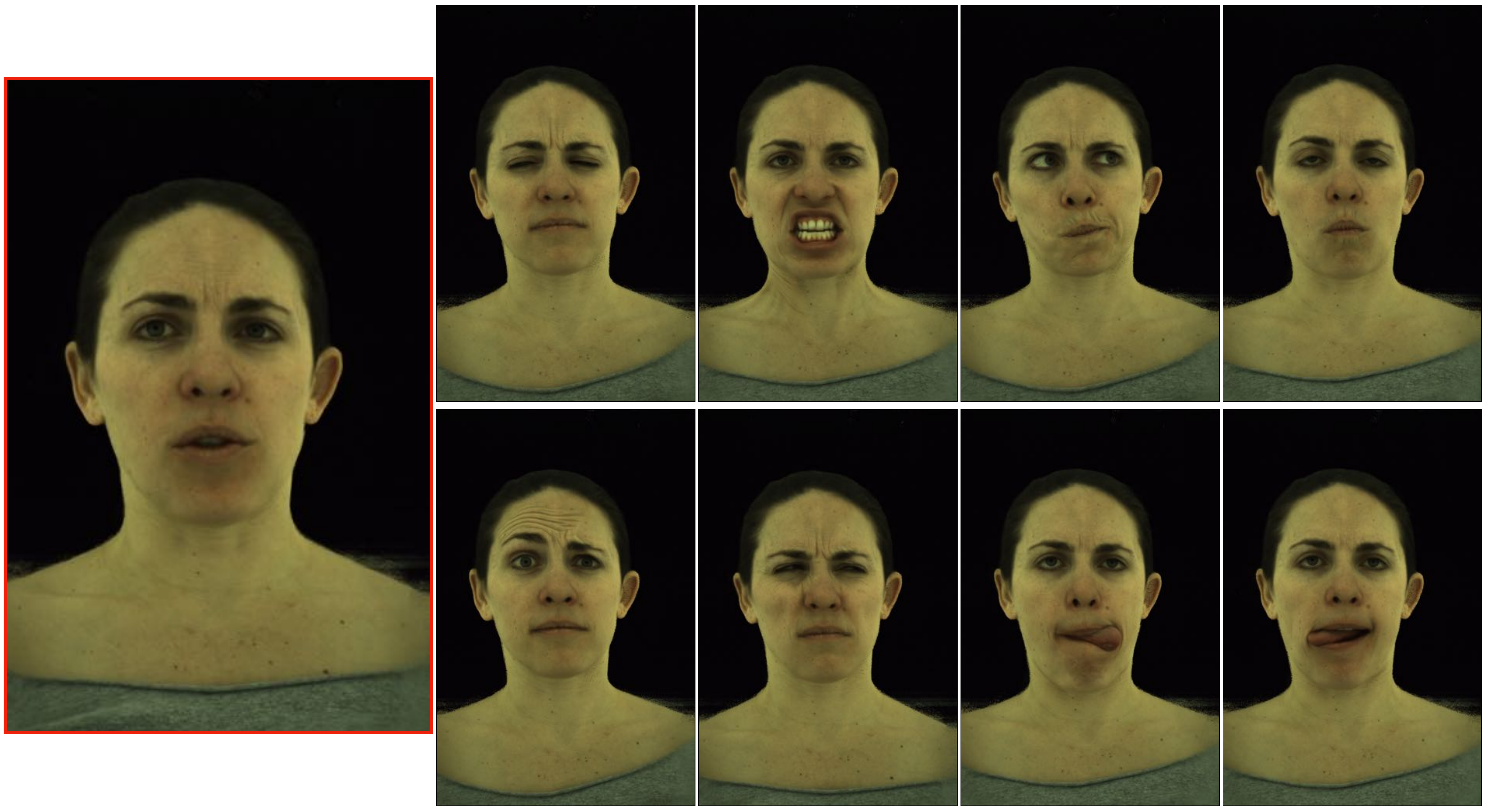}
\includegraphics[width=\linewidth]{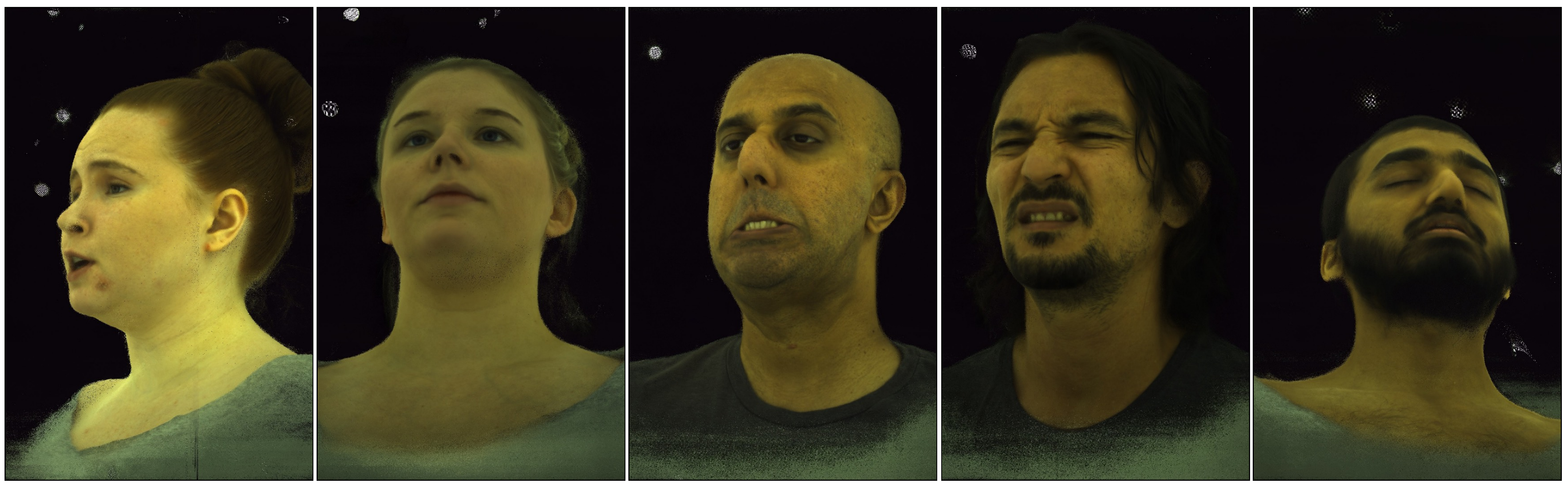}
\caption{\textbf{Generalization to unseen time instants and unseen subjects: } The model is trained on a single time instant -- shown on top-left. Our model generalizes to unseen expressions (top-right) and unseen subjects (bottom row). }
\label{fig:expressions}
\end{figure}

\begin{figure}
\includegraphics[width=\linewidth]{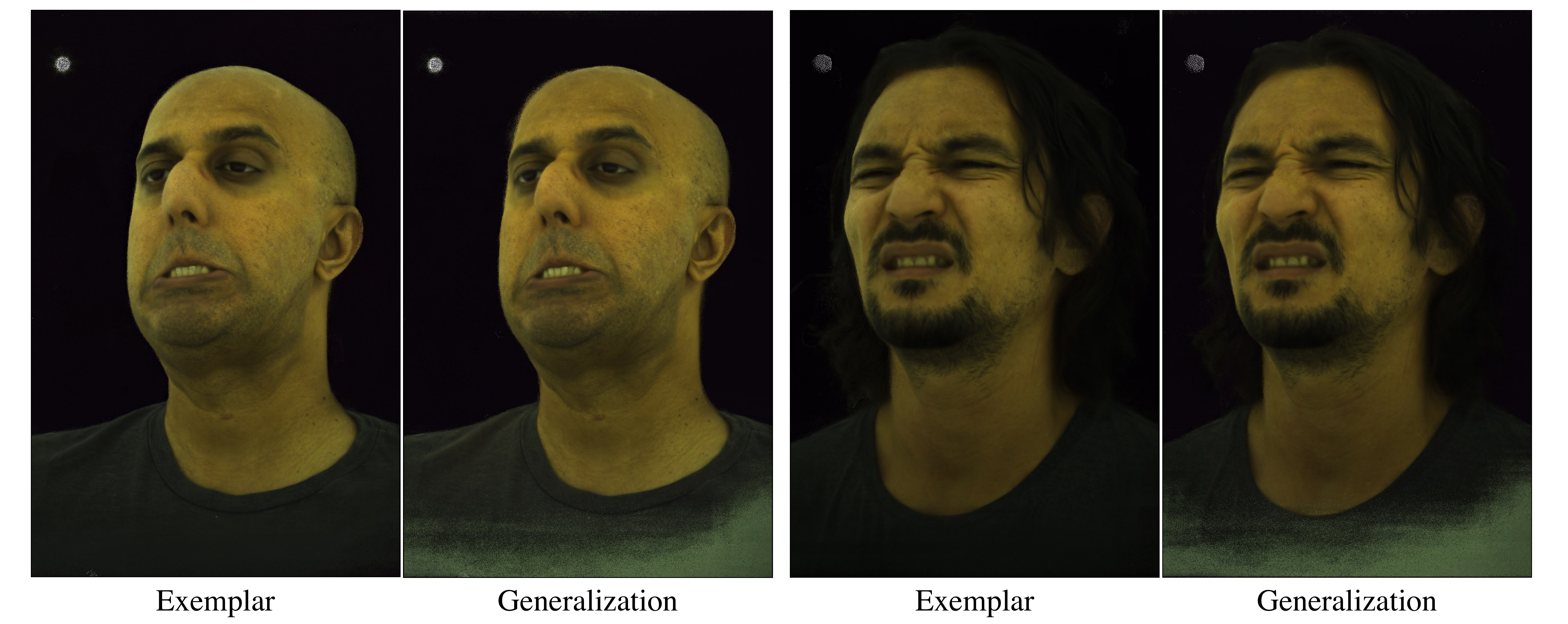}
\caption{\textbf{Contrasting Exemplar Models and Generalization: } We contrast the results from exemplar model with the results obtained using a model that has never seen these subjects. We term it {\bf Generalization} here.}
\label{fig:identities}
\end{figure}

\noindent\textbf{Ability to Generalize: } An important aspect of our approach is to enable generalization to unseen time instants and unknown subjects. We train a model for one time instant of one subject and can use it to synthesize new views for unknown time instants. We show extreme facial expressions and unseen subjects in Figure~\ref{fig:expressions}. We also contrast the results of generalization with a subject-specific model in Figure~\ref{fig:identities}. We observe that the learned model generalizes well except for the clothing in the bottom part of the images. We posit that there isn't sufficient coverage from multi-views in that area. However, an exemplar model learned for a specific subject is able to capture the details. We leave the reader with an open philosophical question as to whether we should think about generalization if we can learn an exemplar model for a given data distribution in a few seconds? 

\subsection{Convergence Analysis}
\label{ssec:conv}

\begin{figure}
\includegraphics[width=\linewidth]{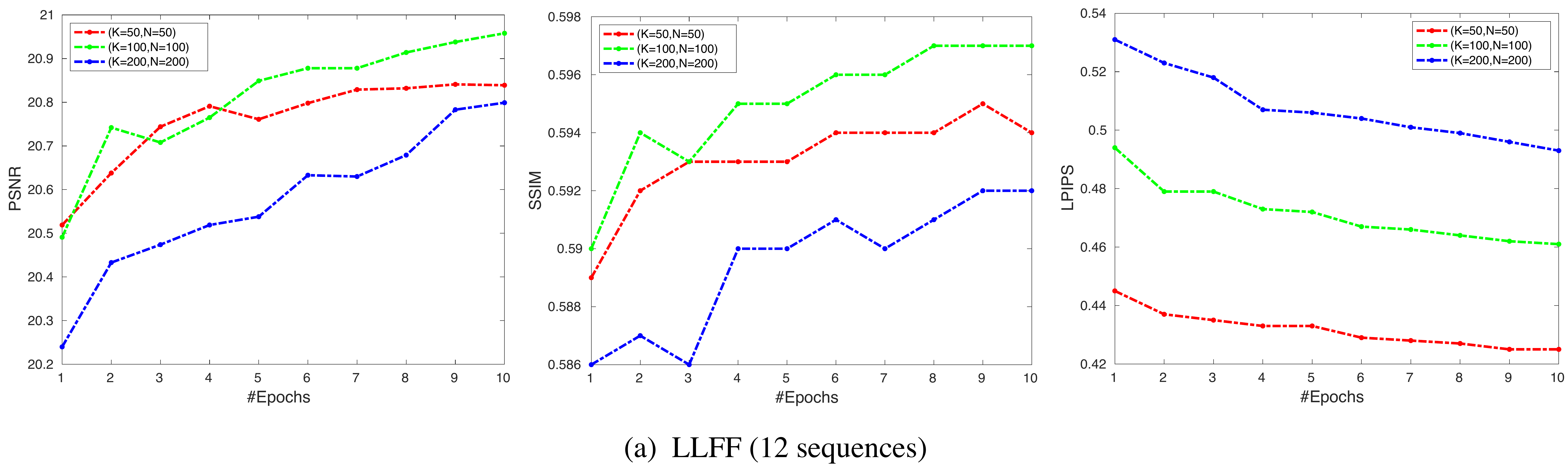}
\includegraphics[width=\linewidth]{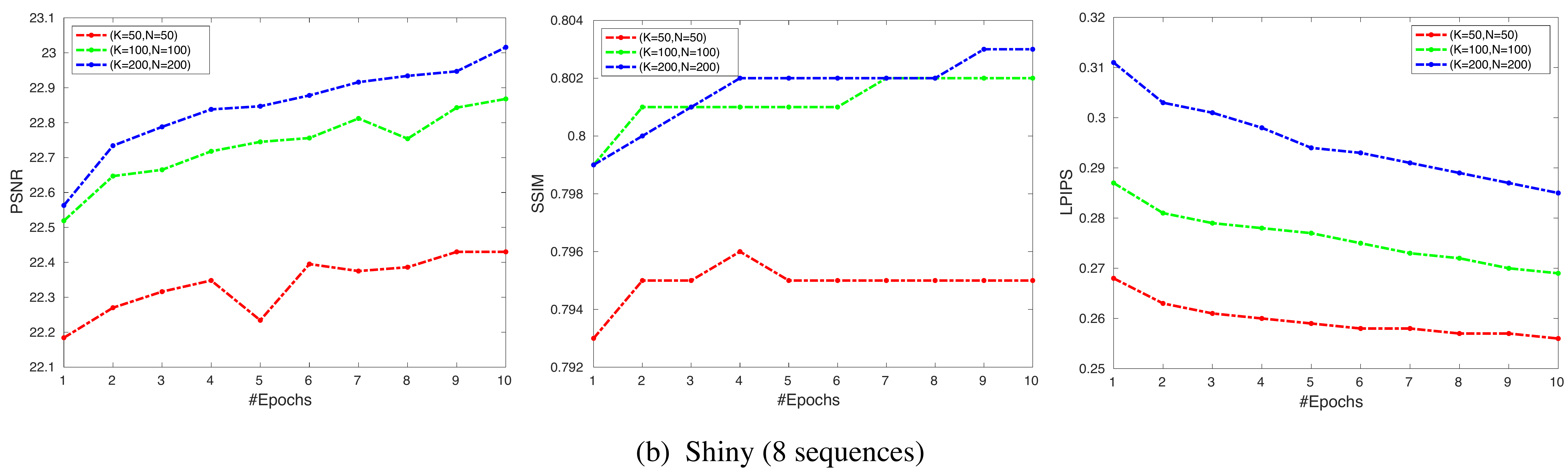}
\includegraphics[width=\linewidth]{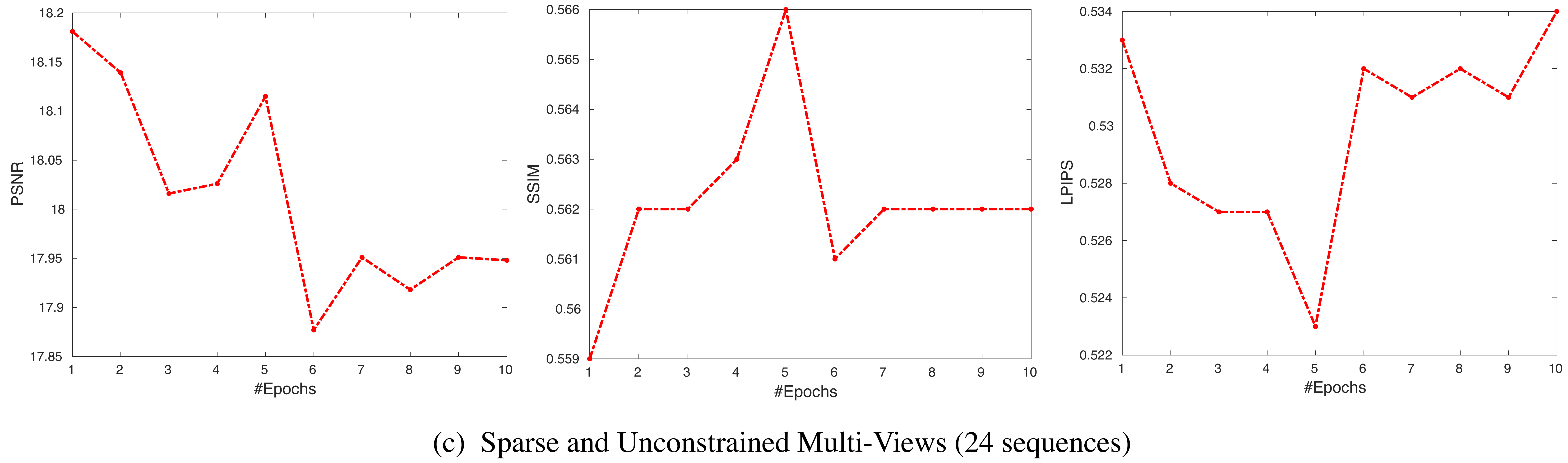}
\caption{\textbf{Convergence Analysis: } We study the convergence properties of our approach using $12$ LLFF sequences~\cite{mildenhall2019local} and $8$ sequences from Shiny Dataset~\cite{Wizadwongsa2021NeX}. We also study convergence using $24$ sparse and unconstrained multi-view sequences. We observe that our model gets close to convergence in the first few seconds. Note the difference in values on y-axis is small.}
\label{fig:plots}
\vspace{-1.cm}
\end{figure}

We study the convergence properties of our approach using $12$ LLFF sequences~\cite{mildenhall2019local} (Appendix~\ref{app:hi-res}) and Shiny Dataset~\cite{Wizadwongsa2021NeX}. We show the plots in Figure~\ref{fig:plots} for model training in the first $10$ epochs, i.e. from 60 seconds to 600 seconds. We observe that our model gets close to convergence in the first few seconds. Crucially, our approach obtains competitive results to prior work on the Shiny dataset within 60 seconds of training as compared to 64 hours for NeRF~\cite{mildenhall2020nerf} on full-resolution and $24-30$ hours of training of NeX~\cite{Wizadwongsa2021NeX} on one-fourth resolution. We also study convergence using $24$ sparse and unconstrained multi-view sequences (Appendix~\ref{app:sparse3d}). Training an epoch on these sequences roughly take $10$ seconds because these are sparse. We observe that model gets close to the best performance in the first $10$ seconds of training. The raw values for the plots are available in Appendix~\ref{app:conv}.

\begin{figure*}[t]
\includegraphics[width=\linewidth]{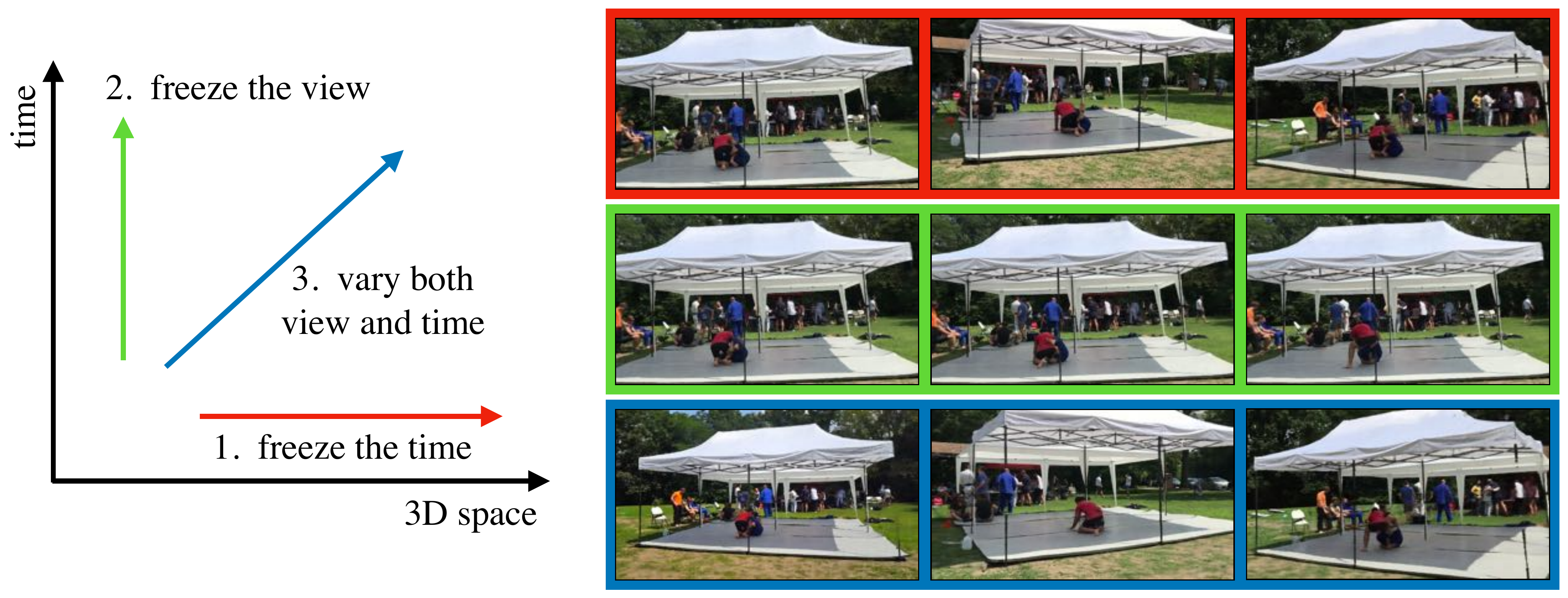}
\caption{\textbf{4D view synthesis: } We demonstrate our approach for 4D view synthesis on the challenging Open4D dataset~\cite{Bansal_2020_CVPR}. Without any background-foreground modeling or any modification, our approach learns to perform 4D visualization of dynamic events. \textbf{(1)}. We can freeze the time/event and move the view. \textbf{(2)}. We can freeze the view and see the event happening. \textbf{(3)}. We can vary both view and time.}
\label{fig:4dv-teaser}
\vspace{-.4cm}
\end{figure*}

\begin{figure*}[h!]
\includegraphics[width=\linewidth]{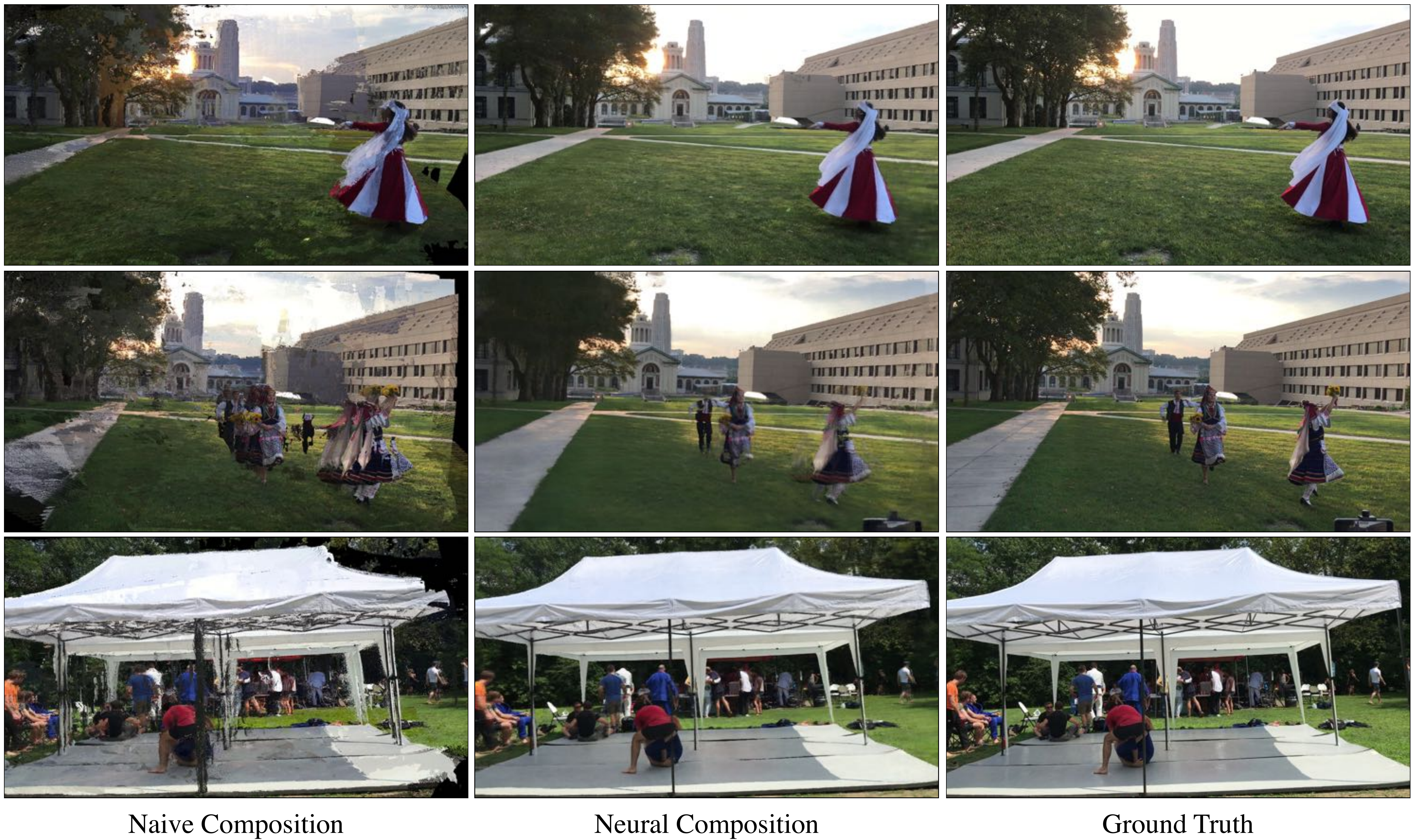}
\caption{\textbf{Naive Composition vs. Neural Composition for 4D View Synthesis: } We contrast the performance of naive composition using depth ordering with neural pixel composition for unseen temporal sequences. We observe that neural composition allows us to generate more realistic views in contrast to the naive composition. }
\label{fig:4d-naive}
\end{figure*}

\section{4D Multi-View View Synthesis}
\label{sss:4dviews}

\begin{figure*}[t]
\includegraphics[width=\linewidth]{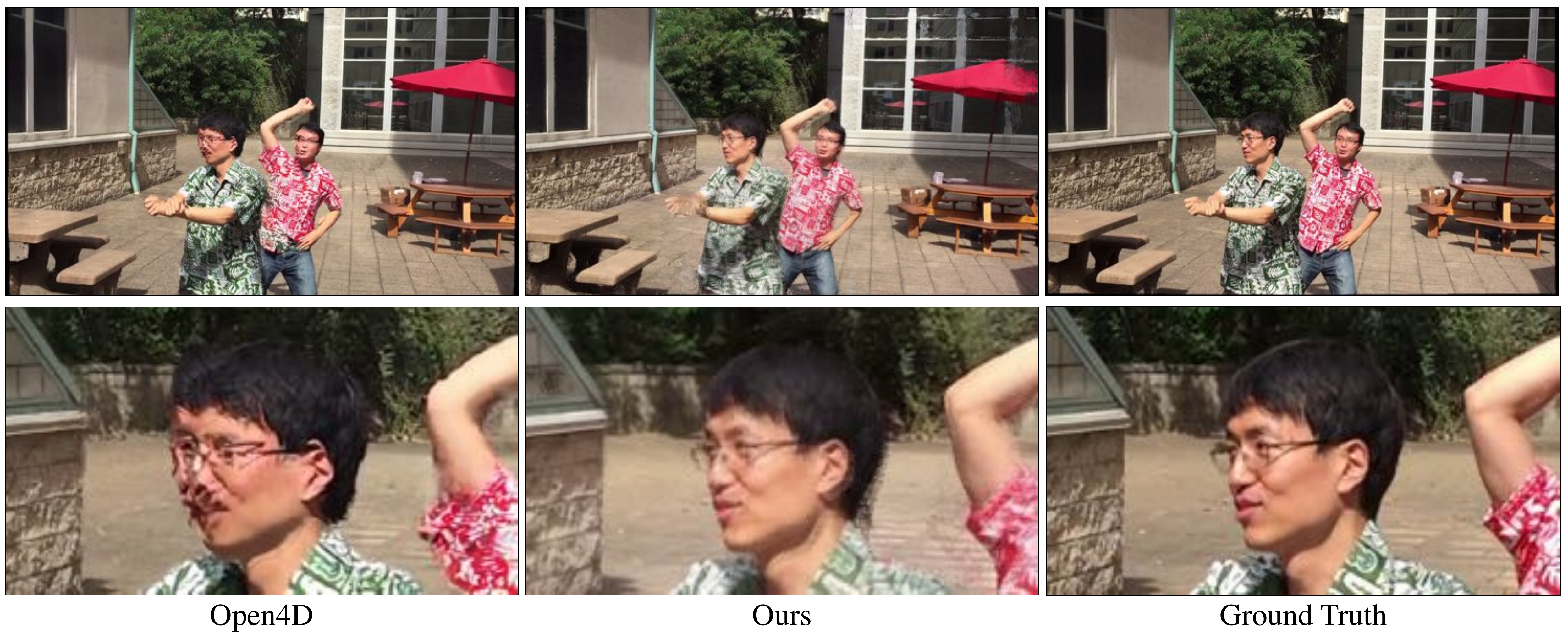}
\includegraphics[width=\linewidth]{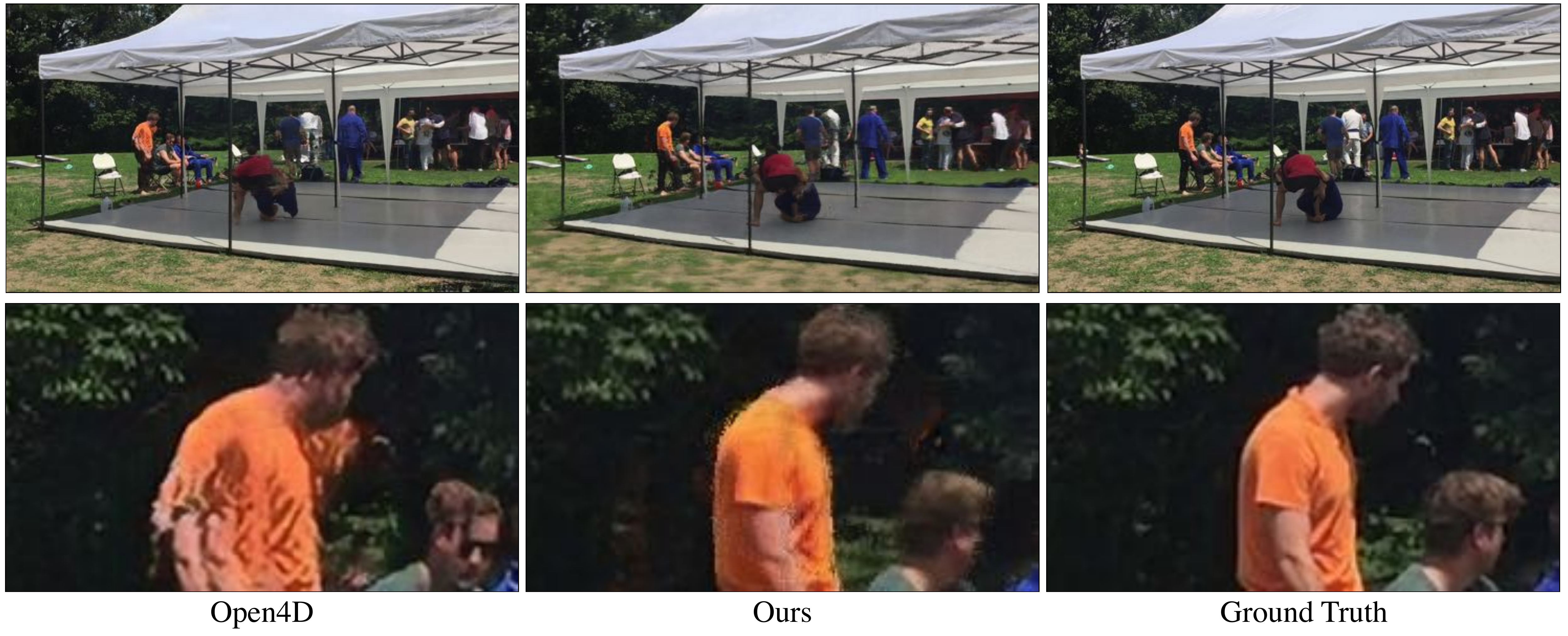}
\caption{\textbf{Unseen Temporal Sequences: } We contrast Open4D with ours for unseen temporal sequences. We observe that our approach allows us to capture details (such as details on human faces) consistently better than Open4D. }
\label{fig:4d-open4d}
\end{figure*}

We study the ability of our approach to perform 4D view synthesis. We train our model on the temporal sequences ($1920\times1080$ resolution) from the Open4D dataset and contrast our approach with their method~\cite{Bansal_2020_CVPR}. Figure~\ref{fig:4dv-teaser} shows different things that we can do using our approach without any modification.

\begin{figure*}[t!]
\includegraphics[width=\linewidth]{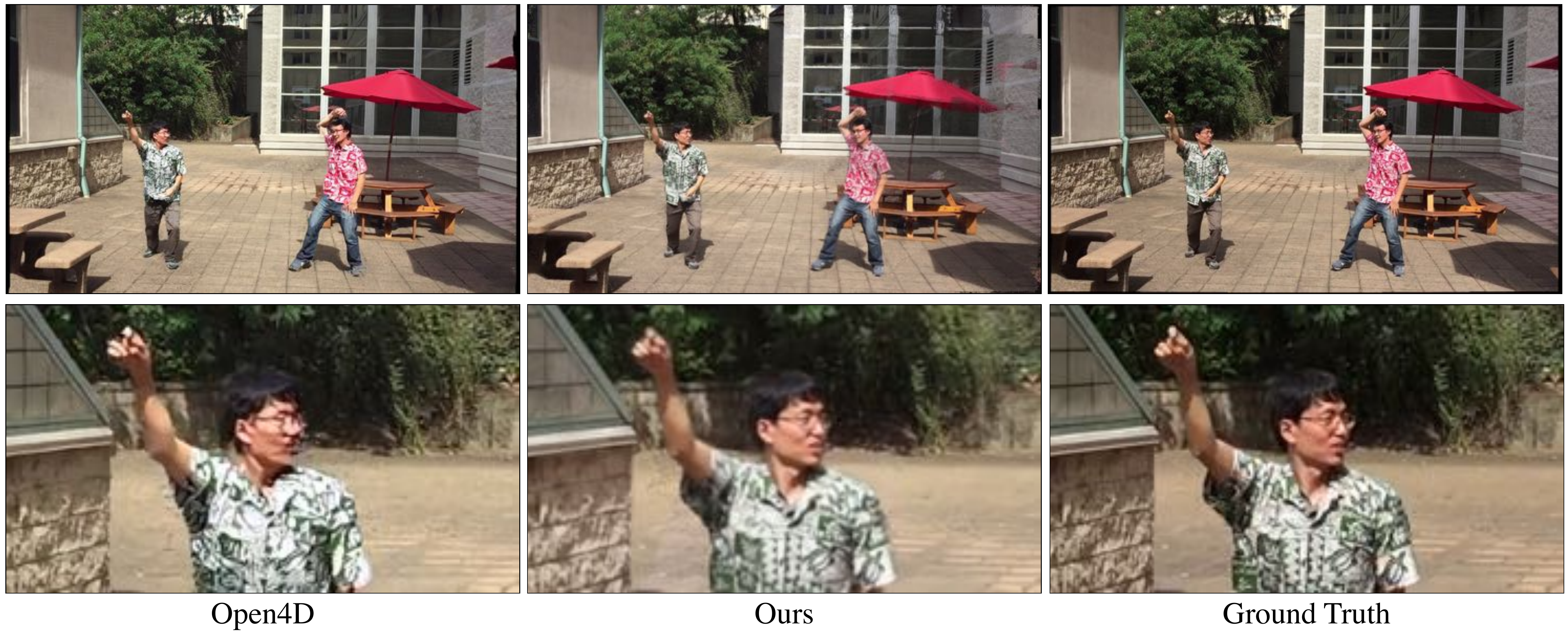}
\includegraphics[width=\linewidth]{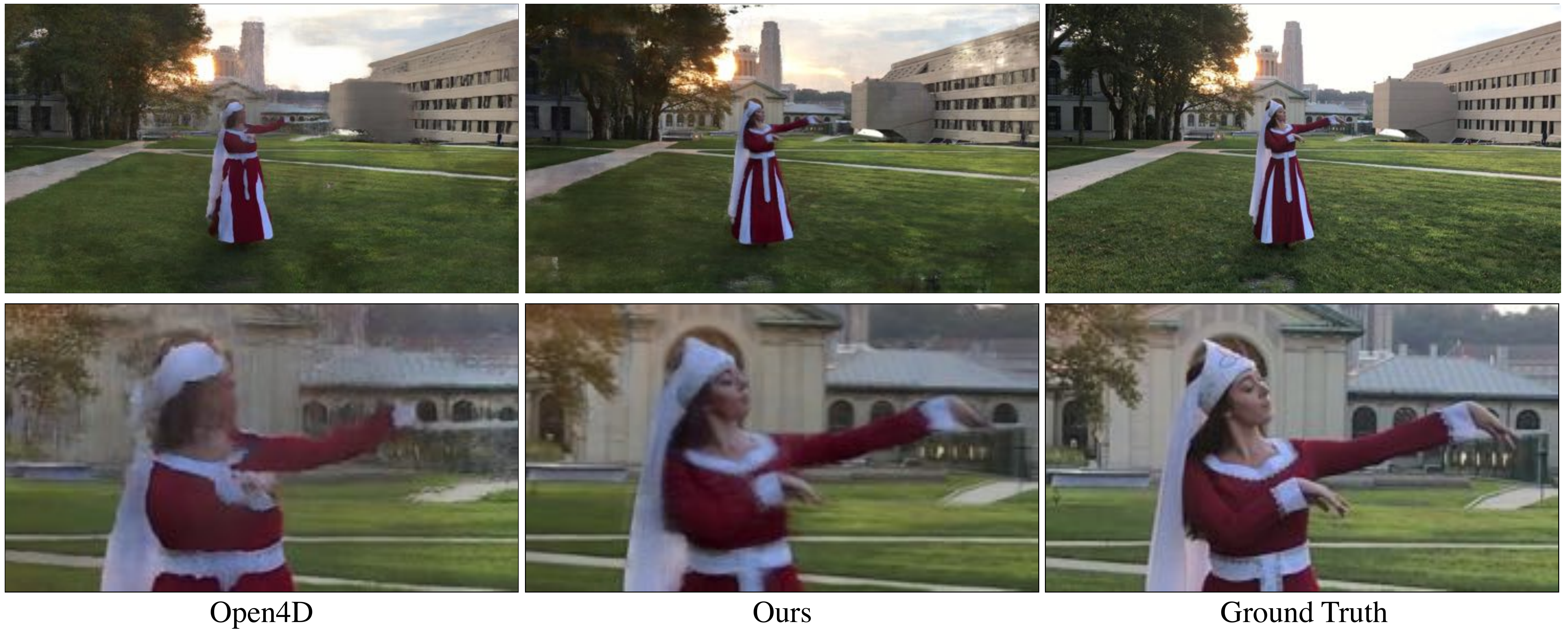}
\includegraphics[width=\linewidth]{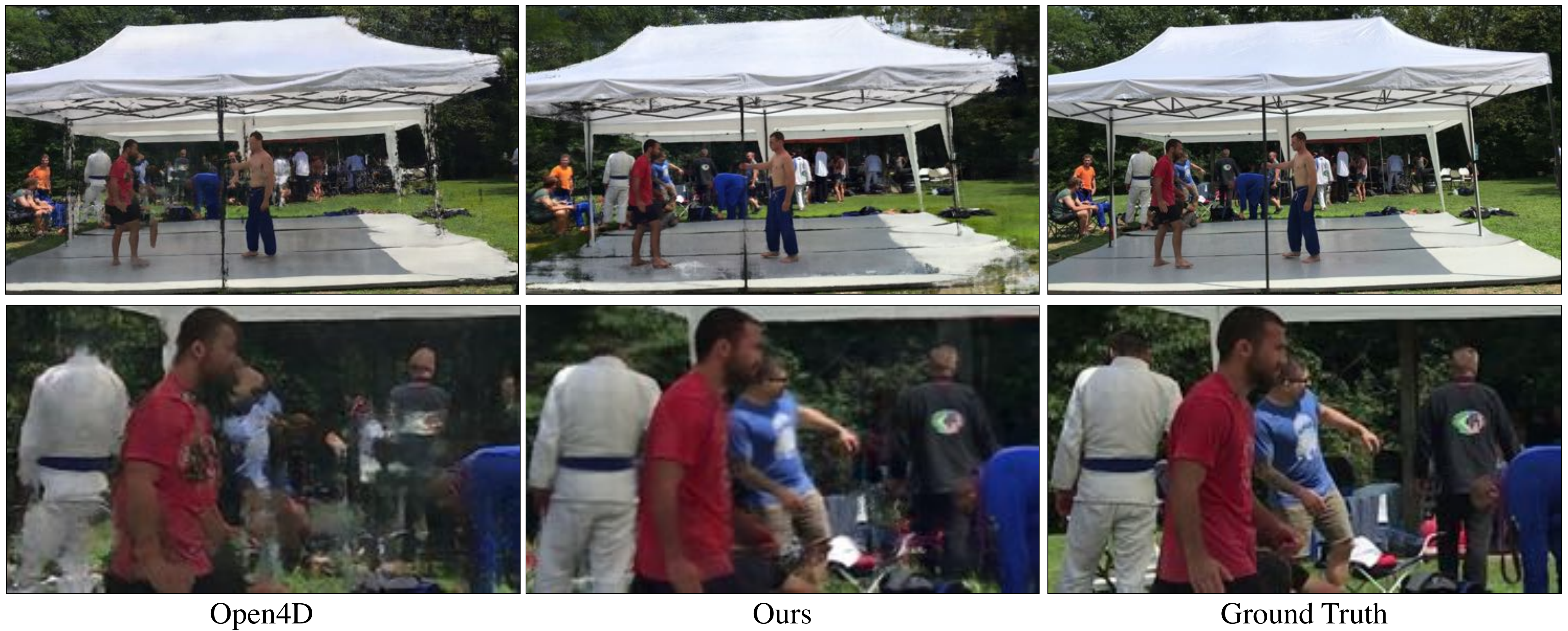}
\caption{\textbf{Held-Out Camera Views: } We contrast Open4D with ours for held-out camera views. Once again, we observe that our approach allows us to capture consistent details (such as details on human faces) better than Open4D. }
\label{fig:4d-open4d-ho}
\end{figure*}

Open4D computes foreground and background images, and trains a modified U-Net model~\cite{ronneberger2015u} for composition. The foreground image is computed by a naive composition of pixels from multi-views using depth ordering (as shown on the left side in Fig~\ref{fig:4d-naive}). The background image is computed by averaging foreground images for various time instances. We conduct two experiments: (1) held-out temporal sequences; and (2) held-out camera views.

\noindent\textbf{Held-out temporal sequences: } In the first experiment, we study the performance of the trained model on unseen temporal sequences. We train the model without temporal constraint. Our goal is to study the compositional ability of our model in contrast to the more explicit Open4D. The model is trained with multi-views available for $300-400$ time instances and evaluated on unseen $100$ time instances. Table~\ref{tab:4d-views} contrasts the performance of our approach with Open4D. Quantitatively, we observe similar performance of our approach as compared to Open4D on unseen temporal sequences. We observe better qualitative results as shown in Figure~\ref{fig:4d-open4d}. Our approach is able to capture details such as human faces consistently better than Open4D. Crucially, our approach does not require explicit foreground-background modeling and can work with arbitrary temporal sequences. The details of the sequences are in Sec~\ref{app:open4d}.

\begin{table}[h]
\scriptsize{
\setlength{\tabcolsep}{3pt}
\def\arraystretch{1.3}
\center
\begin{tabular}{@{}l c c  c c }
\toprule
\textbf{5 sequences} & & \textbf{PSNR}$\uparrow$ & \textbf{SSIM}$\uparrow$   & \textbf{LPIPS} $\downarrow$ \\
\midrule
Naive Composition & & 13.723 $\pm$ 2.759    &  0.342 $\pm$ 0.110  &  0.665 $\pm$ 0.113  \\
Open4D~\cite{Bansal_2020_CVPR} & &  20.355 $\pm$ 4.425  & 0.626 $\pm$ 0.131  &  {\bf 0.306 $\pm$ 0.079} \\
Ours &    & {\bf 21.458 $\pm$ 4.690}  & {\bf 0.645 $\pm$ 0.145 }  & 0.431 $\pm$ 0.139   \\
\bottomrule
\end{tabular}
\caption{\textbf{Unseen Temporal Sequences: }  We study the compositional ability of our model in contrast to the more explicit Open4D. The model is trained with multi-views available for $300-400$ time instances and evaluated on unseen $100$ time instances. There are a total of $5297$ frames used for evaluation. Our approach is able to generate results competitive to Open4D without any modification.}
\label{tab:4d-views}
}
\end{table}

 \noindent\textbf{Held-out camera views: }In the second experiment, we study the performance on unseen camera views but a known temporal sequence. We train the model for $500$ time instances with and without temporal constraint to understand its importance. Table~\ref{tab:4d-views-02} contrasts the performance of our approach with Open4D. Without any heuristics and foreground-background estimation, we are able to learn a representation that allows 4D view synthesis. Our approach use a simple reconstruction loss whereas Open4D use an additional adversarial loss~\cite{goodfellow2014generative}. Using the adversarial loss enables Open4D to generate overall sharp results that leads to lower LPIPS score. We contrast our approach with Open4D in Figure~\ref{fig:4d-open4d-ho}. Once again, we observe that our approach is able to capture details (facial and body details) better than Open4D. Finally, incorporating temporal constraint as the input to the model further improves performance. The details of the sequences are in Sec~\ref{app:open4d-ho}.

\begin{table}[h]
\scriptsize{
\setlength{\tabcolsep}{3pt}
\def\arraystretch{1.3}
\center
\begin{tabular}{@{}l c c  c c }
\toprule
\textbf{5 sequences} & & \textbf{PSNR}$\uparrow$ & \textbf{SSIM}$\uparrow$   & \textbf{LPIPS} $\downarrow$ \\
\midrule
Naive Composition & &  14.584 $\pm$ 3.364   & 0.374 $\pm$ 0.089   & 0.617 $\pm$ 0.064   \\
Open4D~\cite{Bansal_2020_CVPR} & & 16.681 $\pm$ 2.718  & 0.498 $\pm$ 0.071  &  {\bf 0.477 $\pm$ 0.061}  \\
Ours (w/o T) &    &  16.665 $\pm$ 2.365 & 0.519 $\pm$ 0.074  & 0.538 $\pm$ 0.071   \\
Ours (w/ T) &    & {\bf 16.797 $\pm$ 2.523} & {\bf 0.535 $\pm$  0.080}  &   0.522 $\pm$ 0.075 \\
\bottomrule
\end{tabular}
\caption{\textbf{Held-Out Camera Views: } We contrast the performance of our approach with Open4D~\cite{Bansal_2020_CVPR} to synthesize held-out camera views. There are a total of $2092$ frames used for evaluation. Quantitatively, we achieve similar performance. Importantly, our approach does not require heuristics to compute a foreground and background image. Finally, we further improve performance by incorporating temporal information as an input to the model.}
\label{tab:4d-views-02}
}
\end{table}


\section{3D reconstruction and Depth from Multi-Views}
\label{sss:3drecon}

\begin{figure}
\includegraphics[width=\linewidth]{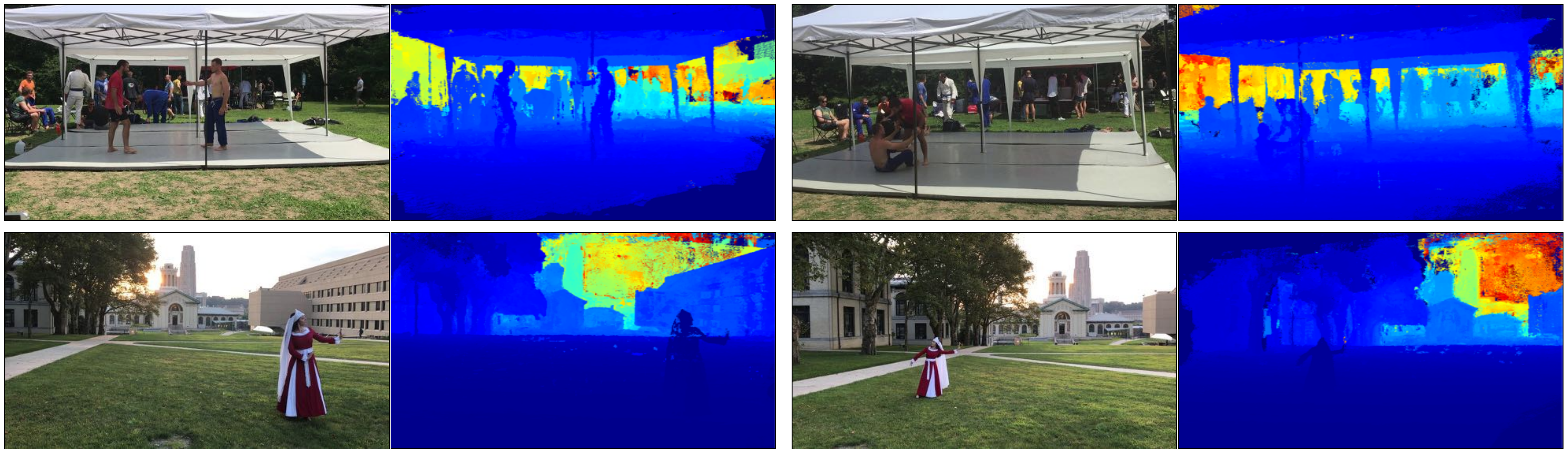}
\caption{\textbf{Depth maps using learned MLPs}: We show depth map for images for various sequences. We use the learned MLPs to select the depth value corresponding to the max $\alpha_i$ value from an array of depth values for a pixel. The ``jet blue'' color corresponds to missing depth values for these images (e.g., the bottom right edge on the depth map of the first image). }
\label{fig:depth-map}
\end{figure}

We use the learned MLPs to construct depth map for a given view. Given an array of depth values for a pixel, we select the depth value corresponding to the maximum $\alpha_i$ value. Figure~\ref{fig:depth-map} shows the depth map for images from various sequences. We do not have ground truth depth values for these sequences.

\begin{figure*}[t]
\includegraphics[width=\linewidth]{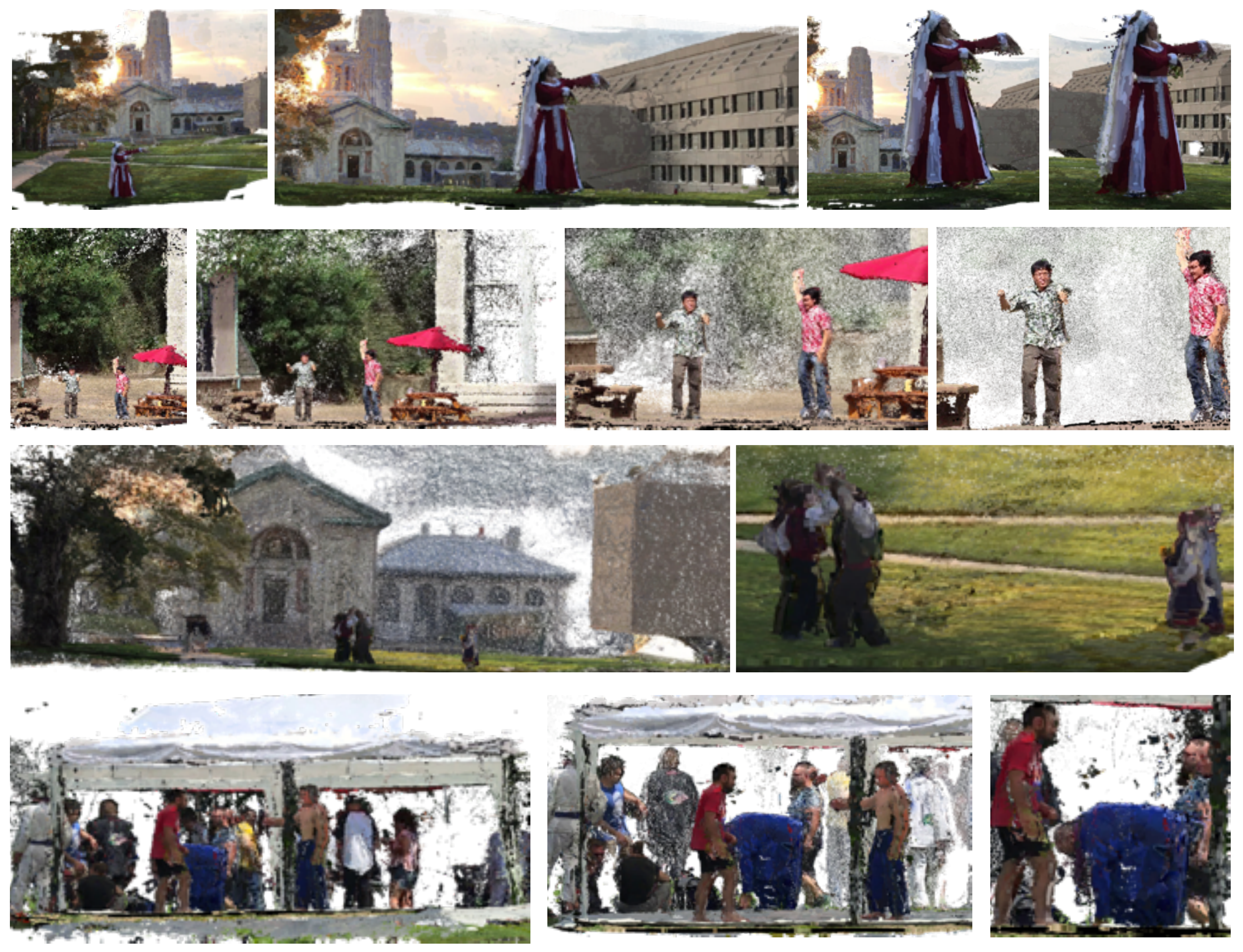}
\caption{\textbf{Dense 3D reconstruction from sparse views}: We show dense 3D point clouds computed using our approach for a specific time instant for four unconstrained multi-view sequences~\cite{Bansal_2020_CVPR}. A user can easily explore the region by navigating the point clouds. We show random views of the point clouds.}
\label{fig:dense3D}
\end{figure*}

Multiple stereo pairs also provide us with dense 3D point clouds. However, correspondences can still be noisy, and using them with noisy camera parameters leads to poor 3D estimates. We observe that the learned MLP enables us to select good 3D points per view that can be accumulated across multi-views to obtain a dense 3D reconstruction. For each pixel, we take the top-$3$ $\alpha_i$ values and check if the corresponding $d_i$ values are in the vicinity of each other (this is done by empirically selecting a distance threshold). If they are, then we select the $3D$ point from a stereo pair corresponding to the maximum $\alpha_i$ value. The process is repeated for all the pixels in the available multi-views.
We show the results of 3D reconstruction using our approach in Fig~\ref{fig:dense3D} for sparse multi-view sequences.
COLMAP~\cite{schoenberger2016sfm,schoenberger2016mvs} struggle to achieve dense 3D reconstruction on these sequences.

\section{Discussion}
\label{sec:discuss}
We propose a novel approach for continuous 3D-4D view synthesis from sparse and wide-baseline multi-view observations.
Leveraging a rich pixel representation that consists of color, depth, and uncertainty information leads to a high performing view-synthesis approach that generalizes well to novel views and unseen time instances.
Our approach can be trained within few minutes from scratch utilizing as few as 1GB of GPU memory. In this work, we strive to provide an extensive analysis of our approach in contrast to existing methods on a wide variety of settings. Importantly, our method works well on numerous settings without incorporating any task-specific or sequence-specific knowledge. 
We see our approach as a first step towards more efficient and general neural rendering techniques via the explicit use of geometric information and hope that it will inspire follow-up work in this exciting field.

\noindent\textbf{Note to reader: } We suggest the reader to see our project page and attached videos for more results and analysis.

\noindent\textbf{Acknowledgements: } AB would like to thank David Forsyth, Deva Ramanan, Minh Vo, and Srinivasa Narasimhan for many wonderful discussions on 3D-4D view synthesis. Many comments and insights from David Forsyth were extremely helpful in designing this work.

\noindent\textbf{Disclaimer: } This academic article may contain images and/or data from sources that are not affiliated with the article submitter. Inclusion should not be construed as approval, endorsement or sponsorship of the submitter, article or its content by any such party.

%
%
\bibliographystyle{splncs04}

\clearpage
\appendix

\section{3D View Synthesis}
\label{app:3dvs}

\subsection{Sparse and Unconstrained Multi-Views} 
\label{app:sparse3d}
We use $24$ time instants from multi-view temporal sequences from the Open4D dataset~\cite{Bansal_2020_CVPR}. The dynamic scenes are captured by a varying number of cameras in these sequences. The number of views vary from $7$ to $11$. We use one held-out view (or camera) for evaluation. Following is the setup for this analysis:

\hrulefill\\
\noindent \textbf{Sequences}

\noindent{\tt WFD-01}: $6$ time-stamps - \{{\tt 2000}, {\tt 2500}, {\tt 3000}, {\tt 3500}, {\tt 4000}, {\tt 4500}\}. Test CAM-ID: \{{\tt 2, 9, 2, 2, 6, 4}\}.

\noindent{\tt WFD-02}: $5$ time-stamps - \{{\tt 1900}, {\tt 3000}, {\tt 3500}, {\tt 4000}, {\tt 4500}\}.  Test CAM-ID: \{{\tt 3, 6, 4, 2, 3}\}.

\noindent{\tt JiuJitsu}: $7$ time-stamps - \{{\tt 3000}, {\tt 3500}, {\tt 4000}, {\tt 4500}, {\tt 5000}, {\tt 5500}, {\tt 6000}\}.  Test CAM-ID: \{{\tt 5, 4, 9, 5, 7, 11, 1}\}.

\noindent{\tt Gangnam}: $3$ time-stamps - \{{\tt 0200}, {\tt 0300}, {\tt 0900}\}.  Test CAM-ID: \{{\tt 4, 4, 4}\}.

\noindent{\tt Jumping}: $3$ time-stamps - \{{\tt 0200}, {\tt 0300}, {\tt 0400}\}.  Test CAM-ID: \{{\tt 0, 0, 0}\}.

\hrulefill

\subsection{Hi-Resolution View Synthesis}
\label{app:hi-res}

We use the following $12$ sequences from LLFF dataset~\cite{mildenhall2019local} for this analysis:

\hrulefill\\
\noindent \textbf{Sequences}: {\tt airplants}, {\tt data2\_apeskeleton}, {\tt data2\_benchflower}, 

\noindent{\tt data2\_bridgecar}, {\tt data2\_chesstable}, {\tt data2\_colorfountain}, 

\noindent{\tt data2\_colorspout}, {\tt data2\_redtoyota}, {\tt data3\_ninjabike}, 

\noindent{\tt data4\_colinepiano}, {\tt data5\_piano}, {\tt pond}.

\noindent \textbf{Test IDs}: For each sequence, we held-out every $8^{th}$ frame for evaluation.

\hrulefill

\subsection{Unbounded Views and Varying Number of Views} 
\label{app:synth}
We use the following $13$ synthetic multi-view sequences for this analysis from the MVS-Synth dataset~\cite{DeepMVS}:

\hrulefill\\
\noindent \textbf{Sequence IDs}: \{{\tt 0000}, {\tt 0001}, {\tt 0002}, {\tt 0003}, {\tt 0004}, {\tt 0005}, {\tt 0006}, {\tt 0007}, {\tt 0008}, {\tt 0009}, {\tt 0010}, {\tt 0011}, {\tt 0012}\}. 

\noindent For each sequence, we held-out every other frame for evaluation: 

\noindent \textbf{Test IDs}: \{{\tt 000}:{\tt 002}:{\tt 098}\}. 

\noindent \textbf{Train IDs}:

\noindent \bm{$10$} views: \{{\tt 003}, {\tt 013}, {\tt 023}, {\tt 033}, {\tt 043}, {\tt 053}, {\tt 063}, {\tt 073}, {\tt 083}, {\tt 093}\}.

\noindent \bm{$20$} views: \{{\tt 003}, {\tt 009}, {\tt 013}, {\tt 019}, {\tt 023}, {\tt 029}, {\tt 033}, {\tt 039}, {\tt 043}, {\tt 049}, {\tt 053}, {\tt 059}, {\tt 063}, {\tt 069}, {\tt 073}, {\tt 079}, {\tt 083}, {\tt 089}, {\tt 093}, {\tt 099}\}.

\noindent \bm{$30$} views: \{{\tt 003}, {\tt 007}, {\tt 009}, {\tt 013}, {\tt 017}, {\tt 019}, {\tt 023}, {\tt 027}, {\tt 029}, {\tt 033}, {\tt 037}, {\tt 039}, {\tt 043}, {\tt 047}, {\tt 049}, {\tt 053}, {\tt 057}, {\tt 059}, {\tt 063}, {\tt 067}, {\tt 069}, {\tt 073}, {\tt 077}, {\tt 079}, {\tt 083}, {\tt 087}, {\tt 089}, {\tt 093}, {\tt 097}, {\tt 099}\}.

\noindent \bm{$40$} views: \{{\tt 001}, {\tt 003}, {\tt 007}, {\tt 009}, {\tt 011}, {\tt 013}, {\tt 017}, {\tt 019}, {\tt 021}, {\tt 023}, {\tt 027}, {\tt 029}, {\tt 031}, {\tt 033}, {\tt 037}, {\tt 039}, {\tt 041}, {\tt 043}, {\tt 047}, {\tt 049}, {\tt 051}, {\tt 053}, {\tt 057}, {\tt 059}, {\tt 061}, {\tt 063}, {\tt 067}, {\tt 069}, {\tt 071}, {\tt 073}, {\tt 077}, {\tt 079}, {\tt 081}, {\tt 083}, {\tt 087}, {\tt 089}, {\tt 091}, {\tt 093}, {\tt 097}, {\tt 099}\}.

\noindent \bm{$50$} views: \{{\tt 001}:{\tt 002}:{\tt 099}\}.

\hrulefill

\subsection{Convergence Analysis} 
\label{app:conv}

In this section, we provide the raw data used in Sec~\ref{ssec:conv}. We use $24$ sparse and unconstrained multi-view sequences (Sec~\ref{app:sparse3d}) from Open4D~\cite{Bansal_2020_CVPR}. Training an epoch on these sequences roughly take $10$ seconds because these are sparse. Table~\ref{tab:app-sparse3d} shows the performance of our model for $10$ epochs (from $10$ seconds to roughly $2$ minutes).  We also use two hi-res (12 MP) datasets for these analysis: {\bf (1)} $12$ sequences (Sec~\ref{app:hi-res}) from LLFF dataset~\cite{mildenhall2019local}; and {\bf (2)} $8$ sequences from Shiny dataset~\cite{Wizadwongsa2021NeX}. We compute the performance of the models for the first $10$ epochs, i.e., from $60$ to $600$ seconds of training. We follow the three settings (as in Sec~\ref{ssec:hi-res}) where we vary the number of stereo-pairs ($K$) and number of 3D points ($N$): {\bf (1)} $(K=50,N=50)$; {\bf (2)} $(K=100,N=100)$; and {\bf (3)} $(K=200,N=200)$. Table~\ref{tab:app-llff-01}, Table~\ref{tab:app-llff-02}, and Table~\ref{tab:app-llff-03} shows the performance for $12$ sequences from LLFF.  Table~\ref{tab:app-shiny-01}, Table~\ref{tab:app-shiny-02}, and Table~\ref{tab:app-shiny-03} shows the performance for $8$ sequences from the Shiny dataset. We observe that our approach gets close to convergence within the first $60$ seconds of training in all the settings.

\begin{table}[h!]
\setlength{\tabcolsep}{7pt}
\def\arraystretch{1.3}
\center
\begin{tabular}{@{}c  c  c c }
\toprule
\textbf{24 sequences} &  \textbf{PSNR}$\uparrow$ & \textbf{SSIM}$\uparrow$   & \textbf{LPIPS} $\downarrow$ \\ 
\midrule
{\bf Num-Epochs} &	 &	 &	\\
1  & 18.181  $\pm$	1.519 &	0.559  $\pm$ 0.079 &	0.533  $\pm$	0.066 \\
2  & 18.139  $\pm$	1.378 &	0.562  $\pm$ 0.077 &	0.528  $\pm$	0.062 \\
3 & 18.016  $\pm$	1.461 &	0.562 $\pm$	0.077 &	0.527 $\pm$	0.063 \\
4 & 18.026 $\pm$	1.420 &	0.563 $\pm$	0.076 &	0.527  $\pm$	0.063 \\
5 & 18.115 $\pm$	1.459 &	0.566 $\pm$	0.076 &	0.523  $\pm$	0.061 \\
6 & 17.877 $\pm$	1.409 &	0.561 $\pm$	0.075 &	0.532  $\pm$	0.061 \\
7 & 17.951 $\pm$	1.456 &	0.562 $\pm$	0.076 &	0.531  $\pm$	0.059 \\
8 & 17.918 $\pm$	1.511 &	0.562 $\pm$	0.077 &	0.532  $\pm$	0.061 \\
9 & 17.951 $\pm$	1.475 &	0.562 $\pm$	0.076 &	0.531  $\pm$	0.058 \\
10 & 17.948 $\pm$	1.472 &	0.562 $\pm$	0.077 &	0.534  $\pm$	0.061 \\
\midrule
NeRF~\cite{mildenhall2020nerf} & 13.693 $\pm$ 2.050 &  0.317 $\pm$ 0.094  & 0.713 $\pm$ 0.089 \\
DS-NeRF~\cite{deng2021depth}  &  14.531 $\pm$ 2.603  & 0.316 $\pm$ 0.099  & 0.757 $\pm$ 0.040 \\ 
LLFF~\cite{mildenhall2019local} &  15.187 $\pm$ 2.166 & 0.384 $\pm$ 0.082  & 0.602 $\pm$ 0.090   \\
\bottomrule
\end{tabular}
\caption{{\bf Sparse and Unconstrained Multi-Views }: We follow the evaluation criterion in Table~\ref{tab:3d-sparse}. We observe that our model gets the best performance in the the first $10$ seconds of training. We contrast the performance of NeRF and DS-NeRF which takes $420$ minutes of training on a single NVIDIA V100 GPU. We also show the performance of LLFF which is an off-the-shelf method and does not require training.}
\label{tab:app-sparse3d}
\end{table}

\begin{table}
\setlength{\tabcolsep}{7pt}
\def\arraystretch{1.3}
\center
\begin{tabular}{@{}c  c  c c }
\toprule
\textbf{12 sequences} &  \textbf{PSNR}$\uparrow$ & \textbf{SSIM}$\uparrow$   & \textbf{LPIPS} $\downarrow$ \\ 
\midrule
{\bf Num-Epochs} &	 &	 &	\\
1 &	20.519 $\pm$	2.805 &	0.589 $\pm$	0.137 &	0.445 $\pm$	0.076 \\
2 &	20.638 $\pm$	2.736 &	0.592 $\pm$	0.137 &	0.437 $\pm$	0.076 \\
3 &	20.744 $\pm$	2.772 &	0.593 $\pm$	0.137 &	0.435 $\pm$	0.076 \\
4 &	20.791 $\pm$	2.783 &	0.593 $\pm$	0.137 &	0.433 $\pm$	0.075 \\
5 &	20.761 $\pm$	2.774 &	0.593 $\pm$	0.137 &	0.433 $\pm$	0.076 \\
6 &	20.798 $\pm$	2.787 &	0.594 $\pm$	0.136 &	0.429 $\pm$	0.076 \\
7 &	20.829 $\pm$	2.807 &	0.594 $\pm$	0.136 &	0.428 $\pm$	0.074 \\
8 &	20.832 $\pm$	2.803 &	0.594 $\pm$	0.136 &	0.427 $\pm$	0.075 \\
9 &	20.841 $\pm$	2.802 &	0.595 $\pm$	0.136 &	0.425 $\pm$	0.074 \\
10 &	20.839 $\pm$	2.798 &	0.594 $\pm$	0.136 &	0.426 $\pm$	0.075 \\
\midrule
NeRF-2M &  21.741  $\pm$ 2.985  & 0.602  $\pm$ 0.147  & 0.584 $\pm$ 0.087\\
\bottomrule
\end{tabular}
\caption{{\bf LLFF-12 sequences and $(K=50,N=50)$}: We use $50$ stereo-pairs and $50$ 3D points. We follow the evaluation criterion in Table~\ref{tab:3d-scalable}. We observe that our model gets close to the best performing model in the the first $60$ seconds of training. For reference, we also show the performance of NeRF which takes $64$ hours of training on a single NVIDIA V100 GPU.}
\label{tab:app-llff-01}
\end{table}

\begin{table}
\setlength{\tabcolsep}{7pt}
\def\arraystretch{1.3}
\center
\begin{tabular}{@{}c  c  c c }
\toprule
\textbf{12 sequences} &  \textbf{PSNR}$\uparrow$ & \textbf{SSIM}$\uparrow$   & \textbf{LPIPS} $\downarrow$ \\ 
\midrule
{\bf Num-Epochs} &	 &	 &	\\
1&	20.491 $\pm$		2.966&	0.590 $\pm$		0.140&	0.494 $\pm$		0.079\\
2&	20.742 $\pm$		2.779&	0.594 $\pm$		0.138&	0.479 $\pm$		0.079\\
3&	20.708 $\pm$		2.851&	0.593 $\pm$		0.139&	0.479 $\pm$		0.080\\
4&	20.765 $\pm$		2.829&	0.595 $\pm$		0.138&	0.473 $\pm$		0.078\\
5&	20.849 $\pm$		2.783&	0.595 $\pm$		0.137&	0.472 $\pm$		0.079\\
6&	20.878 $\pm$		2.787&	0.596 $\pm$		0.137&	0.467 $\pm$		0.079\\
7&	20.878 $\pm$		2.807&	0.596 $\pm$		0.137&	0.466 $\pm$		0.078\\
8&	20.914 $\pm$		2.806&	0.597 $\pm$		0.137&	0.464 $\pm$		0.078\\
9&	20.938 $\pm$		2.801&	0.597 $\pm$		0.136&	0.462 $\pm$		0.079\\
10&	20.958 $\pm$		2.805&	0.597 $\pm$		0.136&	0.461 $\pm$		0.080\\
\midrule
NeRF-2M &  21.741  $\pm$ 2.985  & 0.602  $\pm$ 0.147  & 0.584 $\pm$ 0.087\\
\bottomrule
\end{tabular}
\caption{{\bf LLFF-12 sequences and $(K=100,N=100)$}: We use $100$ stereo-pairs and $100$ 3D points. We follow the evaluation criterion in Table~\ref{tab:3d-scalable}. We observe that our model gets close to the best performing model in the the first $60$ seconds of training. For reference, we also show performance of a NeRF model that takes $64$ hours of training on a single NVIDIA V100 GPU.}
\label{tab:app-llff-02}
\end{table}

\begin{table}
\setlength{\tabcolsep}{7pt}
\def\arraystretch{1.3}
\center
\begin{tabular}{@{}c  c  c c }
\toprule
\textbf{12 sequences} &  \textbf{PSNR}$\uparrow$ & \textbf{SSIM}$\uparrow$   & \textbf{LPIPS} $\downarrow$ \\ 
\midrule
{\bf Num-Epochs} &	 &	 &	\\
1 &	20.240 $\pm$	2.955 &	0.586  $\pm$	0.141 &	0.531  $\pm$	0.083 \\
2 &	20.433 $\pm$	2.859 &	0.587  $\pm$	0.139 &	0.523  $\pm$	0.082 \\
3 &	20.474 $\pm$	2.842 &	0.586  $\pm$	0.138 &	0.518  $\pm$	0.083 \\
4 &	20.519 $\pm$	2.808 &	0.590  $\pm$	0.137 &	0.507  $\pm$	0.082 \\
5 &	20.538 $\pm$	2.816 &	0.590  $\pm$	0.137 &	0.506  $\pm$	0.081 \\
6 &	20.633 $\pm$	2.795 &	0.591  $\pm$	0.136 &	0.504  $\pm$	0.081 \\
7 &	20.630 $\pm$	2.825 &	0.590  $\pm$	0.136 &	0.501  $\pm$	0.082 \\
8 &	20.679 $\pm$	2.841 &	0.591  $\pm$	0.136 &	0.499  $\pm$	0.081 \\
9 &	20.783 $\pm$	2.777 &	0.592  $\pm$	0.136 &	0.496  $\pm$	0.081 \\
10 &	20.799 $\pm$	2.772 &	0.592  $\pm$	0.136 &	0.493  $\pm$	0.081 \\
\midrule
NeRF-2M &  21.741  $\pm$ 2.985  & 0.602  $\pm$ 0.147  & 0.584 $\pm$ 0.087\\
\bottomrule
\end{tabular}
\caption{{\bf LLFF-12 sequences and $(K=200,N=200)$}: We use $200$ stereo-pairs and $200$ 3D points. We follow the evaluation criterion in Table~\ref{tab:3d-scalable}. We observe that our model gets close to the best performing model in the the first $60$ seconds of training. For reference, we also show performance of a NeRF model that takes $64$ hours of training on a single NVIDIA V100 GPU.}
\label{tab:app-llff-03}
\end{table}

\begin{table}[h!]
\setlength{\tabcolsep}{7pt}
\def\arraystretch{1.2}
\center
\begin{tabular}{@{}c  c  c c }
\toprule
\textbf{8 sequences} &  \textbf{PSNR}$\uparrow$ & MC\textbf{SSIM}$\uparrow$   & \textbf{LPIPS} $\downarrow$ \\ 
\midrule
{\bf Num-Epochs} &	 &	 &	\\
1 &	22.184 $\pm$	4.211 & 0.793 $\pm$	0.142 &	0.268 $\pm$	0.110\\
2 &	22.270 $\pm$	4.321 &	0.795 $\pm$	0.142 &	0.263 $\pm$	0.108\\
3 &	22.316 $\pm$	4.372 &	0.795 $\pm$	0.141 &	0.261 $\pm$	0.107\\
4 &	22.348 $\pm$	4.379 &	0.796 $\pm$	0.141 &	0.260 $\pm$	0.107\\
5 &	22.234 $\pm$	4.399 &	0.795 $\pm$	0.141 &	0.259 $\pm$	0.107\\
6 &	22.395 $\pm$	4.542 &	0.795 $\pm$	0.141 &	0.258 $\pm$	0.107\\
7 &	22.375 $\pm$	4.579 &	0.795 $\pm$	0.142 &	0.258 $\pm$	0.017\\
8 &	22.386 $\pm$	4.625 &	0.795 $\pm$	0.142 &	0.257 $\pm$	0.107\\
9 &	22.430 $\pm$	4.677 &	0.795 $\pm$	0.142 &	0.257 $\pm$	0.107\\
10 &	22.430 $\pm$	4.740 &	0.795 $\pm$	0.142 &	0.256 $\pm$	0.107\\
\midrule
NeRF~\cite{mildenhall2020nerf} &     22.009 $\pm$ 3.148  & 0.757 $\pm$ 0.156  & 0.487 $\pm$ 0.180 \\
NeRF-2M    & 21.457 $\pm$ 3.657  & 0.751 $\pm$ 0.155  &   0.498 $\pm$ 0.153  \\
NeX~\cite{Wizadwongsa2021NeX} &    22.292 $\pm$ 3.137  & 0.774 $\pm$ 0.152  & 0.423  $\pm$ 0.156\\
\bottomrule
\end{tabular}
\caption{{\bf Shiny dataset and $(K=50,N=50)$}: We use $50$ stereo-pairs and $50$ 3D points. We follow the evaluation criterion in Table~\ref{tab:shiny}. We observe that our model gets close to the best performing model in the first $60$ seconds of training. For reference, we also show the performance of NeRF models. We also show the performance of NeX models take 24-30 hours of training for one-fourth resolution. }
\label{tab:app-shiny-01}
\vspace{-0.5cm}
\end{table}

\begin{table}[h!]
\setlength{\tabcolsep}{7pt}
\def\arraystretch{1.2}
\center
\begin{tabular}{@{}c  c  c c }
\toprule
\textbf{8 sequences} &  \textbf{PSNR}$\uparrow$ & MC\textbf{SSIM}$\uparrow$   & \textbf{LPIPS} $\downarrow$ \\ 
\midrule
{\bf Num-Epochs} &	 &	 &	\\
1 &	22.519 $\pm$	4.197 &	0.799 $\pm$	0.142 &	0.287 $\pm$	0.126\\
2 &	22.647 $\pm$	4.264 &	0.801 $\pm$	0.140 &	0.281 $\pm$	0.123\\
3 &	22.665 $\pm$	4.305 &	0.801 $\pm$	0.140 &	0.279 $\pm$	0.123\\
4 &	22.718 $\pm$	4.347 &	0.801 $\pm$	0.140 &	0.278 $\pm$	0.123\\
5 &	22.745 $\pm$	4.369 &	0.801 $\pm$	0.141 &	0.277 $\pm$	0.123\\
6 &	22.756 $\pm$	4.437 &	0.801 $\pm$	0.141 &	0.275 $\pm$	0.123\\
7 &	22.812 $\pm$	4.499 &	0.802 $\pm$	0.141 &	0.273 $\pm$	0.123\\
8 &	22.754 $\pm$	4.401 &	0.802 $\pm$	0.141 &	0.272 $\pm$	0.123\\
9 &	22.843 $\pm$	4.455 &	0.802 $\pm$	0.141 &	0.270 $\pm$	0.123\\
10 &	22.868 $\pm$	4.588 &	0.802 $\pm$	0.141 &	0.269 $\pm$	0.123\\
\midrule
NeRF~\cite{mildenhall2020nerf} &     22.009 $\pm$ 3.148  & 0.757 $\pm$ 0.156  & 0.487 $\pm$ 0.180 \\
NeRF-2M   & 21.457 $\pm$ 3.657  & 0.751 $\pm$ 0.155  &   0.498 $\pm$ 0.153  \\
NeX~\cite{Wizadwongsa2021NeX} &    22.292 $\pm$ 3.137  & 0.774 $\pm$ 0.152  & 0.423  $\pm$ 0.156\\
\bottomrule
\end{tabular}
\caption{{\bf Shiny dataset and $(K=100,N=100)$}: We use $100$ stereo-pairs and $100$ 3D points. We follow the evaluation criterion in Table~\ref{tab:shiny}. We observe that our model gets close to the best performing model in the first $60$ seconds of training. For reference, we also show the performance of NeRF models. We also show the performance of NeX models take 24-30 hours of training for one-fourth resolution. }
\label{tab:app-shiny-02}
\vspace{-0.5cm}
\end{table}

\begin{table}[h!]
\setlength{\tabcolsep}{6pt}
\def\arraystretch{1.2}
\center
\begin{tabular}{@{}c  c  c c }
\toprule
\textbf{8 sequences} &  \textbf{PSNR}$\uparrow$ & MC\textbf{SSIM}$\uparrow$   & \textbf{LPIPS} $\downarrow$ \\ 
\midrule
{\bf Num-Epochs} &	 &	 &	\\
1 &	22.563 $\pm$	4.269 &	0.799 $\pm$	0.147 &	0.311 $\pm$	0.141\\
2 &	22.734 $\pm$	4.385 &	0.800 $\pm$	0.146 &	0.303 $\pm$	0.138\\
3 &	22.788 $\pm$	4.413 &	0.801 $\pm$	0.145 &	0.301 $\pm$	0.137\\
4 &	22.838 $\pm$	4.428 &	0.802 $\pm$	0.145 &	0.298 $\pm$	0.137\\
5 &	22.847 $\pm$	4.467 &	0.802 $\pm$	0.145 &	0.294 $\pm$	0.135\\
6 &	22.878 $\pm$	4.478 &	0.802 $\pm$	0.145 &	0.293 $\pm$	0.135\\
7 &	22.916 $\pm$	4.543 &	0.802 $\pm$	0.145 &	0.291 $\pm$	0.134\\
8 &	22.934 $\pm$	4.571 &	0.802 $\pm$	0.145 &	0.289 $\pm$	0.133\\
9 &	22.947 $\pm$	4.603 &	0.803 $\pm$	0.144 &	0.287 $\pm$	0.133\\
10 &	23.016 $\pm$	4.698 &	0.803 $\pm$	0.144 &	0.285 $\pm$	0.132\\
\midrule
NeRF~\cite{mildenhall2020nerf} &     22.009 $\pm$ 3.148  & 0.757 $\pm$ 0.156  & 0.487 $\pm$ 0.180 \\
NeRF-2M &  21.457 $\pm$ 3.657  & 0.751 $\pm$ 0.155  &   0.498 $\pm$ 0.153  \\
NeX~\cite{Wizadwongsa2021NeX} &    22.292 $\pm$ 3.137  & 0.774 $\pm$ 0.152  & 0.423  $\pm$ 0.156\\
\bottomrule
\end{tabular}
\caption{{\bf Shiny dataset and $(K=200,N=200)$}: We use $200$ stereo-pairs and $200$ 3D points. We follow the evaluation criterion in Table~\ref{tab:shiny}. We observe that our model gets close to the best performing model in the first $60$ seconds of training. For reference, we also show the performance of NeRF models. We also show the performance of NeX models take 24-30 hours of training for one-fourth resolution. }
\label{tab:app-shiny-03}
\vspace{-1.cm}
\end{table}

\begin{table}
\setlength{\tabcolsep}{6pt}
\def\arraystretch{1.1}
\center
\begin{tabular}{@{}c  c  c c }
\toprule
\textbf{Open4D-24 sequences} &  \textbf{PSNR}$\uparrow$ & \textbf{SSIM}$\uparrow$   & \textbf{LPIPS} $\downarrow$ \\ 
\midrule
{\tt no gamma}  &	17.034 $\pm$	2.663 &	0.539	$\pm$ 0.099	& 0.539 $\pm$	0.075 \\
{\tt no spatial}  &	18.387 $\pm$	2.308 &	0.569 $\pm$	0.089 &	0.527 $\pm$	0.066 \\
{\tt no entropy}  &	17.893 $\pm$	1.481 &	0.551 $\pm$	0.074 &	0.573 $\pm$	0.064 \\
{\tt direct MLP}  &	17.905 $\pm$ 	1.808 &	0.562 $\pm$	0.081 &	0.546 $\pm$	0.064 \\
{\tt full} & 17.948 $\pm$	1.472 &	0.562 $\pm$	0.077 &	0.534 $\pm$	0.061 \\
\bottomrule
\textbf{LLFF-12 sequences} &  \textbf{PSNR}$\uparrow$ & \textbf{SSIM}$\uparrow$   & \textbf{LPIPS} $\downarrow$ \\ 
\midrule
{\tt no gamma} & 18.831 $\pm$	2.904 & 	0.579 $\pm$	0.133 & 	0.444 $\pm$	0.070\\
{\tt no spatial} & 20.536 $\pm$	2.798 & 	0.594 $\pm$	0.135 & 	0.429 $\pm$	0.074\\
{\tt no entropy} & 20.792 $\pm$	2.777 &	0.595 $\pm$	0.136 &	0.435 $\pm$	0.075\\
{\tt direct MLP} & 20.816 $\pm$	2.796 &	0.593 $\pm$	0.135 &	0.423 $\pm$	0.076\\
{\tt full} & 20.834 $\pm$	2.784 &	0.594 $\pm$	0.136 &	0.426 $\pm$	0.075\\
\bottomrule
\textbf{Shiny-8 sequences} &  \textbf{PSNR}$\uparrow$ & MC\textbf{SSIM}$\uparrow$   & \textbf{LPIPS} $\downarrow$ \\ 
\midrule
{\tt no gamma} & 17.724 $\pm$	2.313 & 	0.765 $\pm$	0.138 & 	0.288 $\pm$	0.094\\
{\tt no spatial} & 21.047 $\pm$	3.177 & 	0.791 $\pm$	0.140 & 	0.266 $\pm$	0.097\\
{\tt no entropy} & 22.529 $\pm$	4.787 &	0.796 $\pm$	0.142 &	0.258 $\pm$	0.110\\
{\tt direct MLP} & 22.419 $\pm$	4.757 &	0.794 $\pm$	0.142 &	0.259 $\pm$	0.105\\
{\tt full}& 22.430 $\pm$	4.748 &	0.795 $\pm$	0.142 &	0.256 $\pm$	0.108\\
\bottomrule
\end{tabular}
\caption{{\bf }: We study the influence of different components on our approach and see their benefits in our approach.}
\label{tab:ma-no}
\vspace{-1.cm}
\end{table}

\section{4D View Synthesis}

We use temporal sequences from Open4D dataset~\cite{Bansal_2020_CVPR} for these analysis.

\subsection{Unseeen Temporal Sequences}
\label{app:open4d}

We use all the available views of the following $5$ publicly available temporal sequences.

\hrulefill\\
\noindent \textbf{Sequences} 

\noindent{\tt WFD-01}: Training - \{{\tt 0011}:{\tt 0411}\}. Testing - \{{\tt 0412}:{\tt 0511}\}.

\noindent{\tt WFD-02}: Training - \{{\tt 0400}:{\tt 0800}\}. Testing - \{{\tt 0801}:{\tt 0900}\}.

\noindent{\tt JiuJitsu}: Training - \{{\tt 0001}:{\tt 0400}\}. Testing - \{{\tt 0401}:{\tt 0500}\}.

\noindent{\tt Gangnam}: Training - \{{\tt 0100}:{\tt 0400}\}. Testing - \{{\tt 0401}:{\tt 0500}\}.

\noindent{\tt Birds}: Training - \{{\tt 0309}:{\tt 0709}\}. Testing - \{{\tt 0710}:{\tt 0809}\}.

\hrulefill

\subsection{Held-out Camera Views}
\label{app:open4d-ho}

We held-out one camera view from the following $5$ publicly available temporal sequences.

\hrulefill\\
\noindent \textbf{Sequences} 

\noindent{\tt WFD-01}: time - \{{\tt 0011}:{\tt 0511}\}. Test CAM-ID: \{{\tt 4}\}.

\noindent{\tt WFD-02}: time - \{{\tt 0400}:{\tt 0900}\}. Test CAM-ID: \{{\tt 4}\}.

\noindent{\tt JiuJitsu}: time - \{{\tt 0001}:{\tt 0500}\}. Test CAM-ID: \{{\tt 0}\}.


\noindent{\tt Gangnam}: time - \{{\tt 0100}:{\tt 0500}\}. Test CAM-ID: \{{\tt 4}\}.

\noindent{\tt Birds}: time - \{{\tt 0309}:{\tt 0809}\}. Test CAM-ID: \{{\tt 7}\}.

\hrulefill

\section{More Analysis}
\label{ss:more}

We run more analysis on our model for various settings and study their impact on performance of our approach. In these experiments, we train the model for $10$ epochs using LLFF-12 sequences (Sec~\ref{app:hi-res}) and Shiny Dataset~\cite{Wizadwongsa2021NeX}, and we use $K=50$ stereo-pairs and $N=50$ 3D points. We also use $24$ sparse and unconstrained sequences from Open4D (Sec~\ref{app:sparse3d}).

\noindent\textbf{Number of Filters: } We vary the number of filters in our MLP model, $n_f = \{16,32.64,128,256,512\}$. Our default setting is $n_f = 256$. Table~\ref{tab:ma-nf1} shows the performance for Open4D-24 sequences, LLFF-12 sequences and Shiny dataset. The performance improves as we increase the number of filters. The use of $n_f = 256$ is a good balance between performance and size of model. We also observe that we can make extremely compact model at the loss of slight performance.

\begin{table}
\setlength{\tabcolsep}{7pt}
\def\arraystretch{1.3}
\center
\begin{tabular}{@{}c  c  c c }
\toprule
\textbf{Open4D-24 sequences} &  \textbf{PSNR}$\uparrow$ & \textbf{SSIM}$\uparrow$   & \textbf{LPIPS} $\downarrow$ \\ 
\midrule
{\bf Num-Filters} &	 &	 &	\\

16 & 17.716 $\pm$	1.485 &	0.555 $\pm$	0.079 &	0.538 $\pm$	0.067 \\
32 & 17.775 $\pm$	1.453 &	0.557 $\pm$	0.081 &	0.536 $\pm$	0.069 \\
64 & 17.828 $\pm$	1.584 &	0.556 $\pm$	0.082 &	0.541 $\pm$	0.068 \\
128 & 17.985 $\pm$	1.561 &	0.561 $\pm$	0.079 &	0.535 $\pm$	0.066 \\
{\tt default} = 256  &	17.948 $\pm$	1.472 &	0.562 $\pm$	0.077 &	0.534 $\pm$	0.061 \\
512 & 18.091 $\pm$ 	1.707 &	0.564 $\pm$	0.081 &	0.534  $\pm$	0.067 \\
\bottomrule
\textbf{LLFF-12 sequences} &  \textbf{PSNR}$\uparrow$ & \textbf{SSIM}$\uparrow$   & \textbf{LPIPS} $\downarrow$ \\ 
\midrule
{\bf Num-Filters} &	 &	 &	\\
16 & 20.320 $\pm$ 2.401 &	0.591 $\pm$	0.135 &	0.441 $\pm$	0.07\\
32 & 20.667 $\pm$ 2.755	 &	0.592 $\pm$	0.136 &	0.433 $\pm$	0.075\\
64 & 20.737 $\pm$	2.818 &	0.594 $\pm$	0.136 &	0.429 $\pm$	0.075\\
128 & 20.771 $\pm$	2.785 &	0.594 $\pm$	0.136 &	0.428 $\pm$	0.075\\
{\tt default} = 256 & 20.834 $\pm$	2.784 &	0.594 $\pm$	0.136 &	0.426 $\pm$	0.075\\
512 & 20.833 $\pm$	2.781 &	0.594 $\pm$	0.135 &	0.422 $\pm$	0.075\\
\bottomrule
\textbf{Shiny-8 sequences} &  \textbf{PSNR}$\uparrow$ & MC\textbf{SSIM}$\uparrow$   & \textbf{LPIPS} $\downarrow$ \\ 
\midrule
{\bf Num-Filters} &	 &	 &	\\

16 & 21.202 $\pm$	4.283 &	0.787 $\pm$	0.144 &	0.273 $\pm$	0.110\\
32 & 22.058 $\pm$	4.150 &	0.794 $\pm$	0.141 &	0.262 $\pm$	0.105\\
64 & 22.188 $\pm$	4.322 &	0.795 $\pm$	0.142 &	0.261 $\pm$	0.106\\
128 & 20.371$\pm$	4.600 &	0.795 $\pm$	0.142 &	0.259 $\pm$	0.109\\
{\tt default} = 256 & 22.430 $\pm$	4.748 &	0.795 $\pm$	0.142 &	0.256 $\pm$	0.108\\
512 & 22.516 $\pm$	4.809 &	0.795 $\pm$	0.143 &	0.254 $\pm$	0.108\\
\bottomrule
\end{tabular}
\caption{{\bf Number of Filters}: We follow the evaluation criterion in Table~\ref{tab:3d-sparse} for Open4D-24 sequences, Table~\ref{tab:3d-scalable} for LLFF-12 sequences and Table~\ref{tab:shiny} for Shiny-8 sequences. The performance improves as we increase the number of filters. We use $n_f = 256$ as a good balance between performance and size of model.  }
\label{tab:ma-nf1}
\vspace{-1.cm}
\end{table}

\noindent\textbf{Number of Layers: } We vary the number of layers in our MLP model, $n_l = \{1,2,3,4,5,6\}$. Our default setting is $n_l = 5$. Table~\ref{tab:ma-nl} shows the performance for Open4D-24 sequences, LLFF-12 sequences and Shiny dataset respectively.

\begin{table}
\setlength{\tabcolsep}{7pt}
\def\arraystretch{1.3}
\center
\begin{tabular}{@{}c  c  c c }
\toprule
\textbf{Open4D-24 sequences} &  \textbf{PSNR}$\uparrow$ & \textbf{SSIM}$\uparrow$   & \textbf{LPIPS} $\downarrow$ \\ 
\midrule
{\bf Num-Layers} &	 &	 &	\\

1 &	17.601 $\pm$	1.779 &	0.527 $\pm$	0.086 &	0.587 $\pm$	0.101 \\
2 &	18.014  $\pm$	1.688 &	0.549  $\pm$ 0.081 &	0.555 $\pm$	0.082 \\
3 &	17.971 $\pm$	1.621 &	0.555 $\pm$	0.081 &	0.541  $\pm$	0.073 \\
4 &	18.066 $\pm$	1.565 &	0.559 $\pm$	0.081 &	0.535 $\pm$	0.065 \\
{\tt default} = 5 &	17.948 $\pm$	1.472 &	0.562 $\pm$	0.077 &	0.534 $\pm$	0.061 \\
6 &	17.996 $\pm$	1.669 &	0.562 $\pm$	0.079 &	0.534 $\pm$	0.067\\
\bottomrule
\textbf{LLFF-12 sequences} &  \textbf{PSNR}$\uparrow$ & \textbf{SSIM}$\uparrow$   & \textbf{LPIPS} $\downarrow$ \\ 
\midrule
{\bf Num-Layers} &	 &	 &	\\

1 & 20.433 $\pm$	2.972 & 	0.573 $\pm$	0.138 & 	0.450 $\pm$	0.090\\
2 & 20.707 $\pm$	2.874 & 	0.588 $\pm$	0.136 & 	0.432 $\pm$	0.081\\
3 & 20.833 $\pm$	2.805 &	0.593 $\pm$	0.136 &	0.424 $\pm$	0.076\\
4 & 20.828 $\pm$	2.818 &	0.594 $\pm$	0.136 &	0.424 $\pm$	0.075\\
{\tt default} = 5 & 20.834 $\pm$	2.784 &	0.594 $\pm$	0.136 &	0.426 $\pm$	0.075\\
6 & 20.820 $\pm$	2.775 &	0.594 $\pm$	0.136 &	0.426 $\pm$	0.075\\
\bottomrule
\textbf{Shiny-8 sequences} &  \textbf{PSNR}$\uparrow$ & MC\textbf{SSIM}$\uparrow$   & \textbf{LPIPS} $\downarrow$ \\ 
\midrule
{\bf Num-Layers} &	 &	 &	\\

1 & 22.334 $\pm$	4.485 &	0.793 $\pm$	0.142 &	0.256 $\pm$	0.108\\
2 & 22.447 $\pm$	4.659 &	0.794 $\pm$	0.143 &	0.254 $\pm$	0.108\\
3 & 22.412 $\pm$	4.688 &	0.795 $\pm$	0.143 &	0.254 $\pm$	0.108\\
4 & 20.391$\pm$	4.730 &	0.794 $\pm$	0.142 &	0.255 $\pm$	0.108\\
{\tt default} = 5 & 22.430 $\pm$	4.748 &	0.795 $\pm$	0.142 &	0.256 $\pm$	0.108\\
6 & 22.367 $\pm$	4.657 &	0.795 $\pm$	0.143 &	0.256 $\pm$	0.108\\
\bottomrule
\end{tabular}
\caption{{\bf Number of Layers}: We follow the evaluation criterion in Table~\ref{tab:3d-sparse} for Open4D-24 sequences, Table~\ref{tab:3d-scalable} for LLFF-12 sequences and Table~\ref{tab:shiny} for Shiny-8 sequences. We use $n_l = 5$ in this work. }
\label{tab:ma-nl}
\vspace{-1.1cm}
\end{table}

\noindent\textbf{Influence of Gamma:} We use $\gamma$ as a correction term that helps us to obtain sharp outputs. Table~\ref{tab:ma-no} (first row))  shows the performance for Open4D-24 sequences, LLFF-12 sequences and Shiny dataset. We observe that the additional $\gamma$ term helps in inpainting the missing information. 

\noindent\textbf{Influence of Spatial Information:} The second row in Table~\ref{tab:ma-no} shows the performance of our approach without using spatial information as an input to MLP. We observe that using spatial information enables us to provide smooth outputs and better inpaints missing information.

\noindent\textbf{Influence of Uncertainty/Entropy:} The third row in Table~\ref{tab:ma-no} shows the performance of our approach without using the uncertainty of the depth estimates ($\bm{\mathfrak{H}}$). Using uncertainty provides slightly better performance.

\noindent\textbf{Direct MLP: } Finally, we observe the benefits of using depth explicitly in computing $\bm{\alpha}$ to do a proper color composition. The fourth row in Table~\ref{tab:ma-no} shows the performance for Open4D-24 sequences, LLFF-12 sequences and Shiny dataset. We observe that using depth explicitly allows to do better view synthesis.

\end{document}